\documentclass[10pt,twocolumn,letterpaper]{article}

\usepackage[pagenumbers]{cvpr} 

\usepackage{graphicx}
\usepackage{amsmath}
\usepackage{amssymb}
\usepackage{booktabs}
\makeatletter
\@namedef{ver@everyshi.sty}{}
\makeatother
\usepackage{tikz}
\usepackage{comment}
\usepackage{color}
\usepackage{multicol}
\usepackage{multirow}
\usepackage{makecell}
\usepackage{subcaption}
\usepackage{microtype}
\usepackage{enumitem}
\usepackage{float}

\newcommand{\boldparagraph}[1]{\vspace{0.1em}\noindent{\bf #1} }
\newcommand{\OURS}{DeepLSD}
\newcommand{\fst}[1]{\textbf{#1}}
\newcommand{\snd}[1]{\underline{#1}}

\usepackage[pagebackref,breaklinks,colorlinks]{hyperref}

\usepackage[capitalize]{cleveref}
\crefname{section}{Sec.}{Secs.}
\Crefname{section}{Section}{Sections}
\Crefname{table}{Table}{Tables}
\crefname{table}{Tab.}{Tabs.}

\begin{document}

\title{DeepLSD: Line Segment Detection and Refinement with Deep Image Gradients}

\newcommand{\autSpace}{-0.2}
\author{
Rémi Pautrat${}^1$ \and
\hspace{\autSpace cm}
Daniel Barath${}^1$ \and
\hspace{\autSpace cm}
Viktor Larsson${}^2$ \and
\hspace{\autSpace cm}
Martin R. Oswald${}^{1, 3}$ \and
\hspace{\autSpace cm}
Marc Pollefeys${}^{1, 4}$
\and
${}^1$ \normalsize{Department of Computer Science, ETH Zurich}
\and
${}^2$ \normalsize{Lund University}
\and
${}^3$ \normalsize{University of Amsterdam}
\and
${}^4$ \normalsize{Microsoft Mixed Reality and AI Zurich Lab}}
\maketitle

\begin{abstract}

Line segments are ubiquitous in our human-made world and are increasingly used in vision tasks. They are complementary to feature points thanks to their spatial extent and the structural information they provide. Traditional line detectors based on the image gradient are extremely fast and accurate, but lack robustness in noisy images and challenging conditions. Their learned counterparts are more repeatable and can handle challenging images, but at the cost of a lower accuracy and a bias towards wireframe lines. We propose to combine traditional and learned approaches to get the best of both worlds: an accurate and robust line detector that can be trained in the wild without ground truth lines. Our new line segment detector, DeepLSD, processes images with a deep network to generate a line attraction field, before converting it to a surrogate image gradient magnitude and angle, which is then fed to any existing handcrafted line detector. Additionally, we propose a new optimization tool to refine line segments based on the attraction field and vanishing points. This refinement improves the accuracy of current deep detectors by a large margin. We demonstrate the performance of our method on low-level line detection metrics, as well as on several downstream tasks using multiple challenging datasets.
The source code and models are available at \url{https://github.com/cvg/DeepLSD}.

\end{abstract}

\section{Introduction}
%

\begin{figure}[t]
    \centering
    \scriptsize
    \setlength{\tabcolsep}{0.3mm}
    \newcommand{\sz}{0.24}
    \begin{tabular}{cc}
        \includegraphics[width=\sz\textwidth]{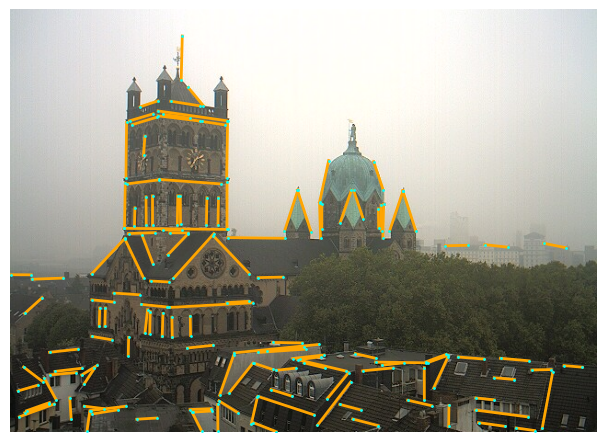} &
        \includegraphics[width=\sz\textwidth]{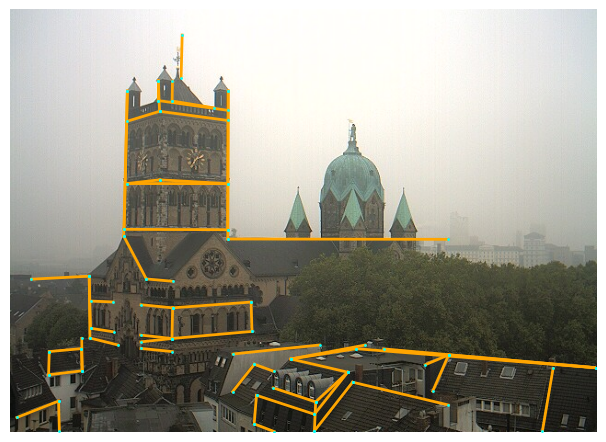} \\
        LSD~\cite{von2008lsd} & HAWP~\cite{hawp} \vspace{0.1cm} \\
        \includegraphics[width=\sz\textwidth]{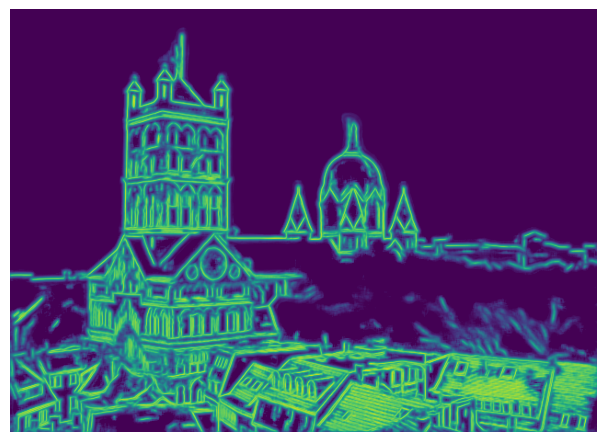} &
        \includegraphics[width=\sz\textwidth]{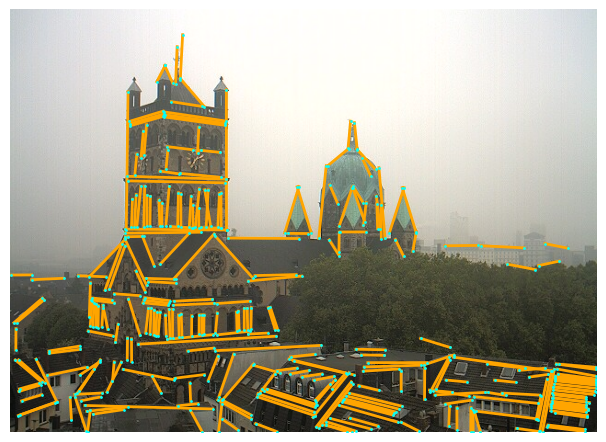} \\
        Line distance field & Ours \\[-5pt]
    \end{tabular}
    \caption{\textbf{Line detection in the wild.} \textbf{Top row:} on challenging images, handcrafted methods such as LSD~\cite{von2008lsd} suffer from noisy image gradients, while current learned methods like HAWP~\cite{hawp} were trained on wireframe images and generalize poorly. \textbf{Bottom row:} we combine deep learning to regress a line attraction field and a handcrafted detector to get both accurate and robust lines.}
    \label{fig:teaser}
\end{figure}

Line segments are ubiquitous in human-made environments and encode the underlying scene structure in a compact way. As such, line features have been used in multiple vision tasks: 3D reconstruction and Structure-from-Motion (SfM)~\cite{hofer_2017,micusik_2017,mateus_2022}, Simultaneous Localization and Mapping~\cite{gomez_2019,pumarola_2017,Lange_2019,fu2020plvins,zuo_2017}, visual localization~\cite{Gao2021PoseRW}, tracking~\cite{quan_2021}, vanishing point estimation~\cite{Tardif_2009}, etc. Thanks to their spatial extent and presence even in textureless areas, they offer a good complement to feature points~\cite{Gao2021PoseRW,gomez_2019,pumarola_2017}.

All these applications require a robust and accurate detector to extract line features from images. Traditionally, line segments are extracted from the image gradient using handcrafted heuristics, such as in the Line Segment Detector (LSD)~\cite{von2008lsd}. These methods are fast and very accurate since they rely on low-level details of the image. However, they can suffer from a lack of robustness in challenging conditions such as in low illumination, where the image gradient is noisy. They also miss global knowledge from the scene and will detect any set of pixels with the same gradient orientation, including uninteresting and noisy lines.

Recently, deep networks offer new possibilities to tackle these drawbacks. This resurgence of line detection methods was initiated by the deep wireframe methods aiming at inferring the line structure of indoor scenes~\cite{wireframe,lcnn,afm,hawp,deephough}. Since then, more generic deep line segment detectors have been proposed~\cite{huang2020tp,hao_2021,dai2021fully,gu2021realtime,teplyakov2022}, including joint line detectors and descriptors~\cite{Pautrat_Lin_2021_CVPR,Zhang_2021_ICCV,abdellali_2021}. 
These methods can, in theory, be trained on challenging images and, thus, gain robustness where classical methods fail. 
As they require a large receptive field to be able to handle the extent of line segments in an image, they can also encode some image context and can distinguish between noisy and relevant lines. 
On the other hand, most of these methods are \textit{fully} supervised and there exists currently only a single dataset with ground truth lines, the Wireframe dataset~\cite{wireframe}. 
Initially designed for wireframe parsing, this dataset is biased towards structural lines and is limited to indoor scenes. Therefore, it is not a suitable training set for generic line detectors, as illustrated in Figure~\ref{fig:teaser}.
Additionally, similarly as with feature points~\cite{sarlin21pixloc,lindenberger2021pixsfm}, current deep detectors are lacking accuracy and are still outperformed by handcrafted methods on easy images. 
The exact localization of line endpoints is often hard to obtain, as lines can be fragmented and suffer from partial occlusion. 
Many applications using lines consequently consider infinite lines and ignore the endpoints~\cite{micusik_2017}.

Based on this assessment, we propose in this work to keep the best of both worlds: use deep learning to process the image and discard unnecessary details, then use handcrafted methods to detect the line segments. We thus retain the benefits of deep learning, namely, to abstract the image and gain more robustness to illumination and noise, while at the same time retaining the accuracy of classical methods. We achieve this goal by following the tracks of two previous methods that used a dual representation of line segments with attraction fields~\cite{afm,hawp}. The latter are continuous representations that are well-suited for deep learning, and we show how to leverage them as input to the traditional line detectors. Contrary to these two previous methods, we do not rely on ground truth lines to train our line attraction field, but propose instead to bootstrap existing methods to create a high-quality pseudo ground truth. 
Thus, our network can be trained on any dataset and be specialized towards specific applications, which we show in our experiments.

We additionally propose a novel optimization procedure to refine the detected line segments. This refinement is based on the attraction field output by the proposed network, as well as on vanishing points, optimized together with the segments. 
Not only can this optimization be used to effectively improve the accuracy of our prediction, but it can also be applied to other deep line detectors.

\noindent
In summary, we propose the following contributions:
\begin{itemize}[topsep=2pt,leftmargin=*,itemsep=0em]
    \item We propose a method \textbf{bootstrapping current detectors to create ground truth line attraction fields} on any image.
    \item We introduce an optimization procedure that can simultaneously \textbf{refine line segments and vanishing points}. This optimization can be used as a stand-alone refinement to improve the accuracy of any existing deep line detector.
    \item We set a new record in several downstream tasks requiring line segments by combining the \textbf{robustness of deep learning approaches} with the \textbf{precision of handcrafted methods} in a single pipeline.
\end{itemize}

\section{Related Work}
%
\boldparagraph{Handcrafted Line Detectors.}
Detecting line segments in images is traditionally performed based on the image gradient. Early methods threshold the gradient magnitude to keep only strong edges and search for aligned sets of pixels sharing the same gradient angle. LSD~\cite{von2008lsd} grows line regions, fits a rectangle to the resulting set of pixels, and finally extract a line segment. EDLines~\cite{akinlar2011edlines} grows the line regions in one direction only, orthogonal to the image gradient. Several extensions of these methods have been proposed, such as the multi-scale version of LSD, MLSD~\cite{salaun_2016}, and ELSED~\cite{suarez2021elsed}, a faster version of EDLines which avoids breaking lines in case of small discontinuities. AG3Line~\cite{zhang2021} proposes to actively group the seed points and adds line geometry constraints. Another approach consists in detecting full lines with the Hough transform~\cite{hough1962} in a first step, then finding segments within these lines~\cite{Elder2020MCMLSDAP}. Since all these methods rely on low-level details of the image, they are highly accurate and fast, but lack robustness to noise and low illumination.

\boldparagraph{Learned Line Detectors.}
Deep line detection was first introduced through the task of wireframe parsing, i.e. estimating the structural lines of a scene~\cite{wireframe}. Several approaches have been proposed to parameterize and represent the line segments, e.g., with two endpoints~\cite{lcnn}, attraction fields~\cite{afm,hawp}, center and offset to the endpoints~\cite{huang2020tp}, graphs~\cite{zhang2019ppgnet,meng_2020}, and transformers~\cite{Xu_2021_CVPR}. 
Wireframes can be further improved through a Deep Hough transform~\cite{deephough}. 
All these methods are trained on a single dataset, the Wireframe dataset~\cite{wireframe}, and they are not necessarily suitable for other tasks such as visual localization and SfM.
\begin{figure*}[t]
    \centering
    \includegraphics[width=\textwidth]{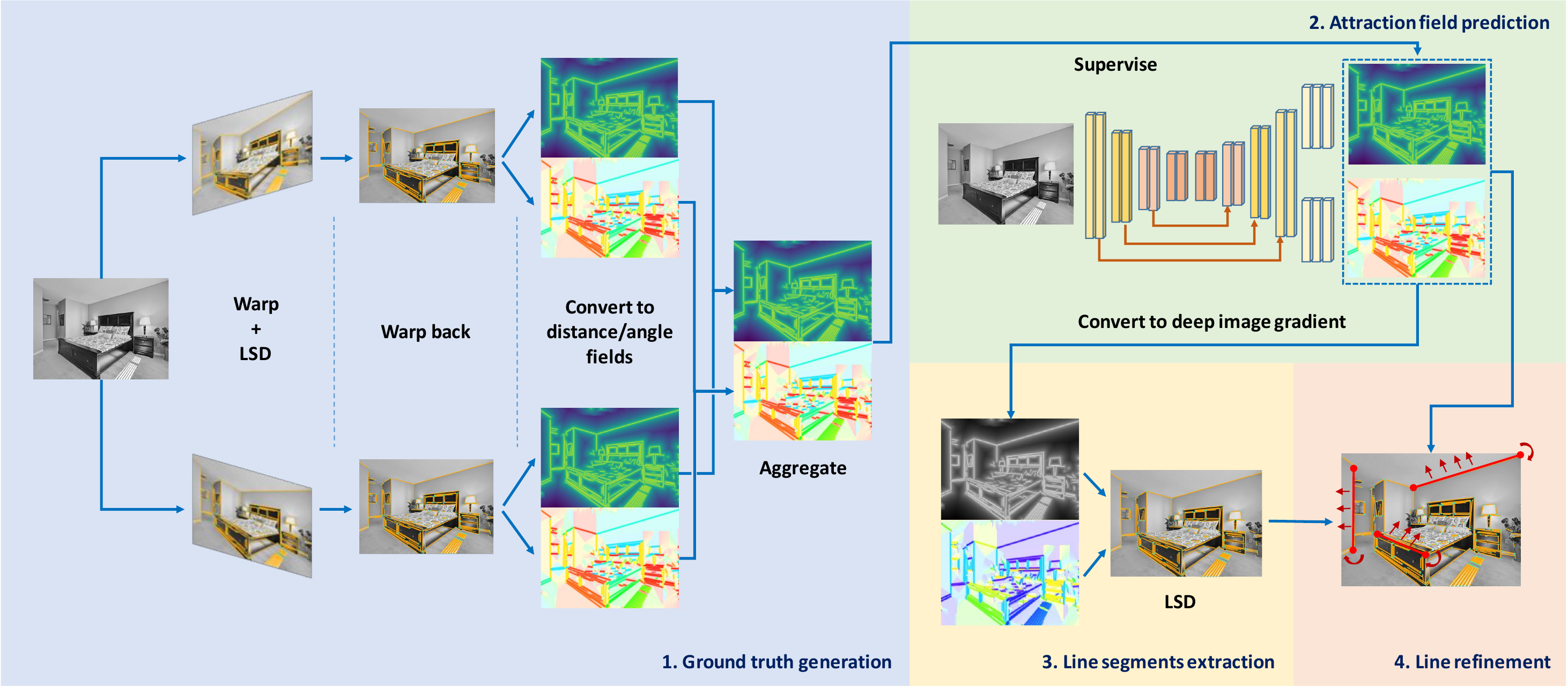}
    \caption{\textbf{Overview of the method.} (1) We generate ground truth line distance and angle fields (DF/AF) by bootstrapping LSD~\cite{von2008lsd}. (2) A deep network is trained to predict the DF/AF, which is then converted to a surrogate image gradient. (3) Line segments are extracted with LSD and (4) refined based on the DF/AF.}
    \label{fig:overview}
\end{figure*}
Generic deep line segment detectors have also been proposed, with a focus on efficiency~\cite{dai2021fully,gu2021realtime}, and can improve visual localization with points and lines~\cite{Gao2021PoseRW}.
However, these methods are again trained solely on the Wireframe dataset and their predicted lines are biased towards structural lines and indoor scenes.

Some works also perform a joint detection and description of line segments. 
SOLD2~\cite{Pautrat_Lin_2021_CVPR} introduced a self-supervised training, using the homography adaptation technique initially described in SuperPoint~\cite{superpoint}. ELSD~\cite{Zhang_2021_ICCV} and L2D2~\cite{abdellali_2021} both propose similar networks, but ELSD is again trained on the Wireframe dataset, while L2D2 uses a novel process to extract a line ground truth from LiDAR scans. 
Though these approaches are a first step towards unsupervised line detection, they still lack accuracy.

\boldparagraph{Attraction Fields.}
This work proposes to combine deep learning methods with classical line extractors. 
The key component for this is to use a dual representation of lines through an attraction field. 
This representation was first introduced by Xue et al.~\cite{afm} for the wireframe task, and later improved with HAWP~\cite{hawp,HAWP-journal}. 
They represent the set of discrete lines of an image with a continuous 2D vector field, suitable for deep networks.
We adopt a similar approach, with small modifications to make the prediction more accurate.
While not exactly an attraction field, Teplyakov et al.~\cite{teplyakov2022} also proposed to predict a line mask and line angle field with a network, then used LSD~\cite{von2008lsd} to get line segments. Our method obtains better accuracy by predicting a distance field instead of a simple binary mask.
Attraction fields have also been leveraged for keypoint detection~\cite{huang2021vs}, where 2D vectors are voting for the closest keypoint in the image. 
These detections-by-voting offer a convenient way to represent discrete quantities through continuous ones, and are also a key aspect of our approach when it comes to generating a reliable ground truth for line detection.

\section{Hybrid Line Detector}
%
We demonstrate how to combine the robustness of deep networks together with the accuracy of handcrafted line detectors.
We train a deep network to predict a line attraction field, convert it to a surrogate image gradient, and feed it to a handcrafted line detector to obtain the segments.
Finally, an optimization based on the attraction field is used to refine the lines, as depicted in Figure~\ref{fig:overview}.

\subsection{Line Attraction Field}
%
Representing line segments through an attraction field was first proposed by Xue et al.~\cite{afm}. 
They initially proposed to regress a 2D vector field for each pixel of an image, indicating the relative position of the closest point on a line. 
This approach allows to represent discrete quantities (the line segments) as a smooth 2-channel image well suited for deep learning. 
In~\cite{hawp}, the authors enriched the attraction field by adding two angles pointing at the endpoints of the closest line. 
Recovering the original segments from the attraction field is then straightforward.

\begin{figure}[tb]
    \centering
    \scriptsize
    \newcommand{\szp}{0.22}
    \setlength{\tabcolsep}{4pt}
    \begin{tabular}{cccc}
        \includegraphics[width=\szp\columnwidth]{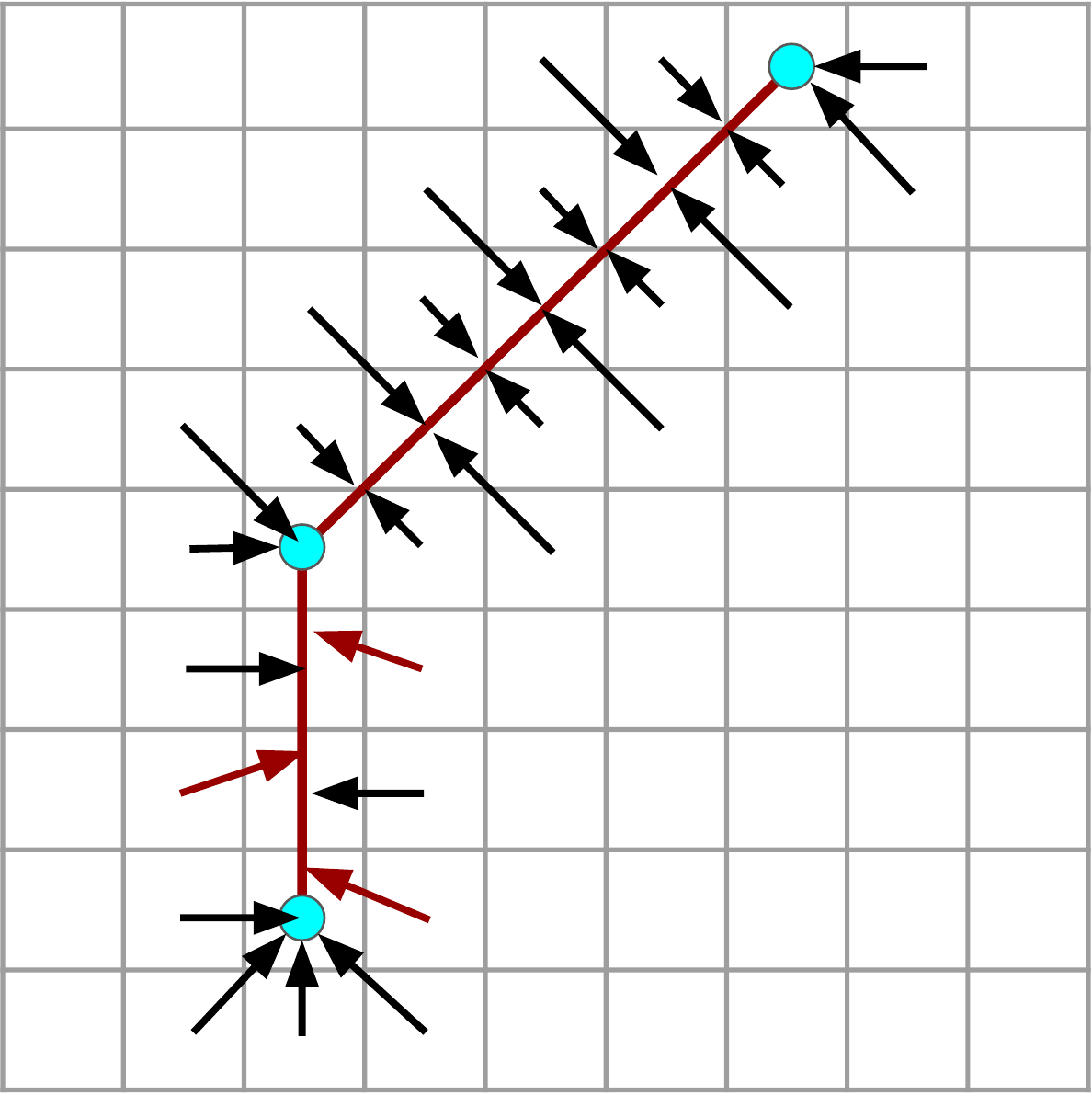} &
        \includegraphics[width=\szp\columnwidth]{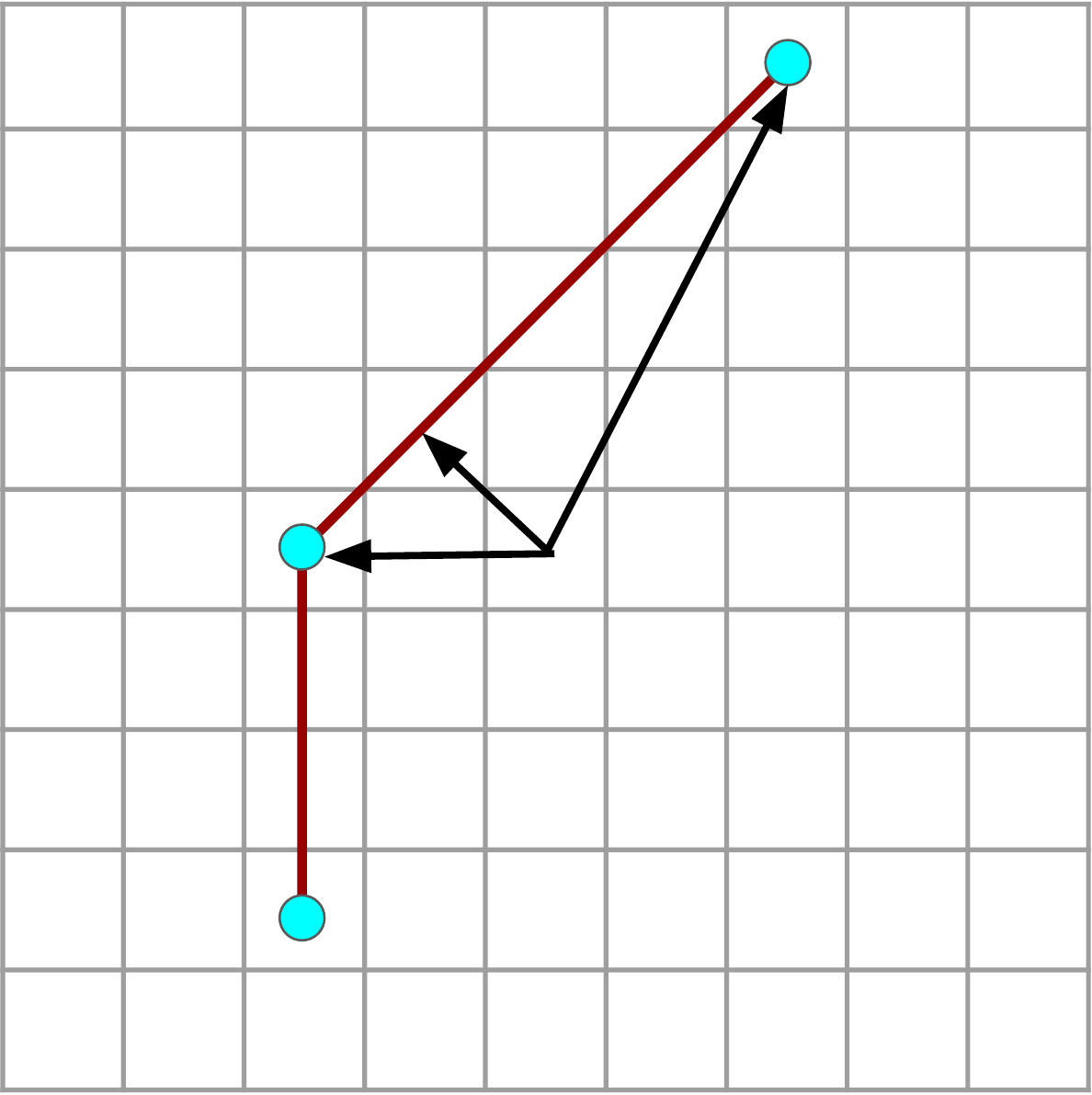} &
        \includegraphics[width=\szp\columnwidth]{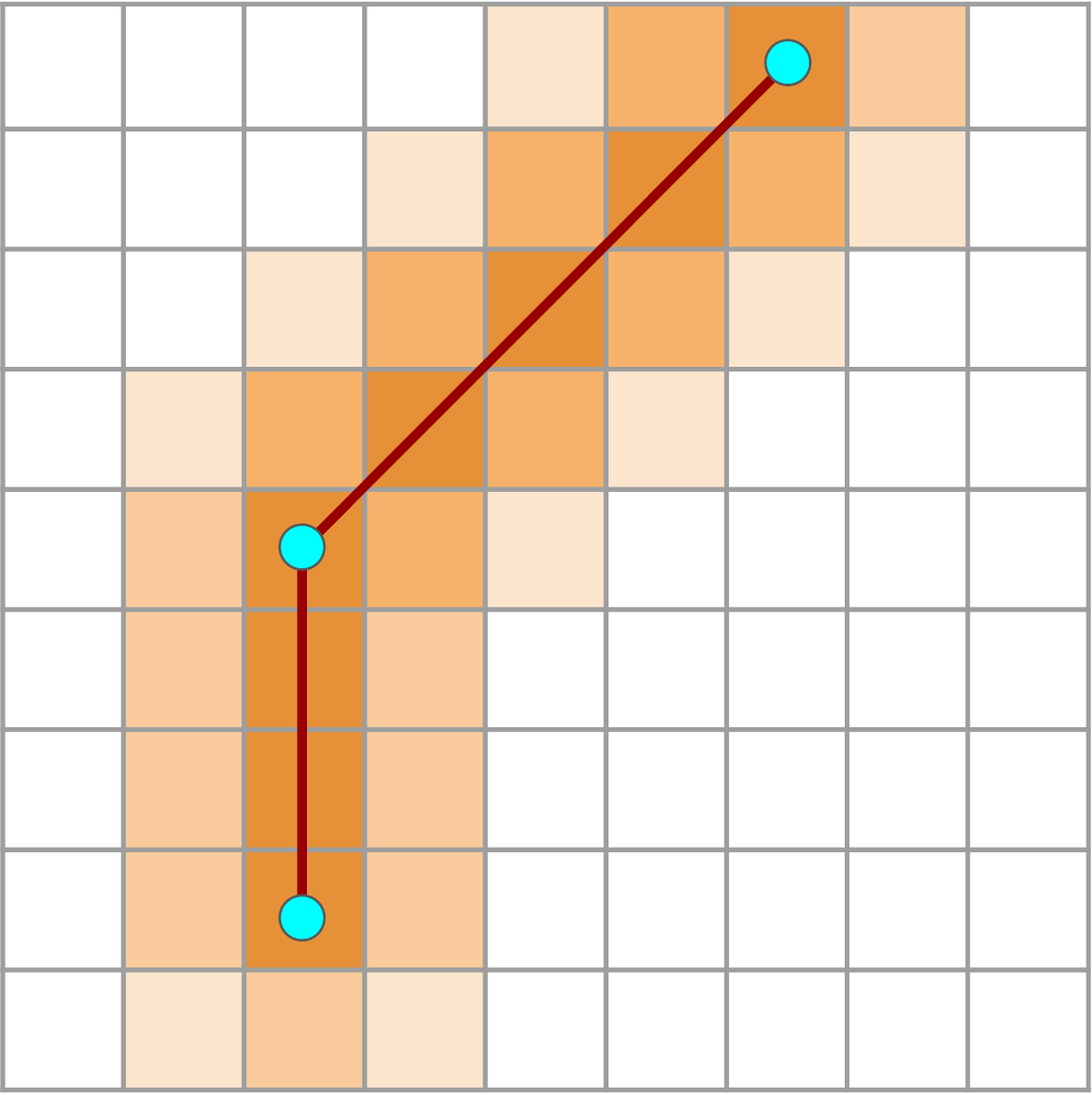} &
        \includegraphics[width=\szp\columnwidth]{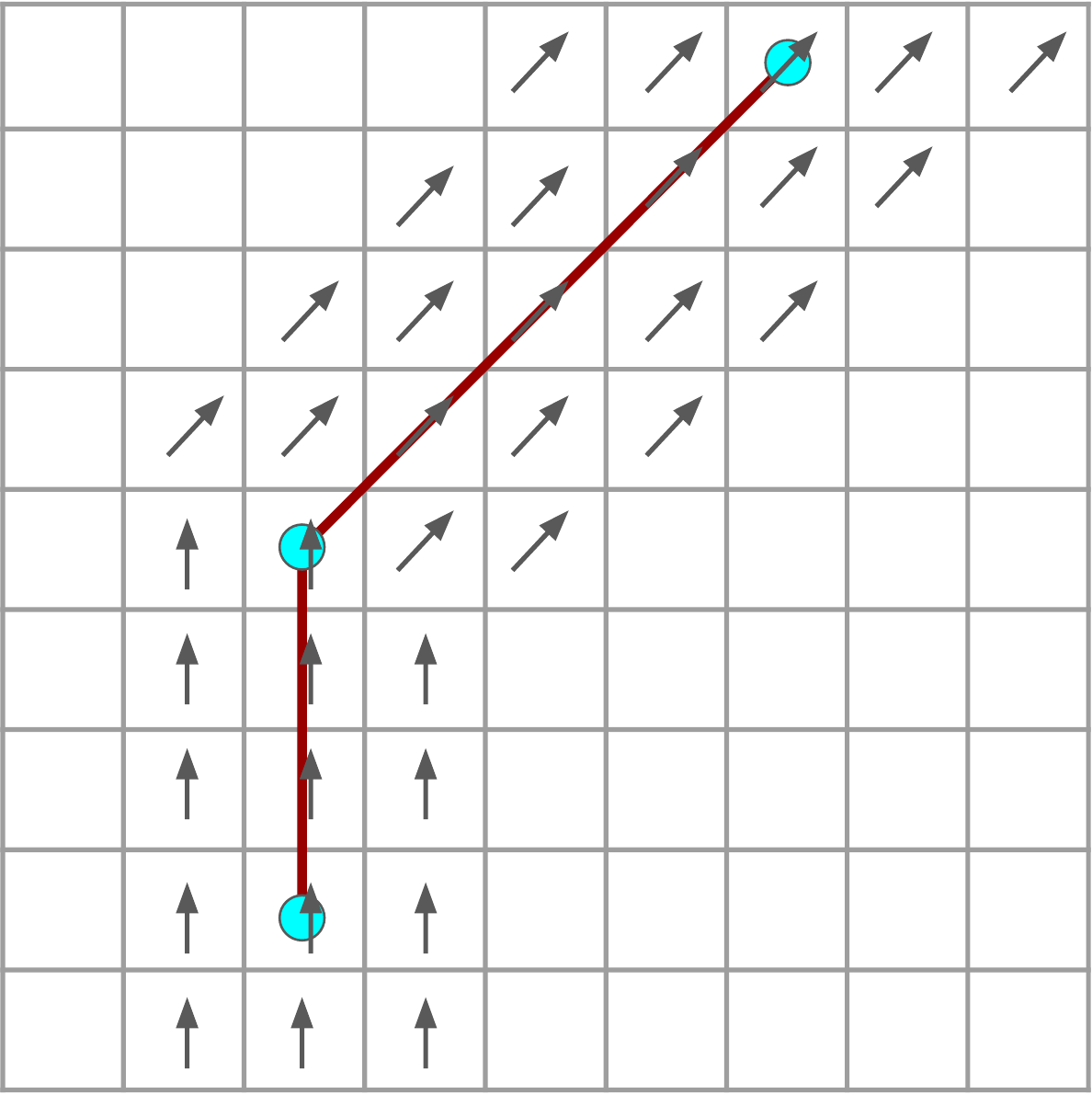} \\
        (a) AFM~\cite{afm} & (b) HAWP~\cite{hawp} & (c) Distance field & (d) Angle field \\[-5pt]
    \end{tabular}
    \caption{\textbf{Attraction field parametrizations.} (a) Parametrizing with 2D vectors may produce noisy angles for small vector norms. (b) Adding offsets to the endpoints requires long-range information and is not robust to noisy endpoints. We propose to decouple the distance field (c) and line orientation field (d).}
    \label{fig:parametrization}
\end{figure}

However, this representation is not optimal to obtain accurate line segments, as illustrated in Figure~\ref{fig:parametrization}. 
Directly predicting the position of the endpoints as done in HAWP~\cite{hawp} requires a larger receptive field to be able to get information from far-away endpoints, so that the network will focus on higher-level details instead of low-level ones. 
Additionally, deep networks are still struggling to yield accurate keypoint detections~\cite{sarlin21pixloc,lindenberger2021pixsfm}, which holds even more for line endpoints, which are notoriously noisy and unstable. 
On the contrary, handcrafted methods such as LSD~\cite{von2008lsd} are very low-level and gradually grow a line, so that endpoints are recovered only at the end of the region growing process. 
In this work, we propose to restrict our network to a smaller receptive field and to let the traditional heuristics determine the endpoints.

We adopt a similar attraction field representation as HAWP~\cite{hawp} but without the additional two angles pointing at endpoints, yielding only a \emph{line distance field} (DF) and a \emph{line angle field} (AF).
For every pixel in these two images, the line distance field $\mathbf{\mathcal{D}}$ gives the distance from the current pixel to the closest point on a line, and the line angle field $\mathbf{\mathcal{A}}$ returns the orientation of the closest line. 
These two quantities can be easily obtained from the 2D offset field $(\mathbf{x}, \mathbf{y})\in \mathbb{R}^{H \times W} \times \mathbb{R}^{H \times W}$ pointing at the closest point on a line, where $(H, W)$ are the dimension of the image:
\begin{equation}
    \label{eq:attr_field}
    \mathbf{\mathcal{D}} = \sqrt{\mathbf{x}^2 + \mathbf{y}^2}, \qquad \mathbf{\mathcal{A}} = \arctan \frac{\mathbf{y}}{\mathbf{x}} + \pi / 2 \mod{\pi} \ .
\end{equation}
We define here the line angle modulo $\pi$ so that a pixel above or below a line would have the same angle. 
Adopting this parametrization has the advantage of separating the norm from the angle of the 2D offset. 
Traditional detectors are leveraging the image gradient magnitude and angle, so we adopt a similar representation. 
Furthermore, both quantities are continuous close to line segments, and the line angle is even constant close enough to a line.

\subsection{Ground Truth Generation}
\begin{figure}[tb]
    \centering
    \scriptsize
    \newcommand{\s}{0.31}
    \setlength{\tabcolsep}{4pt}
    \begin{tabular}{cccl}
        \includegraphics[width=\s\columnwidth]{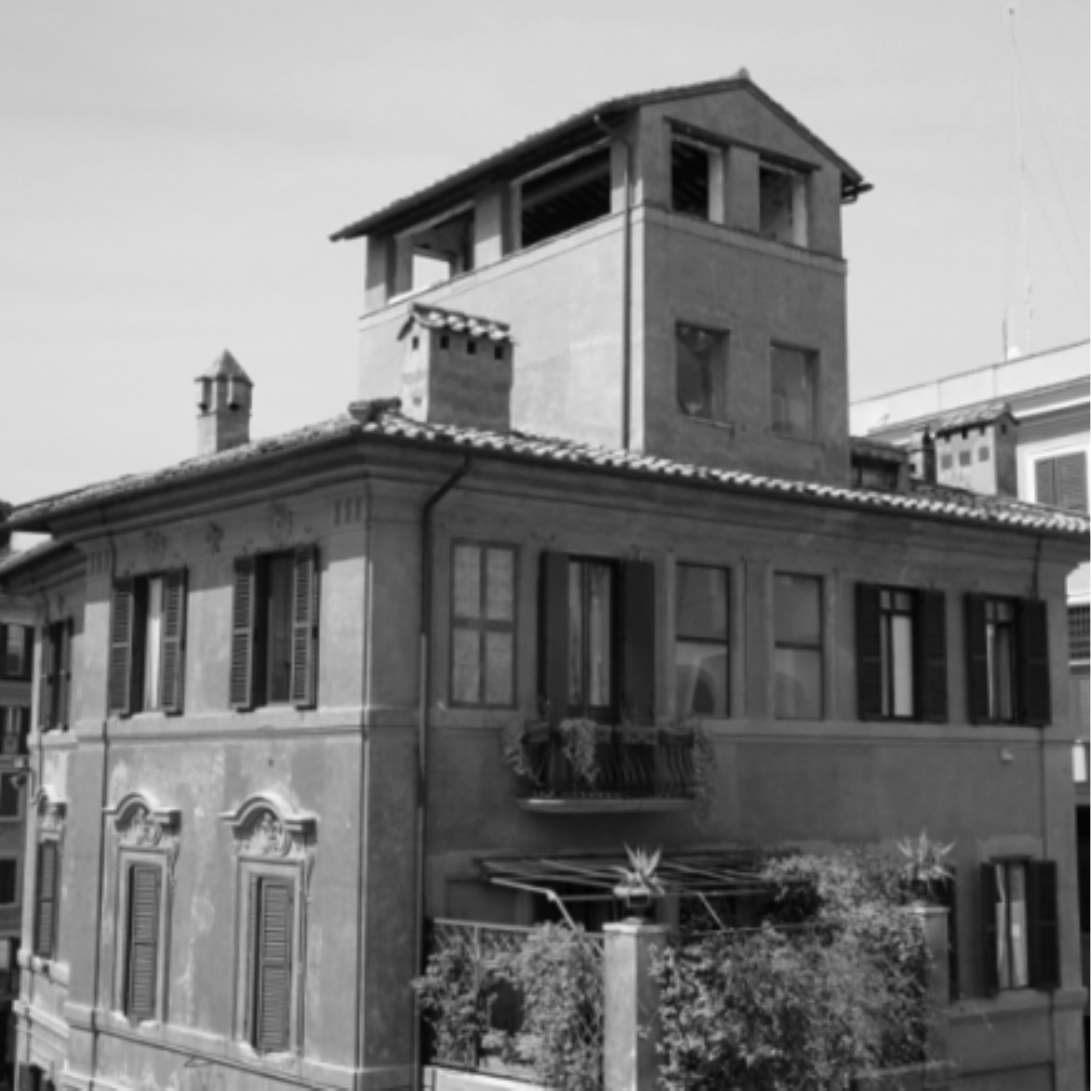} &
        \includegraphics[width=\s\columnwidth]{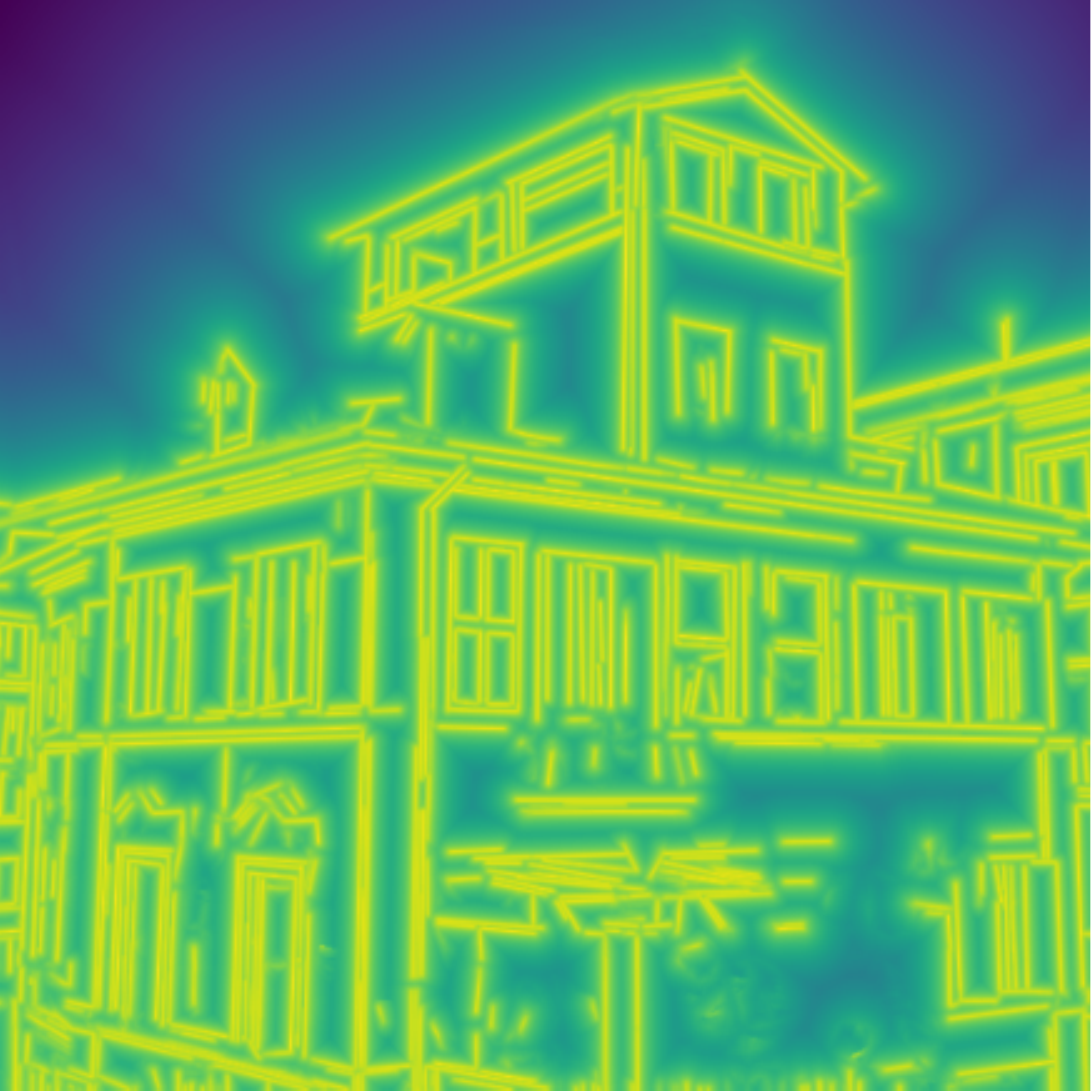} &
        \includegraphics[width=\s\columnwidth]{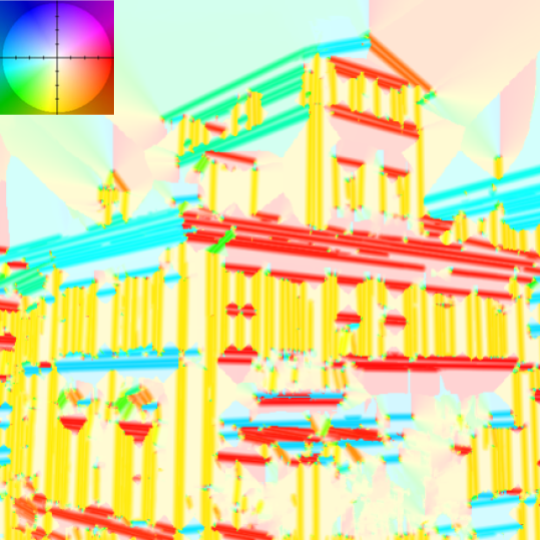} \\
        Input image & Distance field & Angle field \\[-5pt]
    \end{tabular}
    \caption{\textbf{Pseudo GT visualization.} Given an input image, we generate a line distance and angle fields (color coded~\cite{baker_2007}) and use them to supervise a deep network. Noisy lines, such as the ones in the bush at the bottom, are averaged out and ignored.}
    \label{fig:gt}
\end{figure}

%
To learn the attraction field, a ground truth is needed. 
Both AFM~\cite{afm} and HAWP~\cite{hawp} are supervised with the ground truth lines of the Wireframe dataset~\cite{wireframe}. 
We explore a novel method to acquire our ground truth, by bootstrapping previous line detectors.
Inspired by SuperPoint~\cite{superpoint} and SOLD2~\cite{Pautrat_Lin_2021_CVPR}, we propose to generate the ground truth attraction field through \emph{homography adaptation}. 
Given a single input image $I$, we warp it with $N$ random homographies $H_i$, detect line segments in all the warped images $I_i$ using any existing line detector, and then warp back the segments into $I$ to get a set $L_i$ of lines. 
We use LSD~\cite{von2008lsd} to extract lines as it is currently among the most accurate existing line detector.
The next step is to aggregate all the detections together, however, aggregating discrete quantities such as lines is non trivial. 
SOLD2~\cite{Pautrat_Lin_2021_CVPR} proposed to aggregate the endpoints and line heatmaps, and recover the segments afterwards.
Instead, we propose to convert the sets of lines $L_i$ into a distance field $\mathbf{\mathcal{D}}_i$ and angle field $\mathbf{\mathcal{A}}_i$, and to aggregate them by taking the median value of each pixel $(u, v)$ across all images:
\begin{equation}
    \left\{\begin{array}{l}
        \mathcal{D}(u, v) = \mathrm{median}_{i\in[1, N]} \mathcal{D}_i(u, v) \\
        \mathcal{A}(u, v) = \mathrm{median}_{i\in[1, N]} \mathcal{A}_i(u, v)
    \end{array}\right. \ .
\end{equation}
By taking the median, we remove the noisy lines that were detected in only a few images, as shown in Figure~\ref{fig:gt}.

\subsection{Learning the Line Attraction Field}
%
To regress our line distance and angle fields, we leverage a UNet-like neural network architecture~\cite{ronneberger_2015}. 
The input image of size $(H, W)$ is processed by several convolutional layers and gradually downsampled up to a factor of 8 through 3 successive average pooling operations. 
The features are then upscaled back to the original resolution through another series of convolutional layers and bilinear interpolation. 
The resulting deep features are then split into two branches, one outputting the distance field $\mathbf{\hat{\mathcal{D}}} \in \mathbb{R}^{H \times W}$ and the other one the angle field $\mathbf{\hat{\mathcal{A}}} \in \mathbb{R}^{H \times W}$. Refer to Figure~\ref{fig:overview} and supp. material for the detailed architecture.

While all convolutions are followed by ReLU~\cite{agarap2018deep} and Batch Normalization~\cite{ioffe_2015}, the last two outputs have different activations. The angle field is obtained through a sigmoid activation and is multiplied by $\pi$ to get an angle within $]0, \pi[$. 
Since the distance field can get very small values close to lines, where we also want the highest accuracy, we adopt a special normalization. 
The distance field branch ends with a ReLU activation and outputs a normalized distance field $\mathbf{\hat{\mathcal{D}}}_n \in (\mathbb{R}^+)^{H \times W}$. The final distance field is obtained through the following denormalization:
\begin{equation}
    \mathbf{\hat{\mathcal{D}}} = r \cdot e^{-\mathbf{\hat{\mathcal{D}}}_n} \ ,
\end{equation}
where $r$ is a parameter in pixels that defines a region around each line. Since handcrafted methods mainly need gradient information close to line segments, we supervise our network only on pixels at a distance of less than $r$ pixels from a line. 
By selecting a small value for $r$, large portions of the image may not have any supervision, including areas where the pseudo ground truth was not able to detect real lines, e.g. lines with small contrast. 
Enforcing these lines to be in the background during training, i.e. with high distance field, provides a detrimental training signal and decreases the recall of the prediction. 
On the contrary, with our loose supervision, these low contrast lines are not penalized during training and our trained model can detect them, thus yielding a more complete prediction than the ground truth.

We compute the training loss by comparing with a normalized version of the ground truth: $\mathbf{\mathcal{D}}_n = - \log\left(\frac{\mathbf{\mathcal{D}}}{r}\right)$.
Note that since we only supervise pixels with a distance field below $r$, $\frac{\mathbf{\mathcal{D}}}{r} \in [0, 1]$ and so $\mathbf{\mathcal{D}}_n \in \mathbb{R}^+$. 
We compute the total loss as the sum of the losses for the distance field and the angular field:
\begin{equation}
    \mathcal{L} = \mathcal{L}_{D} + \mathcal{L}_{A} \ ,
\end{equation}
where $\mathcal{L}_{D}$ is an L1 loss between the normalized distance fields and $\mathcal{L}_{A}$ is an L2 angular loss that takes the circularity of the angles into account for the given predicted and ground truth angle fields $\mathbf{\hat{\mathcal{A}}}, \mathbf{\mathcal{A}} \in [0, \pi]^{H \times W}$:
\begin{equation}
\begin{split}
    \mathcal{L}_{D} &= ||\mathbf{\hat{\mathcal{D}}}_n - \mathbf{\mathcal{D}}_n||_1 \ , \\
    \mathcal{L}_{A} &= \min (||\mathbf{\hat{\mathcal{A}}} - \mathbf{\mathcal{A}}||_2,
                            ||\pi - |\mathbf{\hat{\mathcal{A}}} - \mathbf{\mathcal{A}}| ||_2) \ .
\end{split}
\end{equation}

\subsection{Extracting Line Segments}
%
Since handcrafted detectors are based on the image gradient, we propose to convert our distance and angle fields into a surrogate image gradient magnitude $\mathbf{M}$ and angle $\mathbf{\theta}$:
\begin{equation}
    \left\{\begin{array}{ll}
        \mathbf{M}      & = r - \mathbf{\hat{\mathcal{D}}} \\
        \mathbf{\theta} & = \mathbf{\hat{\mathcal{A}}} - \frac{\pi}{2}
    \end{array}\right. \ .
\end{equation}

Our predicted angle follows the directions of the lines and is perpendicular to the image gradient, so we rotate it by $\frac{\pi}{2}$.
The maximal magnitude of a pixel on a line is $r$.

\begin{figure}[tb]
    \centering
    \scriptsize
    \setlength{\tabcolsep}{5pt}
    \begin{tabular}{cc}
        \multicolumn{2}{c}{\includegraphics[width=0.7\columnwidth]{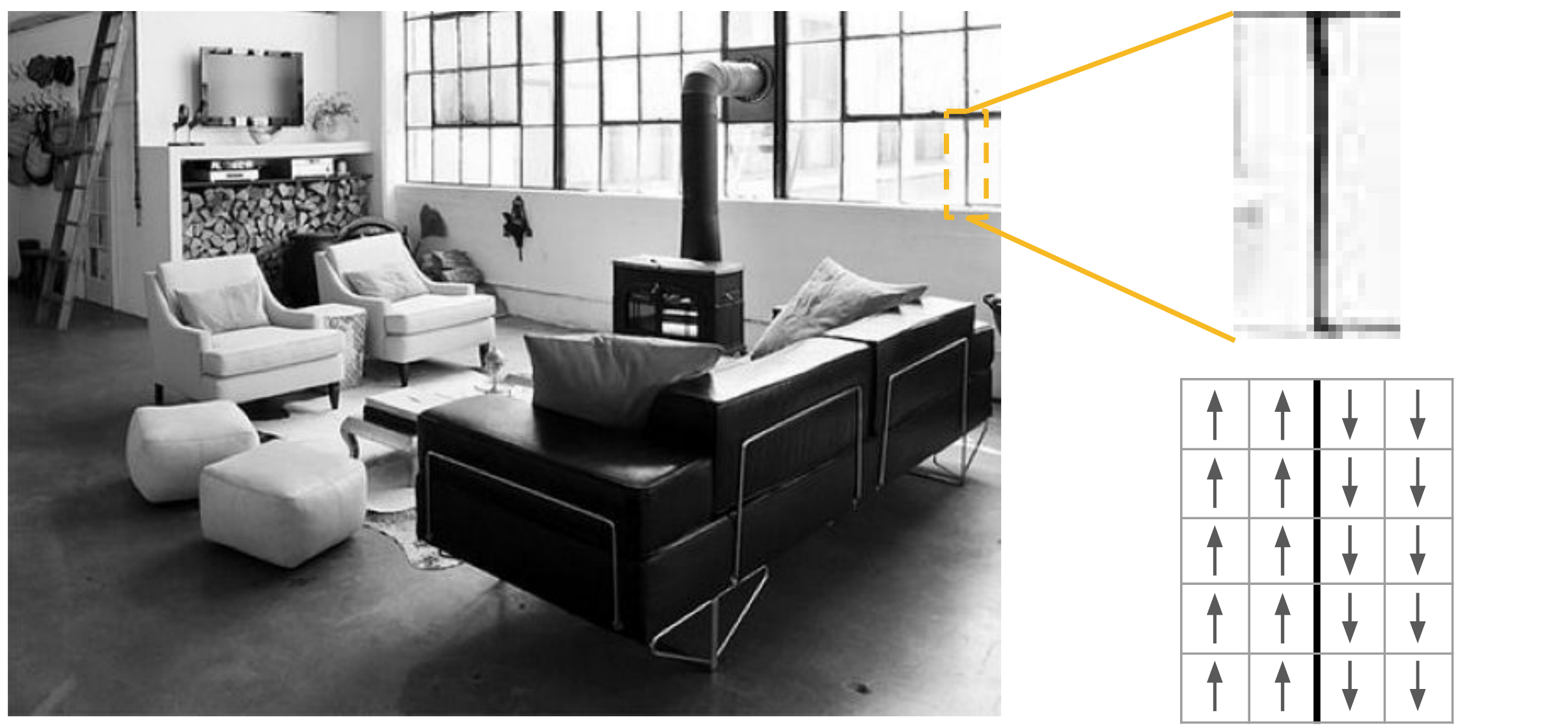}} \\
        \multicolumn{2}{c}{(a) A double edge} \vspace{0.2cm}\\
        \includegraphics[width=0.47\columnwidth]{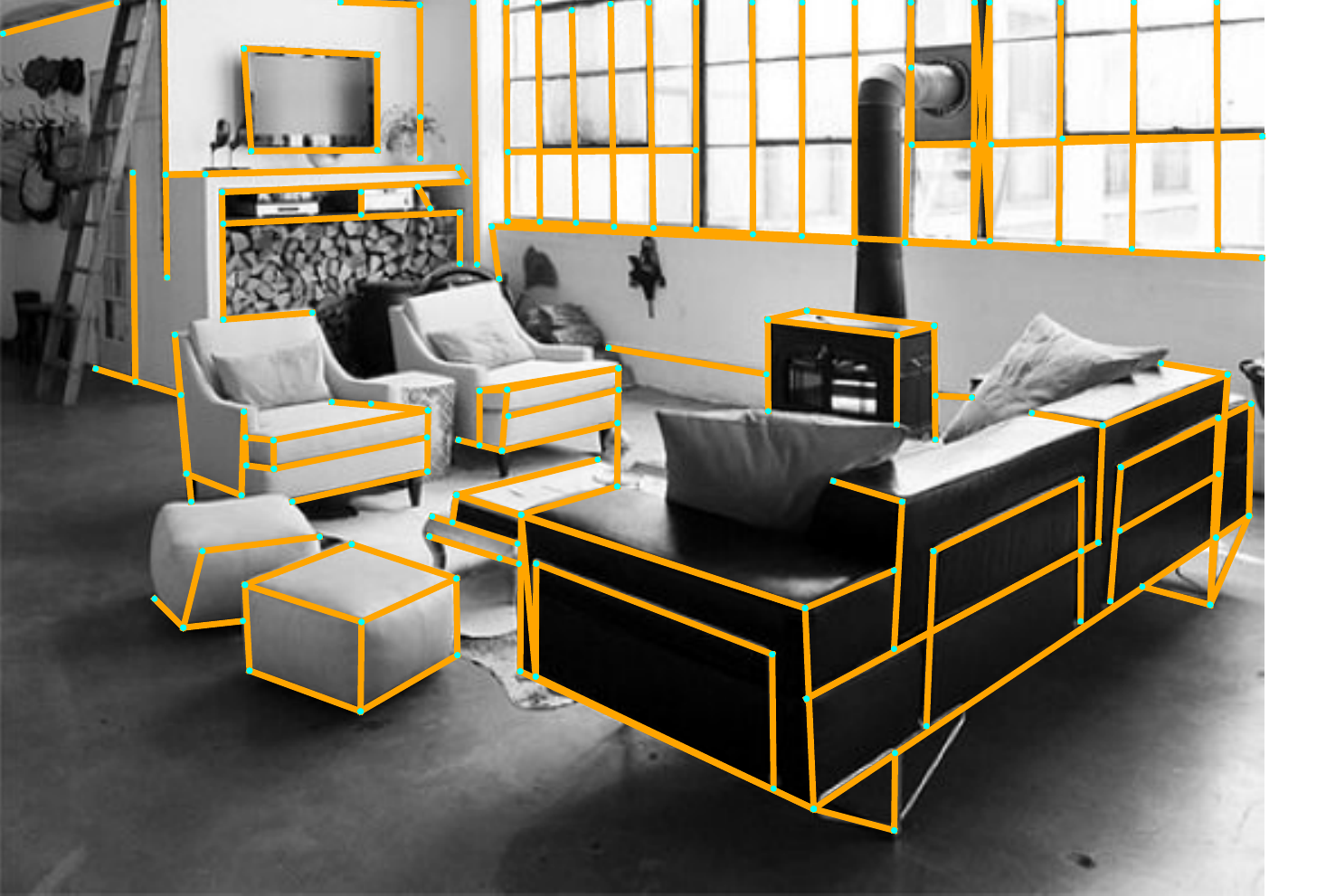} &
        \includegraphics[width=0.47\columnwidth]{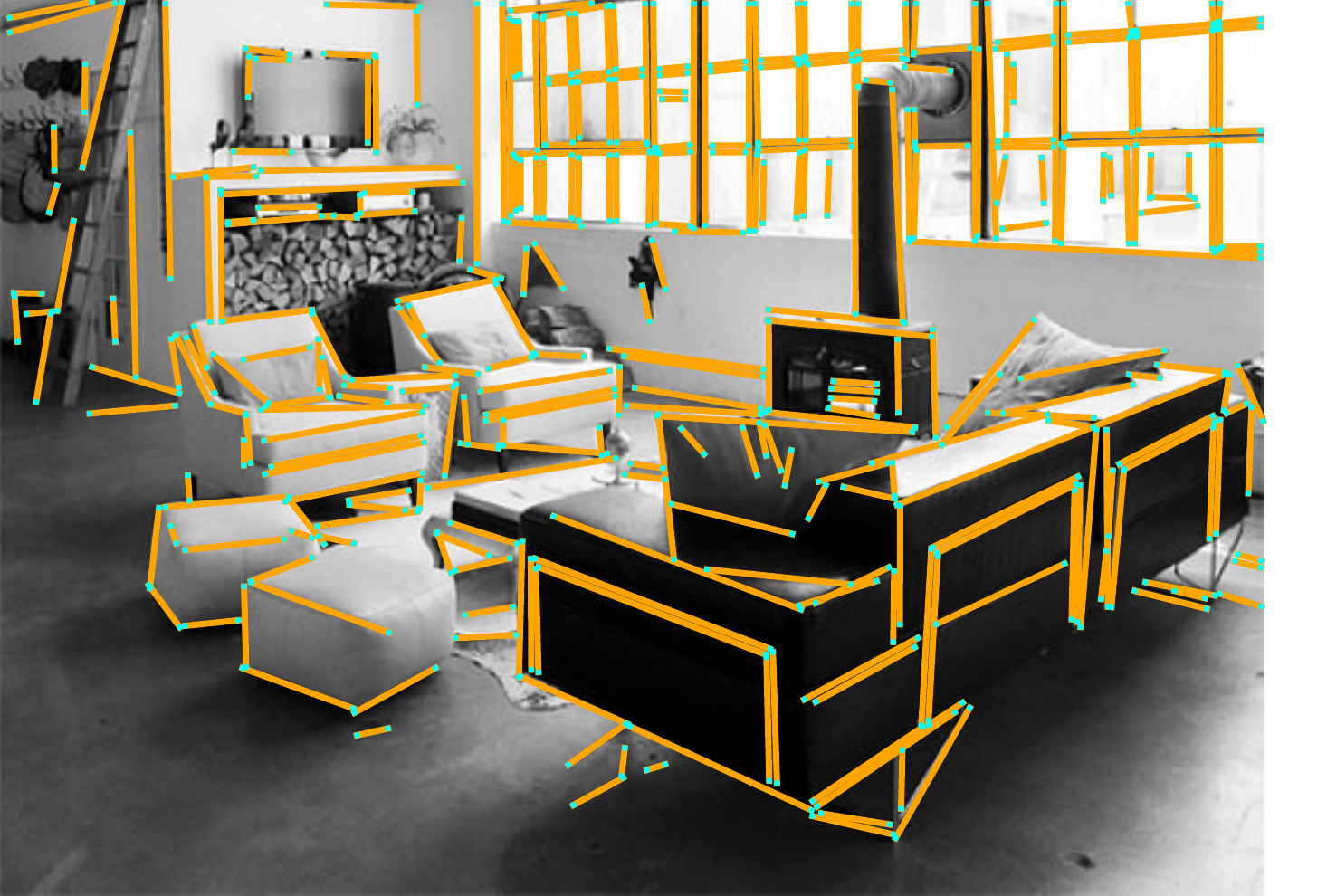} \\
        (b) HAWP~\cite{hawp} & (c) Ours \\[-5pt]
    \end{tabular}
    \caption{\textbf{Distinguishing double edges.} (a) An example of a bright-dark-bright edge and the oriented angle field. (b) Wireframe methods treat it as a single line. (c) We detect it as two lines for better accuracy.}
    \label{fig:double_lines}
\end{figure}

An important difference between the approaches of AFM and LSD is the gradient orientation.
For an edge separating a dark from a bright area, LSD keeps track of the dark-to-bright gradient direction, while AFM does not.
This becomes important when several parallel lines occur next to each other in a dark-bright-dark or bright-dark-bright pattern, as illustrated in Figure~\ref{fig:double_lines}.
For better accuracy and scale-invariance, we advocate to detect these double edges and make our predicted angle \emph{oriented}, based on the sign of the image gradient angle $\theta_I$:
\begin{equation}
    \arraycolsep=12pt
    \theta_o = \left\{\begin{array}{ll}
                      \theta & \mathrm{if} \  d(\theta, \theta_I) < d(\theta - \pi, \theta_I) \\
                      \theta - \pi & \mathrm{otherwise}
                \end{array}\right. \ ,
\end{equation}
where $d(\cdot,\cdot)$ is a circular distance between two angles.
Now equipped with an oriented angle $\mathbf{\theta}_o$ and magnitude $\mathbf{M}$, we can directly apply any existing classical line segment detector. Unless stated otherwise, we always use the LSD~\cite{von2008lsd} approach in the following, due to its high accuracy. 
In summary, the purpose of the deep net is to suppress image noise and detect low-contrast lines, while the line segments are accurately extracted by LSD afterwards.

We also add a filtering step, leveraging the DF and AF. We sample $n_f$ points along each line, and compute the fraction $p$ of samples whose distance function is below $\eta_{\text{DF}}$ and angle is close enough to the line orientation with tolerance $\eta_\theta$. Only the segments with enough inliers are kept.
\begin{figure*}[tb]
    \centering
    \scriptsize
    \setlength{\tabcolsep}{5pt}
    \begin{tabular}{ccccc}
        LSD~\cite{von2008lsd} & HAWP~\cite{hawp} & TP-LSD~\cite{huang2020tp} & SOLD2~\cite{Pautrat_Lin_2021_CVPR} & \OURS \ (Ours) \\
        \includegraphics[width=0.18\textwidth]{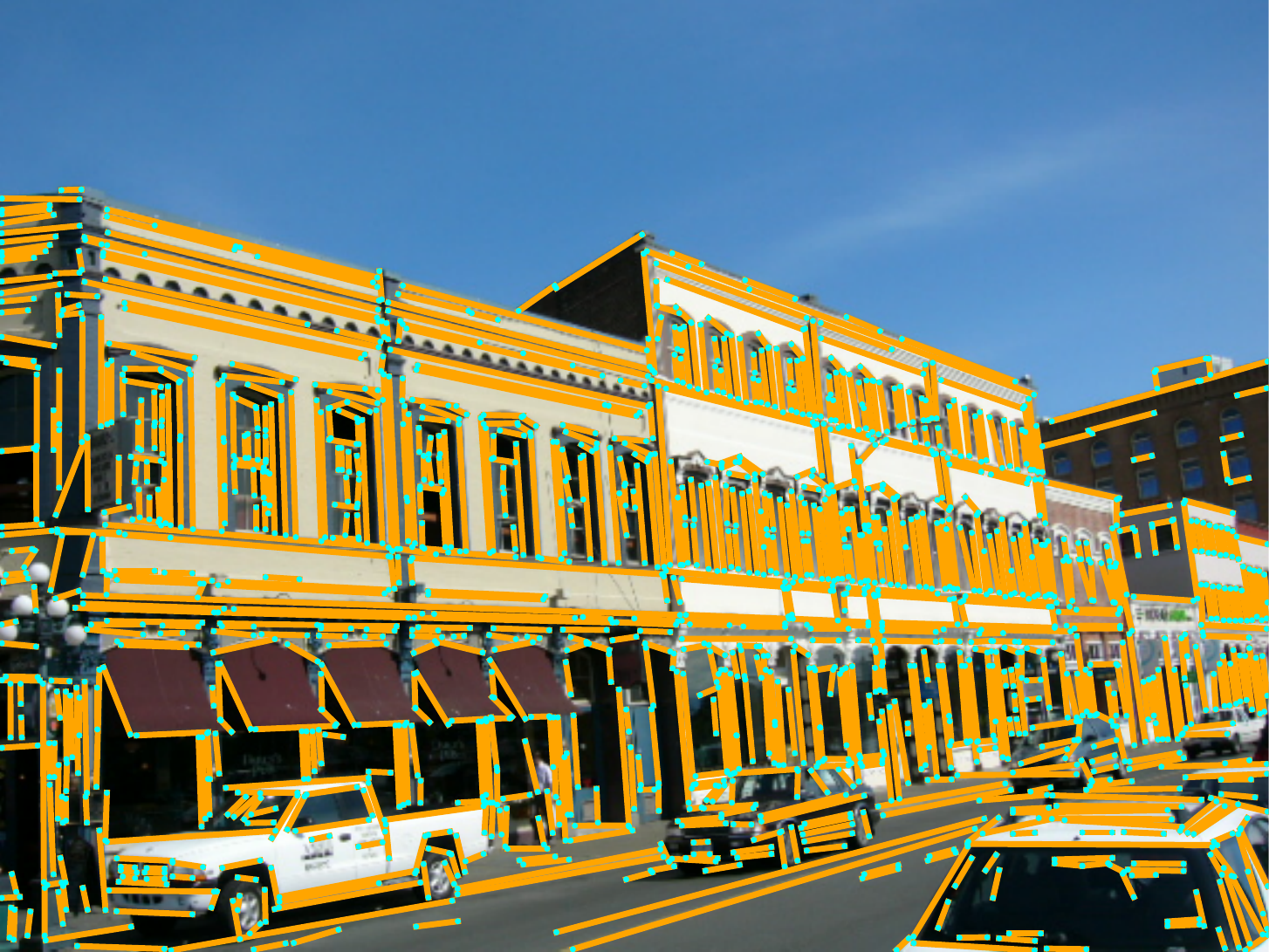} &
        \includegraphics[width=0.18\textwidth]{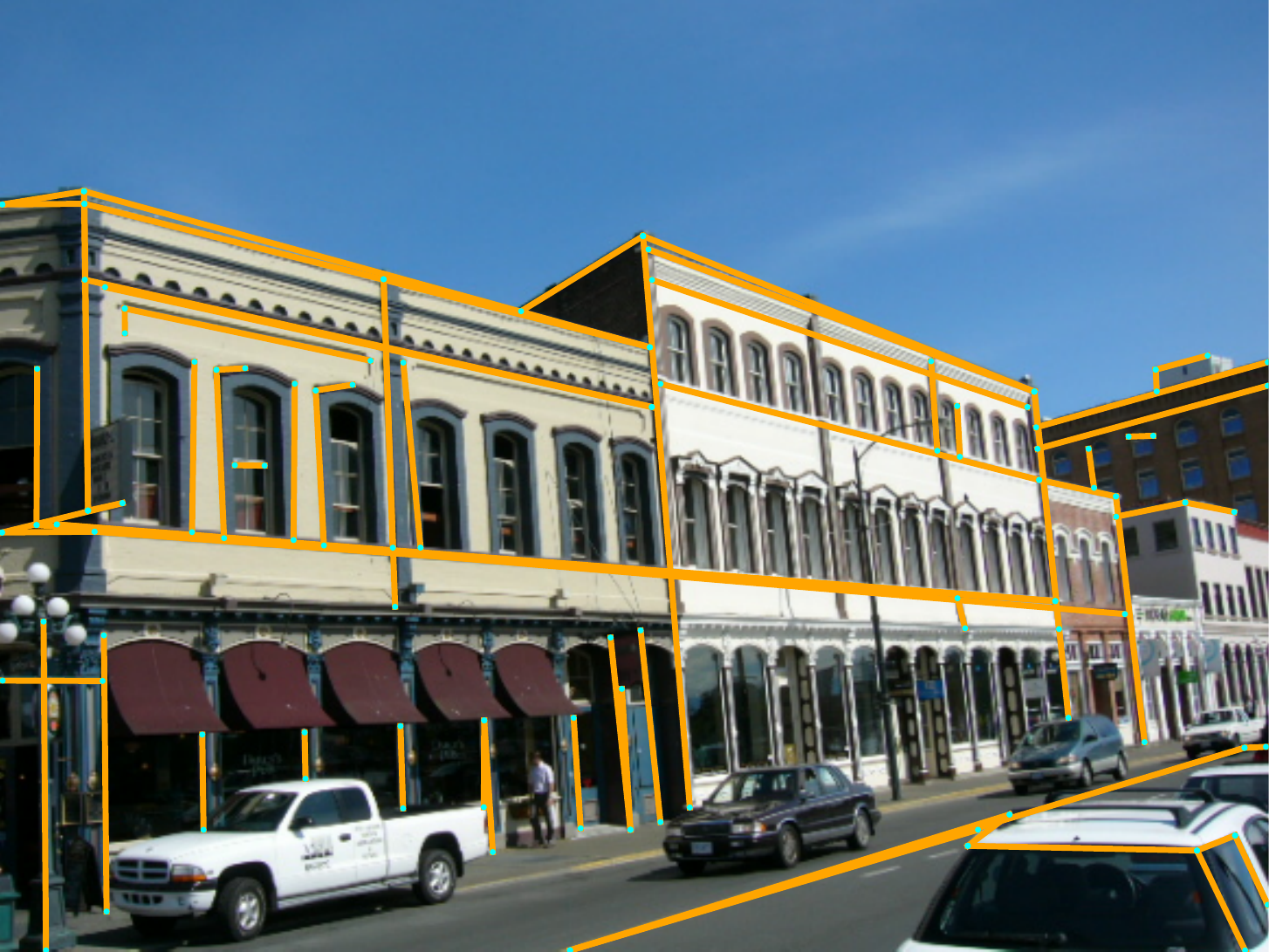} &
        \includegraphics[width=0.18\textwidth]{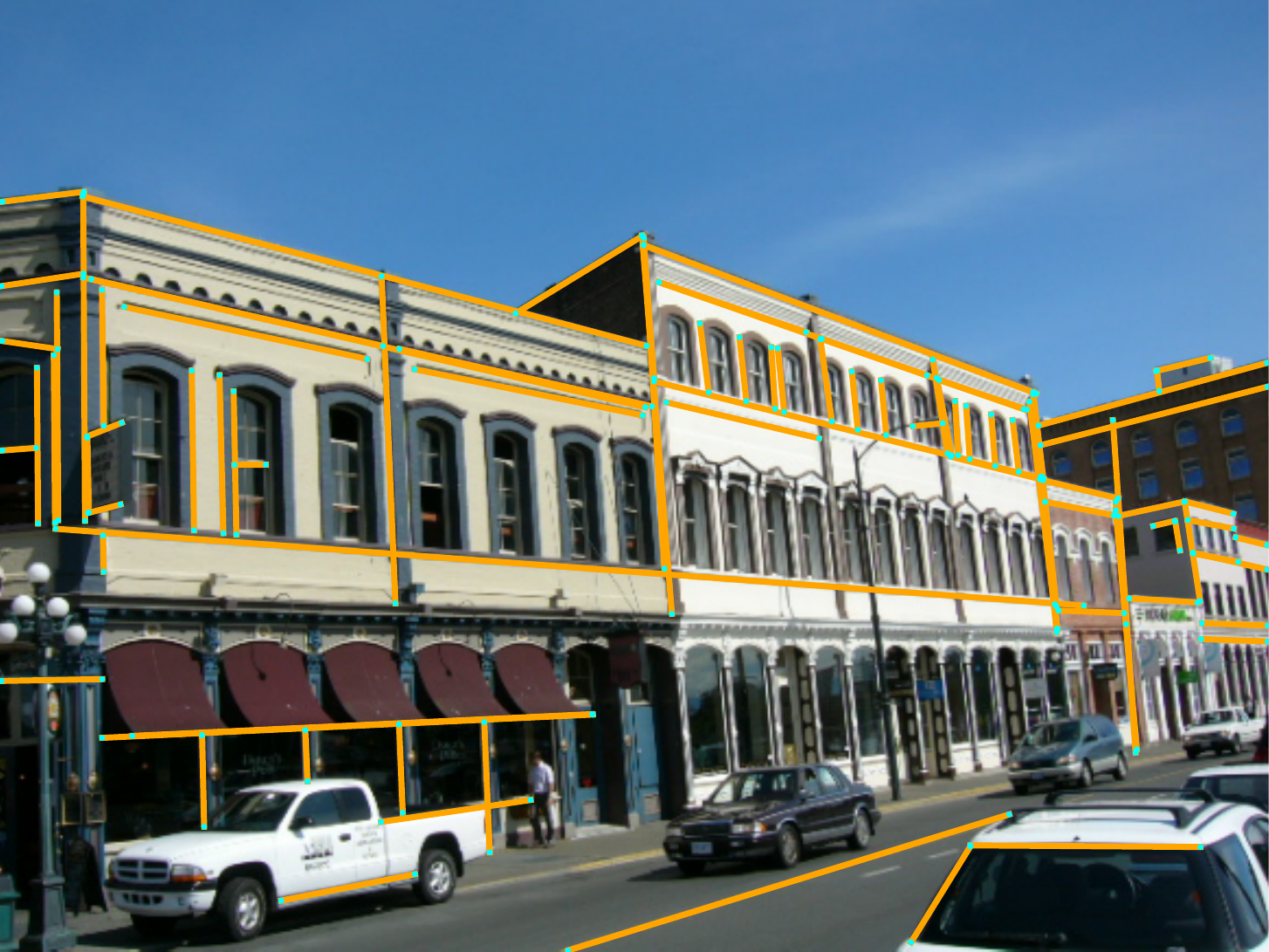} &
        \includegraphics[width=0.18\textwidth]{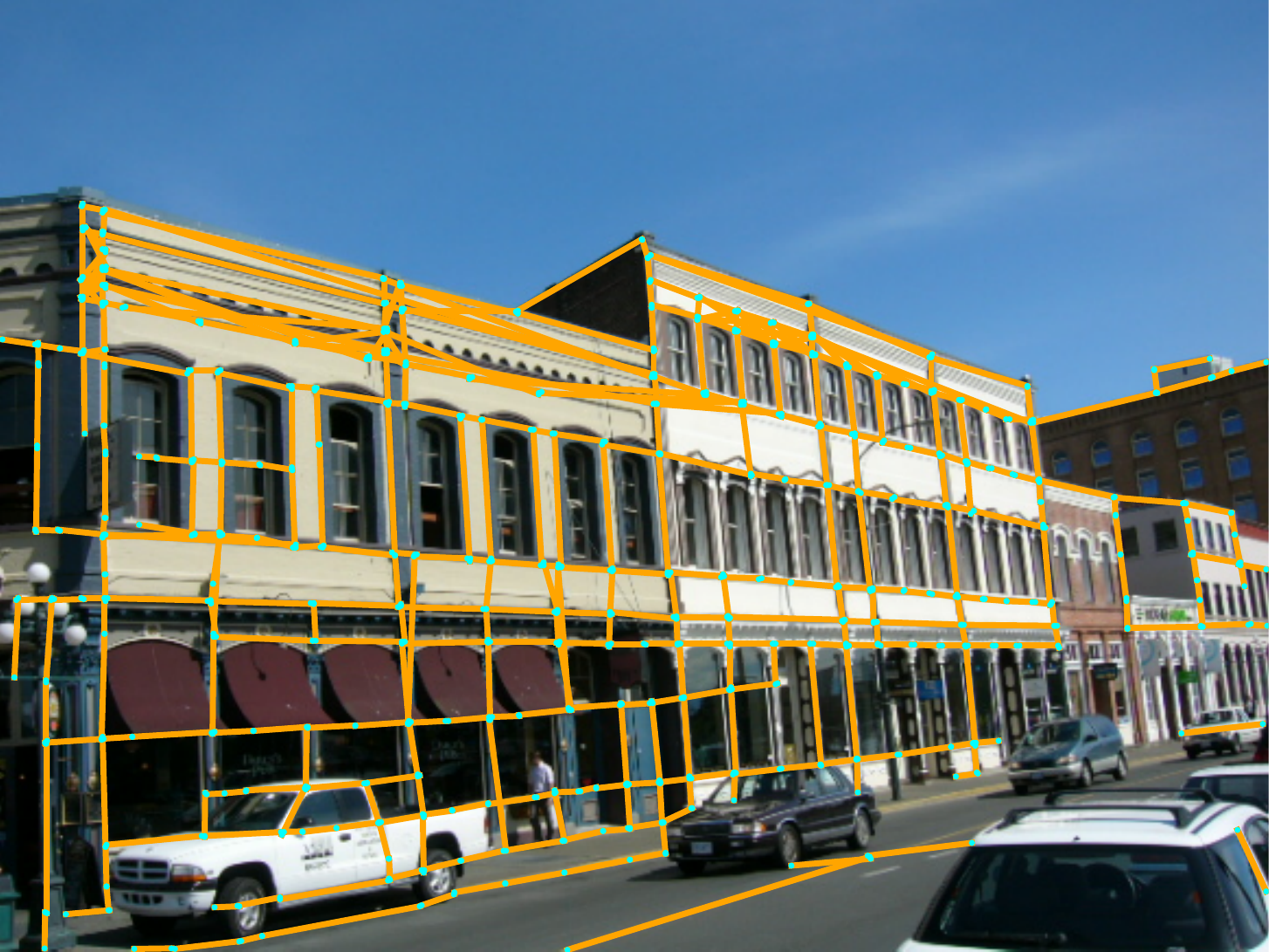} &
        \includegraphics[width=0.18\textwidth]{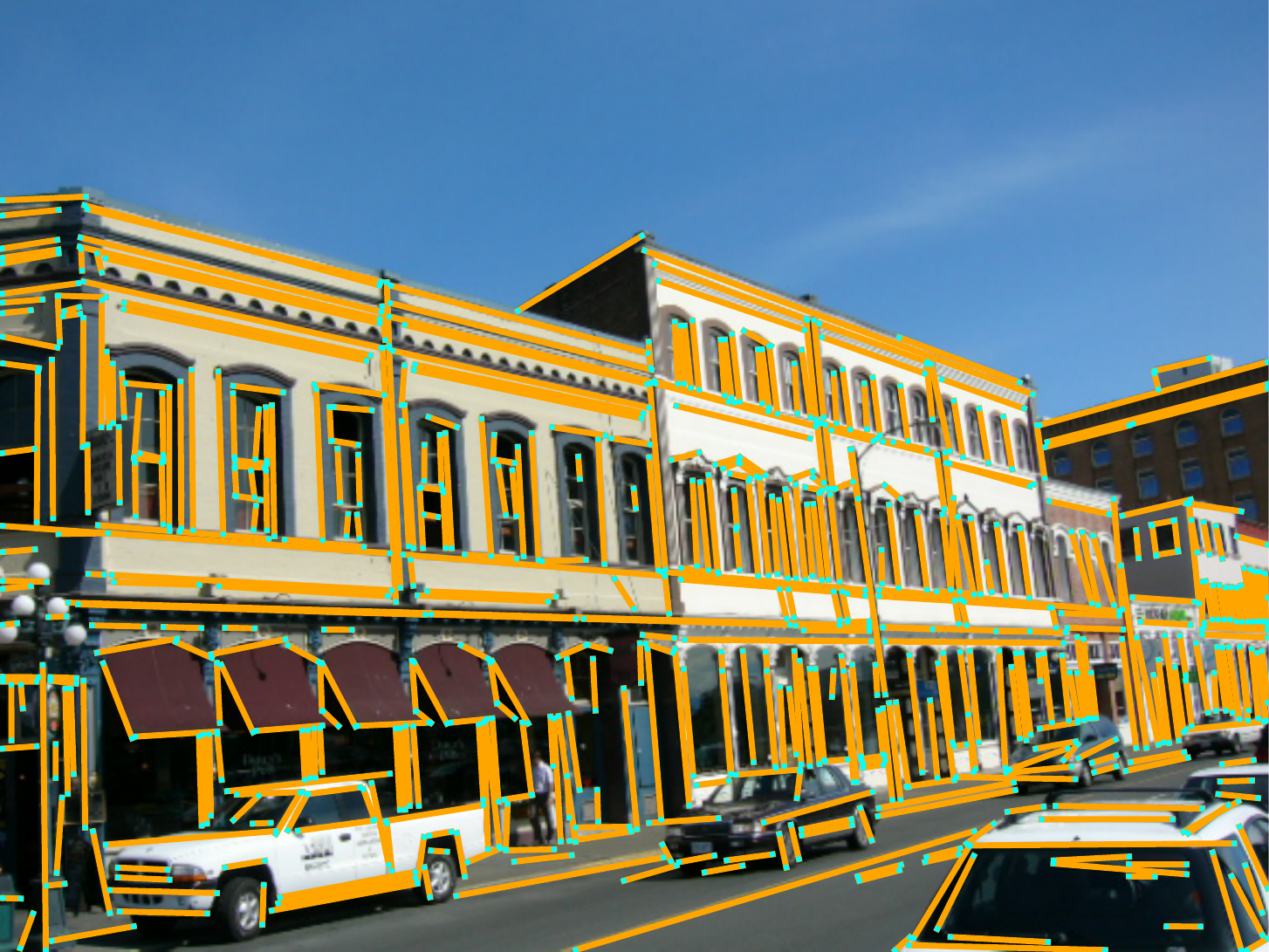}
    \end{tabular}
    \caption{\textbf{Line detection examples.} Wireframe methods~\cite{hawp,huang2020tp} only detect structural lines, while \OURS \ offers more generic detections.}
    \label{fig:visualizations}
\end{figure*}

\subsection{Line Segment Refinement with Optimization}
%
To make lines even more accurate, we propose an optimization step to refine them by leveraging the predicted DF and AF. This refinement can also be used to enhance the lines of any other detector, and we show in Section~\ref{sec:line_refinement} how it can make current deep detectors much more precise.

While lines are detected independently, they usually appear in highly structured configurations in the image. In particular, lines that are parallel in 3D will share vanishing points. We propose to integrate this as soft constraints into our refinement, effectively reducing the degrees of freedom.

We first compute a set of vanishing points (VPs) associated with the predicted line segments, using the multi-model fitting algorithm Progressive-X~\cite{Barath_2019_ICCV}.
We use a strict inlier threshold to be sure to associate only relevant lines to a VP. 
The optimization is then performed independently for each line and is a weighted unconstrained least square minimization of three different costs:
\begin{equation}
    \mathcal{C} = \lambda_{A} \mathcal{C}_{A} + \lambda_{D} \mathcal{C}_{D} + \lambda_{V} \mathcal{C}_{V} \ .
\end{equation}
Given a set $P$ of $n_{opt}$ points uniformly sampled along a line segment $l$, we denote each point by $p_i$, the orientation angle of the line as $\theta_l$, and the VP associated with the line as $\mathbf{v}_l$. We use the following three costs:
\begin{equation}
\begin{split}
    \mathcal{C}_{A} &= \frac{1}{n_{opt}} \sum_{p_i \in P}
        \Big(1 - \big(\cos (\hat{\mathcal{A}}(p_i) - \theta_l) \big) \Big) \ , \\
    \mathcal{C}_{D} &= \frac{1}{n_{opt}} \sum_{p_i \in P} \hat{\mathcal{D}}(p_i) \ , \qquad \qquad
    \mathcal{C}_{V} = d_{\text{VP}} (l, \mathbf{v}_l) \ ,
\end{split}
\end{equation}
where $d_{\text{VP}}$ is a distance measure between a line and a VP. We adopt the perpendicular distance of the line endpoints, projected onto the infinite line passing through the center of the line and the VP, as in~\cite{Tardif_2009}. These objectives are thus minimizing the difference between the sampled angle $\hat{\mathcal{A}}(p_i)$ and the line orientation angle, minimizing the average distance field value over the line, and minimizing the distance between the line and its VP. 
In case the closest VP is farther away from the line than a threshold $t_{\text{VP}}$, we drop the VP constraint as it would push the line towards a wrong VP.
To avoid lines drifting or collapsing to a single point, we keep the length of the line fixed, and we only optimize the lines over two degrees of freedom: the orientation angle of the line $\theta_l$, and a translation of the middle point in the perpendicular direction of the line.

Since the VPs are already computed, we can even optimize the VPs as well, as a by-product of our approach. 
Jointly optimizing lines and VPs empirically led to inferior results, mainly because some lines require more refinement than others, so that a global refinement performs worse than independently optimizing the lines.
We alternate, instead, between refining the lines and refining the VPs, for a fixed number of iterations $k$. 
The VP refinement is performed through a least square minimization of the distance $d_{\text{VP}}$ between the VP and all associated lines, and the line-VP association is recomputed after each iteration.

\subsection{Implementation Details}
%
We train two versions of our network, one indoors on the Wireframe dataset~\cite{wireframe}, but without using the ground truth lines, and one outdoors on MegaDepth~\cite{MegaDepthLi18}. 
Given the large size of MegaDepth, we keep 150 scenes for training and 17 for validation, and only sample 50 images from each scene. 
We use the Adam optimizer~\cite{kingma2014} and an initial learning rate of $1e^{-3}$, which is divided by 10 each time the validation loss reaches a plateau. 
The training takes roughly 12 hours on a single NVIDIA RTX 2080 GPU.
For the line detection, we set the line region $r$ to 5 pixels and ignore magnitudes in $\mathbf{M}$ below 3 when applying LSD. 
We use $n_f = 50$ samples in the filtering step, $\eta_{\text{DF}} = 1.5$, $\eta_\theta = \frac{\pi}{9}$ and accept lines with more than $50\%$ inliers.
The parameters for VP estimation are tuned for each method on a validation set, but the usual threshold $t_{VP}$ ranges from 1 to 2 pixels. 
The optimization weights are empirically chosen as $\lambda_D = 1$, $\lambda_A = 1$, and $\lambda_V = 0.2$. 
We adopt $n_{opt} = 10$ samples, perform a fixed set of $k=5$ alternating iterations, and optimize with Ceres~\cite{ceres}.

\section{Experiments}
%
To evaluate the performance of our method, we cannot use labeled lines as the existing ones are usually biased towards wireframes. 
We are more interested in evaluating the potential to use these lines for downstream applications, such as homography estimation, 3D line reconstruction, and visual localization. We also provide a visual comparison of various line detectors in Figure~\ref{fig:visualizations}.

\begin{table*}
    \centering
    \scriptsize
    \setlength{\tabcolsep}{1.9pt}
    \newcommand{\groupspace}{5pt}
    \begin{tabular}{llr@{\hskip10pt}cc@{\hskip\groupspace}cccc@{\hskip\groupspace}cc@{\hskip5pt}l@{\hskip5pt}cc@{\hskip\groupspace}cccc@{\hskip\groupspace}cc}
    \toprule
     & & & \multicolumn{2}{c}{Traditional} & \multicolumn{4}{c}{Learned} & \multicolumn{2}{c}{Hybrid\phantom{xxx}} & & \multicolumn{2}{c}{Traditional} & \multicolumn{4}{c}{Learned} & \multicolumn{2}{c}{Hybrid}\\ \cmidrule(r{\groupspace}){4-5} \cmidrule(r{\groupspace}){6-9} \cmidrule(r{\groupspace}){10-11} \cmidrule(r{\groupspace}){13-14} \cmidrule(r{\groupspace}){15-18} \cmidrule{19-20}
     & & & \multirow{2}{*}{\makecell{LSD\\\cite{von2008lsd}}} & \multirow{2}{*}{\makecell{ELSED\\\cite{suarez2021elsed}}} & \multirow{2}{*}{\makecell{HAWP\\\cite{hawp}}} & \multirow{2}{*}{\makecell{HAWPv3\\\cite{HAWP-journal}}} & \multirow{2}{*}{\makecell{TP-LSD\\\cite{huang2020tp}}} & \multirow{2}{*}{\makecell{SOLD2\\\cite{Pautrat_Lin_2021_CVPR}}} & \multirow{2}{*}{\makecell{LSDNet\\ \cite{teplyakov2022}}} & \multirow{2}{*}{\makecell{\OURS\\(Ours)}} & & \multirow{2}{*}{\makecell{LSD\\\cite{von2008lsd}}} & \multirow{2}{*}{\makecell{ELSED\\\cite{suarez2021elsed}}} & \multirow{2}{*}{\makecell{HAWP\\\cite{hawp}}} & \multirow{2}{*}{\makecell{HAWPv3\\\cite{HAWP-journal}}} & \multirow{2}{*}{\makecell{TP-LSD\\\cite{huang2020tp}}} & \multirow{2}{*}{\makecell{SOLD2\\\cite{Pautrat_Lin_2021_CVPR}}} & \multirow{2}{*}{\makecell{LSDNet\\ \cite{teplyakov2022}}} & \multirow{2}{*}{\makecell{\OURS\\(Ours)}} \\
     & & & & & & & & & & & & & & & & & & &  \\
     \midrule
     \multirow{7}{*}{\rotatebox{90}{HPatches~\cite{hpatches_2017_cvpr}\phantom{xxx}}} & \multirow{2}{*}{Struct} & Rep $\uparrow$ & 0.314 & 0.240 & 0.330 & 0.272 & \fst{0.413} & 0.308 & 0.108 & \snd{0.367} & \multirow{7}{*}{\rotatebox{90}{RDNIM~\cite{Pautrat_2020_ECCV}\phantom{xxx}}} & 0.283 & 0.209 & 0.284 & \snd{0.320} & \fst{0.344} & 0.307 & 0.047 & 0.285 \\
      &  & LE $\downarrow$ & \snd{1.309} & 1.551 & 2.019 & 2.132 & 1.500 & 1.741 & 2.860 & \fst{1.235} & & 2.039 & 2.303 & 2.206 & 1.939 & \snd{1.779} & 1.879 & 3.331 & \fst{1.733} \\ \cmidrule(r{\groupspace}){2-3}
      & \multirow{2}{*}{Orth} & Rep $\uparrow$ & \snd{0.468} & 0.465 & 0.337 & 0.309 & 0.444 & 0.395 & 0.200 & \fst{0.485} & & \fst{0.403} & 0.392 & 0.284 & 0.354 & 0.377 & 0.386 & 0.130 & \snd{0.394} \\
      &  & LE $\downarrow$ & \fst{0.793} & 0.845 & 1.905 & 1.937 & 1.305 & 1.362 & 2.285 & \snd{0.818} & & 1.369 & \snd{1.248} & 2.215 & 1.704 & 1.625 & 1.449 & 2.752 & \fst{1.098} \\ \cmidrule(r{\groupspace}){2-3}
      & \multicolumn{2}{l}{H estimation $\uparrow$} & \snd{0.697} & 0.617 & 0.260 & 0.231 & 0.388 & 0.421 & 0.316 & \fst{0.705} & & \snd{0.468} & 0.200 & 0.006 & 0.026 & 0.030 & 0.182 & 0.027 & \fst{0.591} \\ \cmidrule(r{\groupspace}){2-11} \cmidrule{13-20}
      & \multicolumn{2}{l}{\# lines / img} & 492.6 & 425.4 & 53.6 & 82.0 & 88.6 & 122.9 & 172.1 & 486.2 & & 191.4 & 112.0 & 31.6 & 23.8 & 24.1 & 138.2 & 109.1 & 400.0 \\
      & \multicolumn{2}{l}{Time [ms] $\downarrow$} & 104 & \fst{10} & 61 & 51 & 179 & 334 & \snd{48} & 271 & & \snd{34} & \fst{3} & 42 & 47 & 75 & 199 & 44 & 96 \\
     \bottomrule
     \end{tabular}
    \vspace{-0.2cm}
     \caption{\textbf{Line detection evaluation on the HPatches~\cite{hpatches_2017_cvpr} and RDNIM~\cite{Pautrat_2020_ECCV} datasets.} We compare repeatability (Rep) and localization error (LE) in structural and orthogonal distances, together with homography estimation. We get the best score on homography estimation and a good trade-off between classical and learned methods for the all metrics. The best score is in bold and the second best is underlined.}
     \label{tab:rdnim_hpatches}
\end{table*}

\subsection{Evaluation on Low-Level Metrics}
%
We first evaluate our line detection on two challenging datasets to test the robustness of the methods. 
First, the HPatches dataset~\cite{hpatches_2017_cvpr}, consisting of 580 pairs of images with ground truth homographies relating them and varying illumination and viewpoint changes. 
Second, the RDNIM dataset~\cite{Pautrat_2020_ECCV}, also with image pairs related by a homography and with challenging day-night variations. 
We use the night reference in our experiments to get more challenging pairs.

Similarly as in~\cite{Pautrat_Lin_2021_CVPR}, we assess the \emph{repeatability} and \emph{localization error} metrics. 
For both metrics, we compute a one-to-one matching of the detected line segments between the two images of a pair using the ground truth homography.
For each match, one can then compute the distance between the line in the reference image and the line of the warped image reprojected into the reference frame. 
We consider two line distance measures: the structural distance evaluating the average distance between the endpoints, and the orthogonal distance measuring the average distance of each endpoint of one line to their orthogonal projection to the other line. 
Repeatability (Rep) measures the ratio of lines whose match has an error below 3 pixels, and the localization error (LE) returns the average distance of the 50 most accurate matches.

We also compute a \emph{homography estimation score}, similarly as in~\cite{superpoint}. We first match line segments between the two images, using the Line Band Descriptor (LBD)~\cite{zhang2013lbd}. To estimate the homography, we sample minimal sets of 4 line matches and run LO-RANSAC~\cite{Lebeda2012loransac} for up to 1M iterations, using the orthogonal line distance as reprojection error.

We compare in Table~\ref{tab:rdnim_hpatches} our method to two classical detectors: LSD~\cite{von2008lsd} and ELSED~\cite{suarez2021elsed}; the best methods using attraction fields: HAWP~\cite{hawp}, its recent update HAWPv3 trained in a self-supervised way~\cite{HAWP-journal}, and LSDNet~\cite{teplyakov2022}: a similar approach as ours combining LSD and a deep network; and two generic deep line detectors: TP-LSD~\cite{huang2020tp} and SOLD2~\cite{Pautrat_Lin_2021_CVPR}. 
We use the implementation of the authors with the biggest model available and default parameters, except for HAWP where we use a threshold of 0.9, as it was not detecting enough lines otherwise. HAWPv3 was trained on ImageNet. 
For LSD, we use the implementation of Rafael Grompone\footnote{\url{http://www.ipol.im/pub/art/2012/gjmr-lsd/}} instead of the OpenCV one as it gets much better results.
Our method is given without the final optimization in the following, unless otherwise specified.

From the results, the learned methods, led by TP-LSD~\cite{huang2020tp}, offer good repeatability, but suffer from a low localization error and inaccurate homography estimation. 
Handcrafted methods and our method are much more accurate, due to the fact that they do not directly regress the endpoints, but gradually grow the line segments using very low-level details. 
\OURS~displays the best improvement over LSD when the changes become the most challenging, i.e. on RDNIM with strong day-night changes. 
It can significantly improve the localization error and homography estimation score. 
In spite of having a similar approach as ours, LSDNet~\cite{teplyakov2022} performs poorly for multiple reasons: they lose accuracy by rescaling images to a fixed low resolution, their line mask is less precise than our distance field, and their training is limited to the Wireframe dataset, while ours can be trained on more diverse images.
Overall, our method offers the best trade-off between handcrafted and learned methods and consistently ranks first in the downstream task of homography estimation.

\subsection{3D Line Reconstruction}
%
The aim of this work is to provide general-purpose lines and as such, the lines generated by DeepLSD should be suitable for 3D reconstruction. 
We leverage Line3D++~\cite{hofer_2017} that takes a collection of images with known poses and the associated 2D line segments, and outputs a 3D reconstruction of lines. 
We propose to compare our method with a few baselines on the first 4 scenes of the Hypersim dataset~\cite{roberts_2021}. 
This synthetic - but highly realistic - dataset has the advantage of offering a ground truth mesh and 3D model, making it suitable for a quantitative evaluation. 
Given the ground truth mesh of the scene, we can compute the recall and precision of the 3D lines.
Recall is the length in meters of all the portions of lines that are within 5 millimeters from the mesh. High values mean that many lines have been reconstructed.
Precision is the percentage of predicted lines that are within 5 millimeters from the mesh. High values indicate that most of the predicted lines are on a real 3D surface.

The results can be seen in Table~\ref{tab:3d_reconstruction}. DeepLSD obtains the best recall overall, and second best precision. While TP-LSD~\cite{huang2020tp} ranks first in precision, it is able to recover very few lines, as shows its average recall, which is 71\% smaller than the one of DeepLSD.
We provide qualitative examples of the reconstructions in the supp. material.
Note that DeepLSD is able to reconstruct more lines and with a higher precision than LSD~\cite{von2008lsd}, the detector that is the most commonly used for line reconstruction~\cite{hofer_2017}.

\begin{table}
    \centering
    \scriptsize
    \setlength{\tabcolsep}{2.7pt}
    \begin{tabular}{lcccccccccc}
    \toprule
     & \multicolumn{2}{c}{ai\_001\_001} & \multicolumn{2}{c}{ai\_001\_002} & \multicolumn{2}{c}{ai\_001\_003}
     & \multicolumn{2}{c}{ai\_001\_004} & \multicolumn{2}{c}{Average} \\
     \cmidrule(lr){2-3} \cmidrule(lr){4-5} \cmidrule(lr){6-7} \cmidrule(lr){8-9} \cmidrule(lr){10-11}
     & R & P & R & P & R & P & R & P & R & P \\
     \midrule
    LSD~\cite{von2008lsd} & 183.6 & 95.8 & 61.8 & 95.3 & \textbf{385.0} & 88.9 & 225.3 & 91.5 & 213.9 & 92.9 \\
    SOLD2~\cite{Pautrat_Lin_2021_CVPR} & 109.9 & 94.7 & 89.3 & 92.8 & 62.0 & 89.0 & 58.6 & 89.1 & 80.0 & 91.4 \\
    HAWPv3~\cite{HAWP-journal} & 15.8 & 79.9 & 15.6 & 81.0 & 24.4 & 68.4 & 18.5 & 77.3 & 18.6 & 76.7 \\
    TP-LSD~\cite{huang2020tp} & 68.8 & 95.3 & 38.9 & 94.7 & 50.7 & \textbf{98.2} & 102.7 & \textbf{94.3} & 65.3 & \textbf{95.6} \\
    DeepLSD & \textbf{204.8} & \textbf{96.5} & \textbf{89.5} & \textbf{98.1} & 378.8 & 88.0 & \textbf{231.1} & 91.9 & \textbf{226.1} & 93.6 \\
    \bottomrule
    \end{tabular}
    \vspace{-0.2cm}
     \caption{\textbf{Line 3D reconstruction evaluation.} We reconstruct lines in 3D with Line3D++~\cite{hofer_2017} and evaluate the line length recall in m (R $\uparrow$) and precision (P $\uparrow$) on the first 4 scenes of Hypersim~\cite{roberts_2021}.}
    \label{tab:3d_reconstruction}
\end{table}

\subsection{Visual Localization}
%
The 7Scenes dataset~\cite{7scenes} is a well-known RGB-D dataset for visual localization, displaying 7 indoor scenes with GT poses and depth. While most scenes are already saturated for point-based localization, the Stairs scene remains very challenging for feature points. Due to the lack of texture and repeated patterns of the stairs, current point-based methods are still struggling on this scene~\cite{dsacstar}. We thus propose to evaluate our method and previous works on this particular scene, by following the pipeline of hloc~\cite{sarlin2019coarse,hloc}, enriched with line features. As points remain important features, we still detect SuperPoint features~\cite{superpoint} and match them with SuperGlue~\cite{sarlin_2020_superglue}. We detect lines with different detectors, and match them between database and query images with the SOLD2 descriptor~\cite{Pautrat_Lin_2021_CVPR}. Since depth is available on 7Scenes, we can directly back-project lines in 3D and do not rely on line mapping. In practice, we sample points along each line, un-project them to 3D, and re-fit a line in 3D to these un-projected points. We use the solvers of \cite{kukelova2016efficient,Zhou2018ASA,PoseLib} to generate poses from a minimal set of 3 features (3 points, 2 points and 1 line, 1 point and 2 lines, or 3 lines), then combine them in a hybrid RANSAC implementation~\cite{Sattler2019Github,Camposeco2018CPVR} to robustly recover the query camera poses. We report the median translation and rotation error, as well as the percentage of successfully recovered poses under various thresholds.

\begin{figure}
    \centering
    \begin{minipage}{0.53\columnwidth}
        \includegraphics[width=\textwidth]{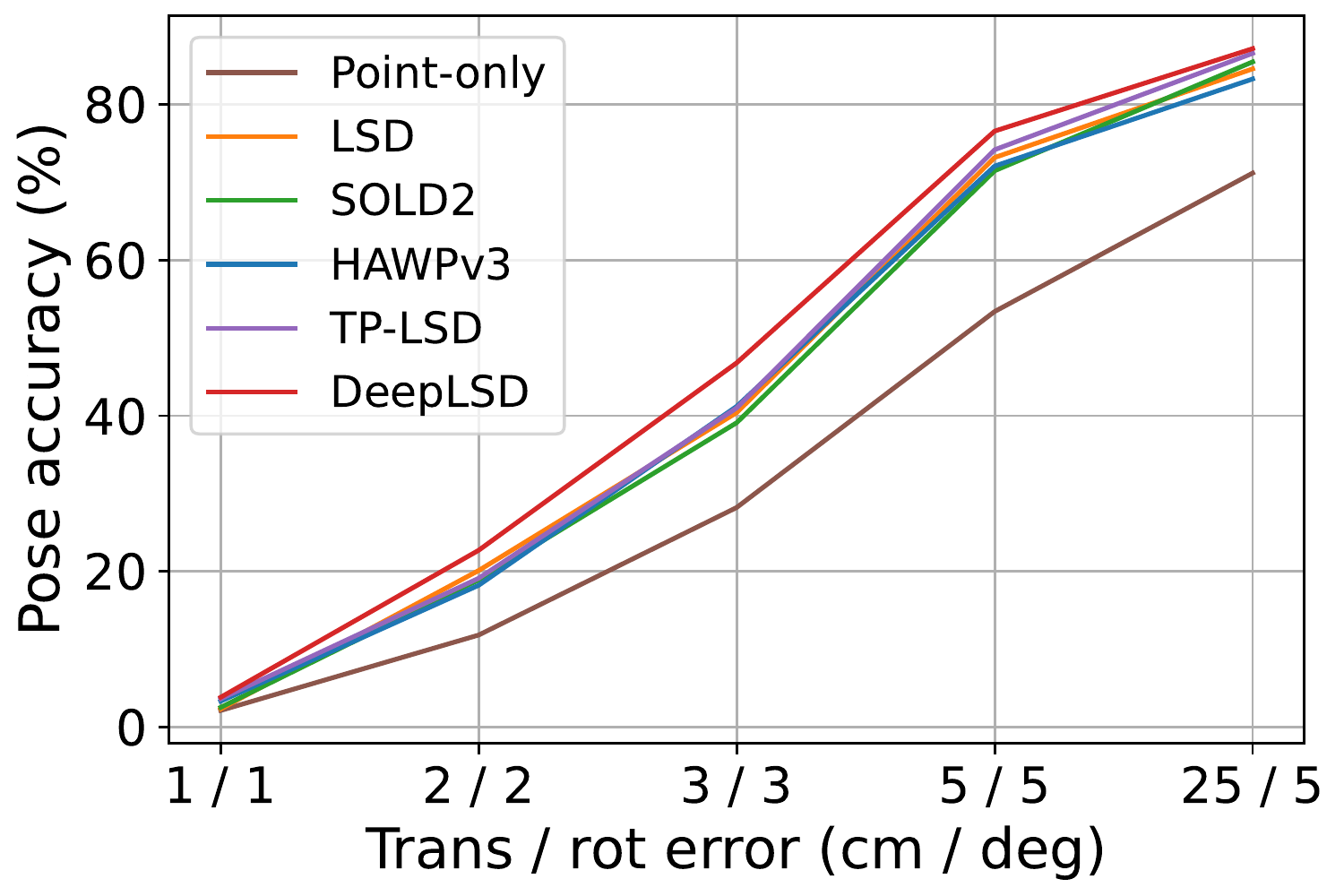}
    \end{minipage}
    \begin{minipage}{0.44\columnwidth}
        \scriptsize
        \setlength{\tabcolsep}{3.5pt}
        \begin{tabular}{lcc}
            \toprule
             & T / R err $\downarrow$ & Acc $\uparrow$ \\
            \midrule
            Point-only & 4.7 / 1.25 & 53.4 \\
            LSD~\cite{von2008lsd} & 3.4 / 0.94 & 73.2 \\
            SOLD2~\cite{Pautrat_Lin_2021_CVPR} & 3.5 / 0.96 & 71.5 \\
            HAWPv3~\cite{HAWP-journal} & 3.4 / 0.93 & 72.1 \\
            TP-LSD~\cite{huang2020tp} & 3.4 / 0.98 & 74.2 \\
            DeepLSD & \textbf{3.1 / 0.85} & \textbf{76.6} \\
            \bottomrule
        \end{tabular}
    \end{minipage}
    \vspace{-0.2cm}
    \caption{\textbf{Visual localization on 7Scenes stairs~\cite{7scenes}.} We evaluate the median translation and rotation errors (cm / deg), the pose accuracy at a 5 cm / 5 deg threshold, and plot the pose accuracy curve for various thresholds.}
    \label{fig:7scenes}
\end{figure}

Figure~\ref{fig:7scenes} shows that DeepLSD obtains the best performance on this challenging dataset. One can highlight the large boost of performance brought by line features compared to using points only. Lines are indeed still present and well localized in indoor environments such as in this scene, and can be matched even when in low-textured scenes.

\subsection{Impact of the Line Refinement}
\label{sec:line_refinement}
%
We evaluate applying our proposed line refinement as a post-processing step for several learned detection methods. 
Classical detectors are usually already accurate enough, so that our refinement would not enhance them much.
For each method, we compare the raw lines with the lines and VPs optimized by our line optimization. Table~\ref{tab:refinement_wireframe} shows results of line detectors on the 462 images of the test set of the Wireframe dataset~\cite{wireframe}. The second image is obtained using a synthetic homographic warp of the first image. We use the Wireframe dataset as it has a lot of well-defined vanishing points, which can be leveraged during the optimization.
We include results for our proposed optimization with and without the VP constraint to show the increased accuracy with VPs. As we want to highlight the gain in accuracy, we compute repeatability with an error threshold of only 1 pixel.

\begin{table}[tb]
    \centering
    \scriptsize
    \setlength{\tabcolsep}{3.2pt}
    \renewcommand{\arraystretch}{1}
    \begin{tabular}{llcccccccc}
    \toprule
      & & \multicolumn{2}{c}{Struct} & \multicolumn{2}{c}{Orth} & \multirow{2}{*}{\makecell{H\\estim}} &\multirow{2}{*}{\makecell{\# lines\\/ img}} & \multirow{2}{*}{\makecell{Time\\ $[$ms$]$ $\downarrow$}}  \\
      \cmidrule(lr){3-4} \cmidrule(lr){5-6}
      & & Rep $\uparrow$ & LE $\downarrow$ & Rep $\uparrow$ & LE $\downarrow$ & & &  \\
     \midrule
     \multirow{3}{*}{\makecell{HAWP\\\cite{hawp}}} & Baseline & 0.253 & 1.34 & 0.253 & 1.43 & 0.701 & \multirow{3}{*}{95.2} & \phantom{1}\fst{40}  \\
     & Opt w/o VP & 0.300 & 1.293 & 0.399 & 1.067 & 0.864 &  & 142  \\
     & Opt w/\phantom{o} VP & \fst{0.318} & \fst{1.245} & \fst{0.431} & \fst{0.967} & \fst{0.892} &  & 300  \\
     \midrule
     \multirow{3}{*}{\makecell{TP-LSD\\\cite{huang2020tp}}} & Baseline & 0.273 & 1.379 & 0.342 & 1.269 & 0.658 & \multirow{3}{*}{90.8} & \phantom{1}\fst{46}  \\
     & Opt w/o VP & 0.314 & 1.326 & 0.470 & 0.949 & 0.898 &  & 145  \\
     & Opt w/\phantom{o} VP & \fst{0.331} & \fst{1.277} & \fst{0.512} & \fst{0.861} & \fst{0.913} &  & 297  \\
     \midrule
     \multirow{3}{*}{\makecell{SOLD2\\\cite{Pautrat_Lin_2021_CVPR}}} & Baseline & \fst{0.197} & \fst{1.277} & 0.333 & 0.894 & 0.848 & \multirow{3}{*}{166.7} & \fst{297}  \\
     & Opt w/o VP & 0.172 & 1.388 & 0.339 & 0.814 & \fst{0.935} &  & 426  \\
     & Opt w/\phantom{o} VP & 0.185 & 1.330 & \fst{0.368} & \fst{0.753} & 0.920 &  & 697  \\
     \midrule
     \multirow{3}{*}{\makecell{DeepLSD\\(Ours)}} & Baseline & 0.318 & 0.941 & 0.489 & 0.574 & 0.991 & \multirow{3}{*}{168.8} & \phantom{1}\fst{68}  \\
     & Opt w/o VP & 0.314 & 0.938 & 0.482 & 0.575 & \fst{0.994} &  & 154  \\
     & Opt w/\phantom{o} VP & \fst{0.319} & \fst{0.927} & \fst{0.501} & \fst{0.544} & 0.981 &  & 542  \\
     \bottomrule
     \end{tabular}
    \vspace{-0.2cm}
    \caption{\textbf{Line refinement on the Wireframe dataset~\cite{wireframe}.} We use an error threshold of 1 pixel for the repeatability metrics. The refinement can significantly improve the localization error and homography score of inaccurate methods.}
     \label{tab:refinement_wireframe}
\end{table}

Results show that the refinement can significantly improve all metrics evaluating the accuracy of the lines, i.e. the localization error and homography estimation. 
This is particularly true for HAWP~\cite{hawp} and TP-LSD~\cite{huang2020tp}, with a decrease in localization error with orthogonal distance of up to $32\%$ for both, and an improvement of homography score of $27\%$ and $39\%$.
The benefits brought by the refinement are lower for our method, as its raw predicted lines are already sub-pixel accurate and the optimization is limited by the resolution of the DF and AF. 
Nonetheless, it can slightly improve most metrics. 
A limitation of this refinement is the execution time, which grows linearly with the number of lines, and requires running two networks.

\subsection{Ablation Study}
%
We validate our design choices on the HPatches dataset~\cite{hpatches_2017_cvpr} with low-level detector metrics. We compare our proposed approach with the same model detecting single edges instead of double ones, our network trained without the DF normalization, and a version of the HAWP~\cite{hawp} backbone re-trained on our line GT on the MegaDepth dataset~\cite{MegaDepthLi18}. The results of Table~\ref{tab:ablation} emphasize the importance of each component. Note that re-training HAWP~\cite{hawp} on our lines yields poor results due to the high number of GT lines, and the fact that generic lines have often noisy endpoints, so that predicting an angle to the two endpoints is noisy as well.

\begin{table}[tb]
    \centering
    \scriptsize
    \setlength{\tabcolsep}{5pt}
    \renewcommand{\arraystretch}{1}
    \begin{tabular}{lccccccc}
    \toprule
      & \multicolumn{2}{c}{Struct} & \multicolumn{2}{c}{Orth} & \multirow{2}{*}{\makecell{H\\estim}} &\multirow{2}{*}{\makecell{\# lines\\/ img}}  \\
      \cmidrule(lr){2-3} \cmidrule(lr){4-5}
      & Rep $\uparrow$ & LE $\downarrow$ & Rep $\uparrow$ & LE $\downarrow$ & &  \\
     \midrule
     Single edge & 0.241 & 2.121 & 0.328 & 1.686 & 0.434 & 130.8 \\ 
     No DF normalization & 0.344 & 1.343 & 0.475 & 0.879 & 0.674 & 439.6 \\ 
     HAWP with our lines & 0.209 & 2.138 & 0.239 & 1.840 & 0.245 & 98.0 \\ 
     DeepLSD (Ours) & \textbf{0.367} & \textbf{1.235} & \textbf{0.485} & \textbf{0.818} & \textbf{0.705} & \textbf{486.2} \\ 
     \bottomrule
     \end{tabular}
    \vspace{-0.2cm}
    \caption{\textbf{Ablation study on the HPatches dataset~\cite{hpatches_2017_cvpr}.} We compare DeepLSD to alternatives detecting single edges, without DF normalization and with HAWP re-trained on our line GT.}
     \label{tab:ablation}
\end{table}

%
\section{Conclusion}
%
We presented a hybrid line segment detector combining the robustness of deep learning and the accuracy of handcrafted detectors, using a learned surrogate image gradient as intermediate representation. 
Without the requirement of ground truth lines, our method can be trained on any dataset and is suitable for most tasks including line segments. 
Finally, we proposed a line refinement able to improve the accuracy of our method and to bridge the gap in line localization between deep line detectors and handcrafted ones.
We believe that our general-purpose lines will open new possibilities to use line segments in the wild.

{\footnotesize
\vspace{0.5em}
\noindent \textbf{Acknowledgments.}
We would like to warmly thank Iago Suarez for reviewing this paper and for the insightful discussions, as well as Yifan Yu for sharing his code for visual localization. Daniel Barath was supported by the ETH Postdoc Fellowship and Viktor larsson by ELLIIT.
}

\clearpage

\appendix

\twocolumn[
    {\centering \Large Supplementary Material \\[1ex]}
    \vspace*{3ex}
]

In the following we provide additional results, insights and visualizations for DeepLSD. Section~\ref{sec:network} describes in details our network architecture, Section~\ref{sec:ablation} introduces additional ablation studies and insights about our approach, Section~\ref{sec:full_7scenes} provides an evaluation of visual localization with points and lines on the full 7Scenes dataset, Section~\ref{sec:vp_estimation} gives additional results about vanishing point estimation from the detected line segments, Section~\ref{sec:3d_reconstruction} displays visualizations of the 3D reconstruction, Section~\ref{sec:limitations} highlights some limitations of our method, and finally Section~\ref{sec:visualizations} offers examples of the line detections.

\section{Network Architecture}  \label{sec:network}
\begin{figure*}
    \centering
    \includegraphics[width=0.98\textwidth]{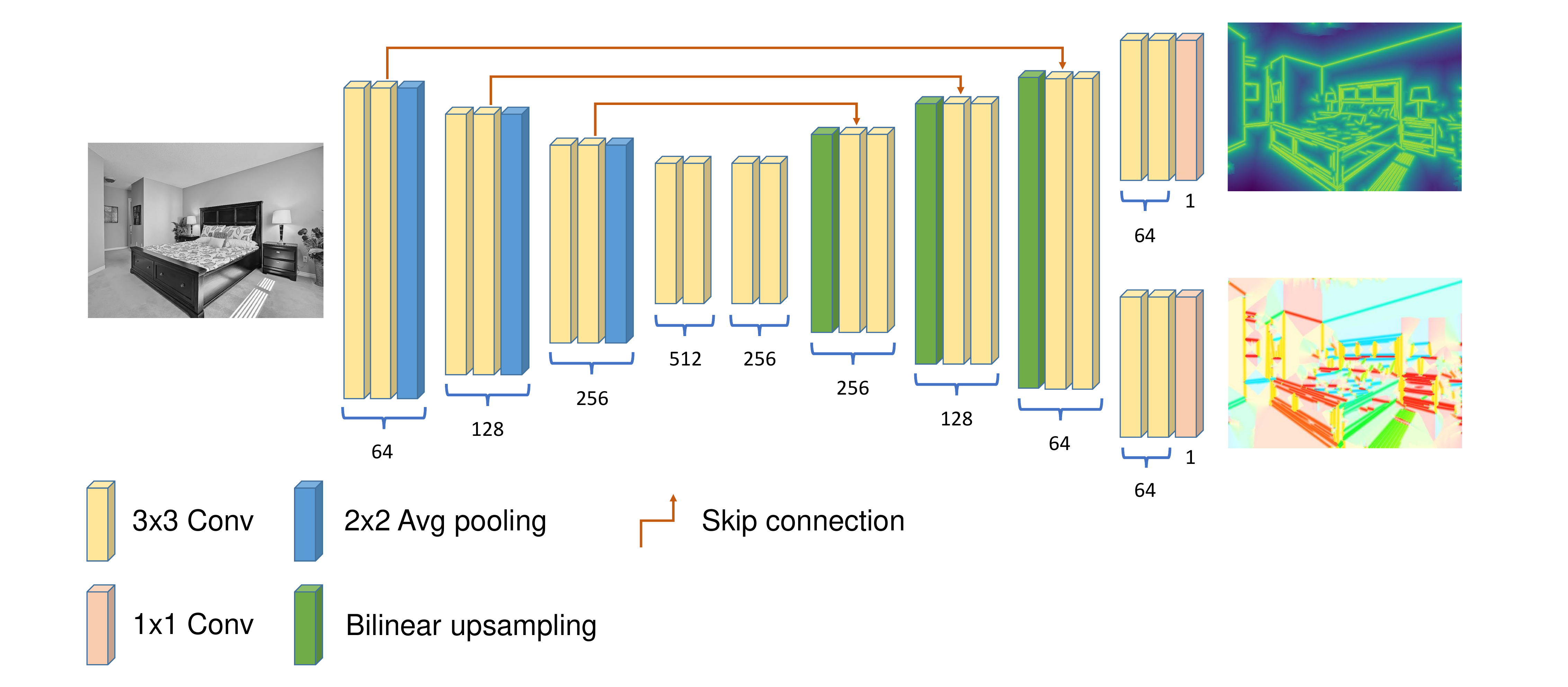}
    \caption{\textbf{Network architecture.} We use a standard UNet~\cite{ronneberger_2015} architecture to predict the distance and angle fields.}
    \label{fig:network_architecture}
\end{figure*}

%
We provide more details about the network architecture that we used to predict attraction fields. 
We use a simple U-Net-like architecture~\cite{ronneberger_2015} with several blocks of convolutions, downsampling the initial image by a factor of 8 and then upsampling it again to the initial resolution. 
Downsampling is performed through 3 successive $2 \times 2$ average poolings and upsampling is done with bilinear interpolation. 
A skip connection is added before each downsampling layer and is concatenated with the output of the corresponding upsampling layer. 
Please refer to Figure~\ref{fig:network_architecture} for the detailed architecture.
Each convolution layer is followed by ReLU activation~\cite{agarap2018deep} and Batch Normalization~\cite{ioffe_2015}, except the final layer of each branch. 
The activations of the two output branches are ReLU for the distance field and Sigmoid for the angle field, without batch normalization.

\section{Additional Ablation Studies}  \label{sec:ablation}
%
\subsection{Generalization to Other Traditional Detectors}
%
While DeepLSD is using LSD~\cite{von2008lsd} as its base line detector, our approach can be applied to any other traditional detector leveraging the image gradient. 
We show here the results of our method using ELSED~\cite{suarez2021elsed} as base detector (coined DeepELSED) and compare it to the original ELSED in Table~\ref{tab:deep_elsed}.
We give the results for the raw lines without any refinement on low-level line detection metrics on the HPatches~\cite{hpatches_2017_cvpr} and RDNIM~\cite{Pautrat_2020_ECCV} datasets. 
For both traditional detectors LSD and ELSED, our deep version can improve most metrics, thanks to the additional robustness brought by the learned processing of the image.

\begin{table}
    \centering
    \scriptsize
    \setlength{\tabcolsep}{5pt}
    \newcommand{\groupspace}{15pt}
    \begin{tabular}{lllcccc}
    \toprule
     & & & \multicolumn{2}{c}{LSD~\cite{von2008lsd}} & \multicolumn{2}{c}{ELSED~\cite{suarez2021elsed}} \\
     \cmidrule(lr){4-5} \cmidrule(lr){6-7}
     & & & Traditional & DeepLSD & Traditional & DeepELSED \\
     \midrule
     \multirow{7}{*}{\rotatebox{90}{HPatches~\cite{hpatches_2017_cvpr}\phantom{xxx}}} & \multirow{2}{*}{Struct} & Rep $\uparrow$ & 0.314 & \fst{0.367} & 0.240 & \fst{0.263} \\
      &  & LE $\downarrow$ & 1.309 & \fst{1.235} & \fst{1.551} & 1.585 \\ \cmidrule{2-3}
      & \multirow{2}{*}{Orth} & Rep $\uparrow$ & 0.468 & \fst{0.485} & 0.465 & \fst{0.478} \\
      &  & LE $\downarrow$ & \fst{0.793} & 0.818 & 0.845 & \fst{0.839} \\ \cmidrule{2-3}
      & \multicolumn{2}{l}{H estimation $\uparrow$} & 0.697 & \fst{0.705} & 0.617 & \fst{0.624} \\ \cmidrule{2-7}
      & \multicolumn{2}{l}{\# lines / img} & 492.6 & 486.2 & 425.4 & 419.4 \\
      & \multicolumn{2}{l}{Time [ms] $\downarrow$} & \fst{104} & 271 & \fst{10} & 144 \\
     \midrule
     \multirow{7}{*}{\rotatebox{90}{RDNIM~\cite{Pautrat_2020_ECCV}\phantom{xxx}}} & \multirow{2}{*}{Struct} & Rep $\uparrow$ & 0.283 & \fst{0.285} & 0.209 & \fst{0.230} \\
      &  & LE $\downarrow$ & 2.039 & \fst{1.733} & 2.303 & \fst{2.258} \\  \cmidrule{2-3}
      & \multirow{2}{*}{Orth} & Rep $\uparrow$ & \fst{0.403} & 0.394 & 0.392 & \fst{0.407} \\
      &  & LE $\downarrow$ & 1.369 & \fst{1.098} & \fst{1.248} & 1.361 \\ \cmidrule{2-3}
      & \multicolumn{2}{l}{H estimation $\uparrow$} & 0.468 & \fst{0.591} & 0.200 & \fst{0.221} \\ \cmidrule{2-7}
      & \multicolumn{2}{l}{\# lines / img} & 191.4 & 400.0 & 112.0 & 162 \\
      & \multicolumn{2}{l}{Time [ms] $\downarrow$} & \fst{34} & 96 & \fst{3} & 88 \\
     \bottomrule
     \end{tabular}
     \caption{\textbf{Generalization to other traditional detectors.} Our method is not limited to LSD~\cite{von2008lsd}, but can also be applied to the ELSED~\cite{suarez2021elsed} line detector for example. We show the comparison between our approach and the original detectors on the HPatches~\cite{hpatches_2017_cvpr} and RDNIM~\cite{Pautrat_2020_ECCV} datasets. The first three columns are identical to Table 1 in the main paper and our results are given without the final line refinement.}
     \label{tab:deep_elsed}
\end{table}

\subsection{Line Refinement on Traditional Methods}
%
The proposed line refinement is mainly aiming at improving the accuracy of previous deep line detectors and DeepLSD, but one can wonder how it performs with traditional methods. 
When refining the lines output by LSD~\cite{von2008lsd} and ELSED~\cite{suarez2021elsed} on the Wireframe dataset, we did not observe any improvement in low-level metrics, except for a boost of performance in homography estimation for ELSED (see Table~\ref{tab:refinement_handcrafted}). 
Traditional detectors are indeed already sub-pixel accurate, so that the limited resolution of the distance field is not high enough to refine the lines further. 
The drop in performance in most metrics can be explained by the fact that some lines detected by these methods are in areas with high distance field values, so that these lines will rather drift than being optimized correctly. 
However, relevant lines for downstream tasks still seem to benefit from the refinement as shown by the large boost in homography estimation for ELSED.

\begin{table}[tb]
    \centering
    \scriptsize
    \setlength{\tabcolsep}{2.7pt}
    \renewcommand{\arraystretch}{1}
    \begin{tabular}{clcccccccc}
    \toprule
      & & \multicolumn{2}{c}{Struct} & \multicolumn{2}{c}{Orth} & \multirow{2}{*}{\makecell{H\\estimation}} &\multirow{2}{*}{\makecell{\# lines\\/ img}} & \multirow{2}{*}{\makecell{Time\\ $[$ms$]$ $\downarrow$}}  \\
      \cmidrule(lr){3-4} \cmidrule(lr){5-6}
      & & Rep $\uparrow$ & LE $\downarrow$ & Rep $\uparrow$ & LE $\downarrow$ & & &  \\
     \midrule
     \multirow{3}{*}{\makecell{LSD\\\cite{von2008lsd}}} & Baseline & \textbf{0.386} & \fst{0.456} & \fst{0.647} & \fst{0.12} & \fst{0.998} & \multirow{3}{*}{352.1} & \phantom{1}\fst{23}  \\
     & Opt w/o VP & 0.332 & 0.593 & 0.485 & 0.35 & 0.994 &  & 217  \\
     & Opt w/\phantom{o} VP & 0.332 & 0.589 & 0.494 & 0.325 & 0.994 &  & 545  \\
     \midrule
     \multirow{3}{*}{\makecell{ELSED\\\cite{suarez2021elsed}}} & Baseline & \fst{0.185} & \fst{1.238} & \fst{0.564} & \fst{0.36} & 0.926 & \multirow{3}{*}{178.2} & \phantom{2}\fst{3}  \\
     & Opt w/o VP & 0.165 & 1.315 & 0.462 & 0.529 & \fst{0.989} &  & 130  \\
     & Opt w/\phantom{o} VP & 0.164 & 1.313 & 0.474 & 0.502 & \fst{0.989} &  & 397  \\
     \bottomrule
     \end{tabular}
    \caption{\textbf{Line refinement of traditional methods on the Wireframe dataset~\cite{wireframe}.} The line refinement can be detrimental for some outlier lines outside of the distance field, but it is still able to improve the accuracy of most lines, as shown by the boost of performance of ELSED in homography estimation.}
     \label{tab:refinement_handcrafted}
\end{table}

\subsection{Training Learned Baselines with our Supervision Strategy}
\begin{figure*}[tb]
    \centering
    \scriptsize
    \begin{tabular}{ccc}
        HAWP~\cite{hawp} & Re-trained HAWP & DeepLSD (Ours) \\
        \includegraphics[width=0.32\textwidth]{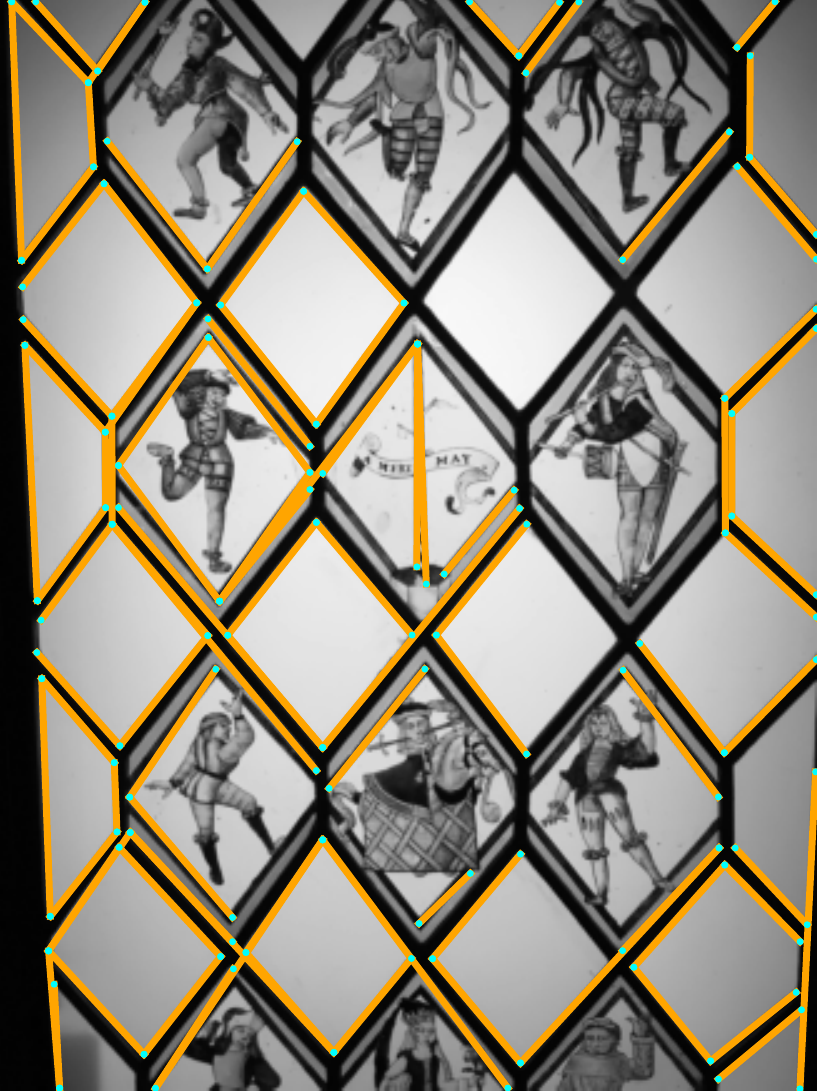} &
        \includegraphics[width=0.32\textwidth]{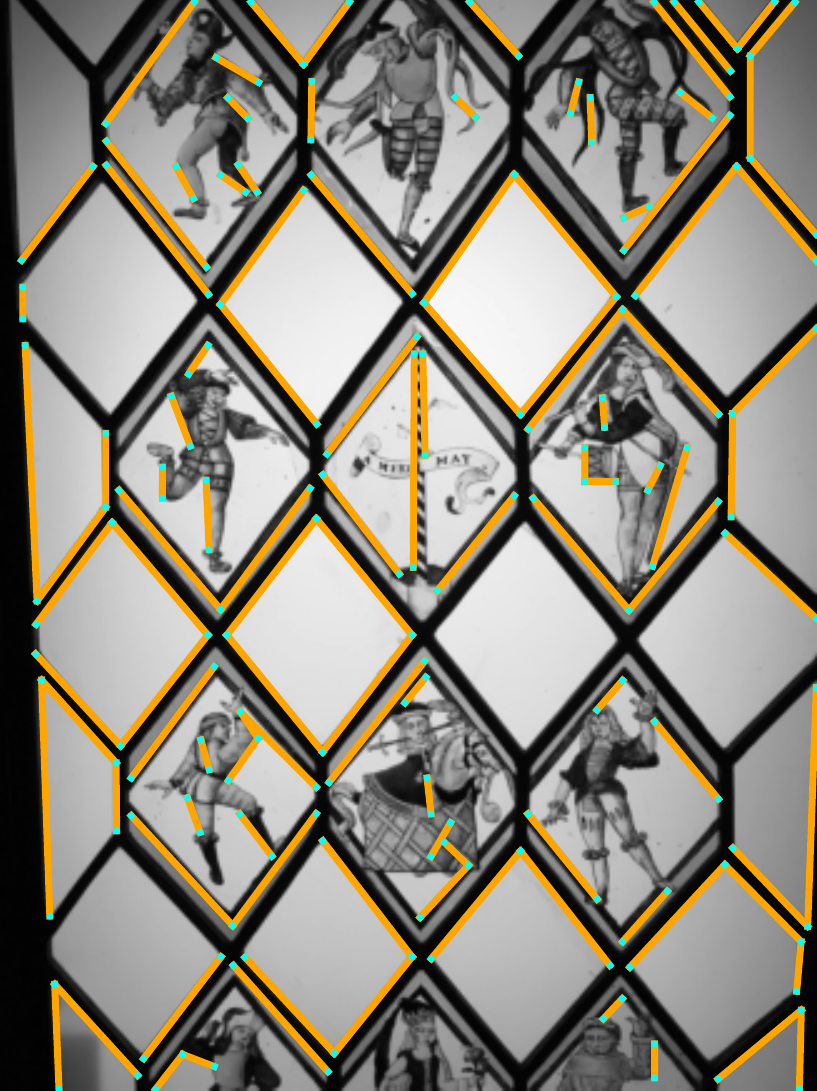} &
        \includegraphics[width=0.32\textwidth]{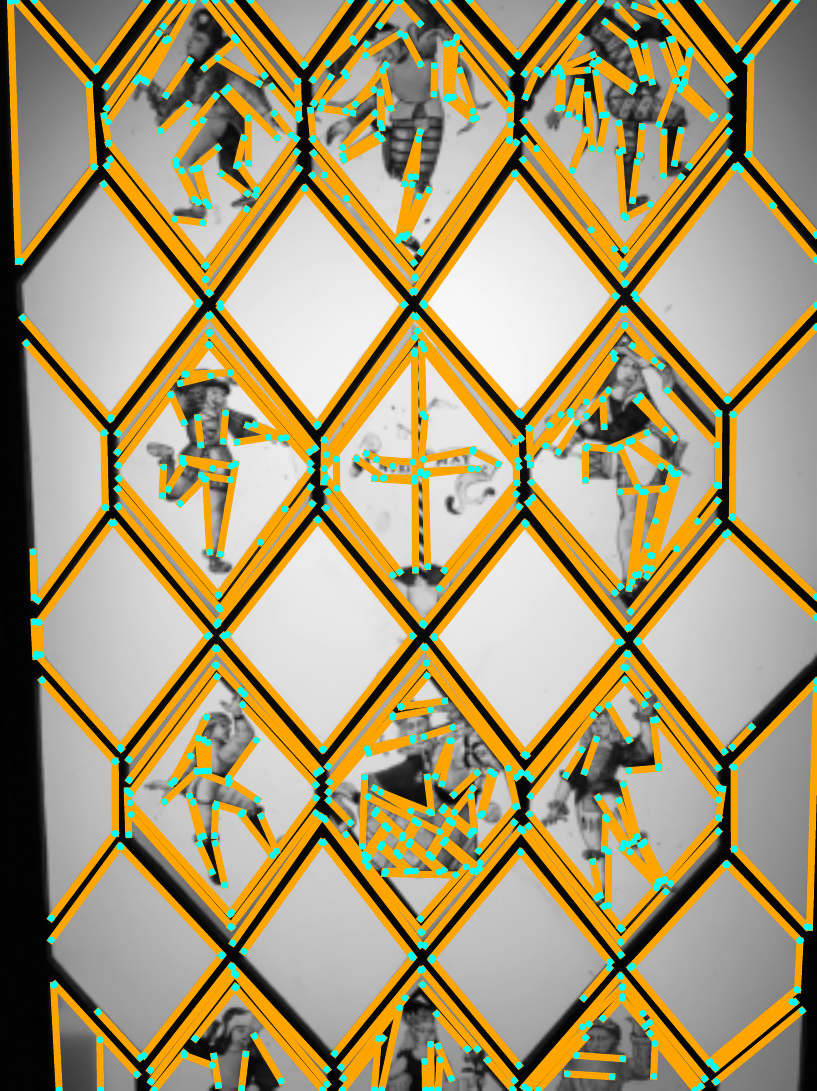} \\
        \includegraphics[width=0.32\textwidth]{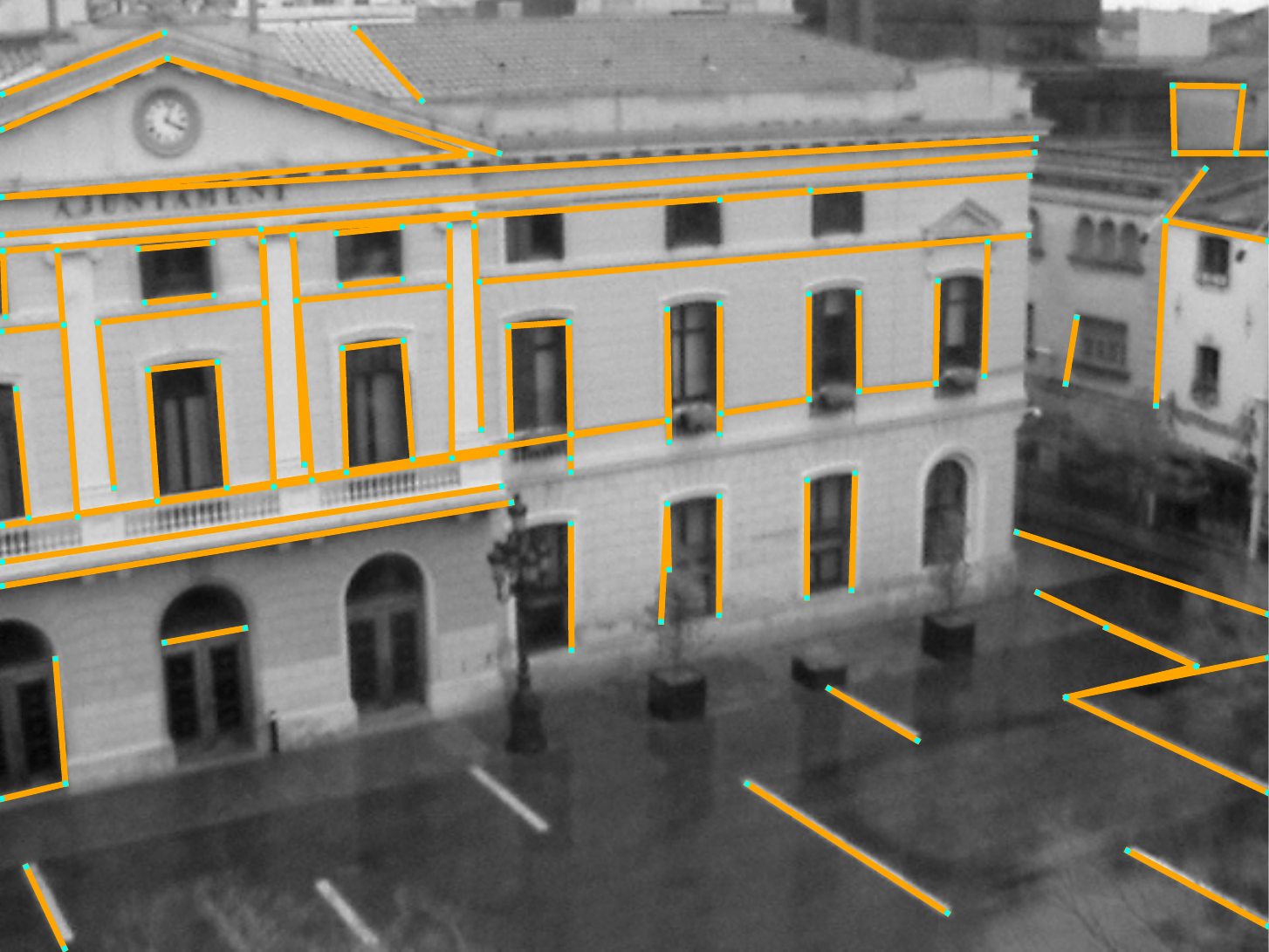} &
        \includegraphics[width=0.32\textwidth]{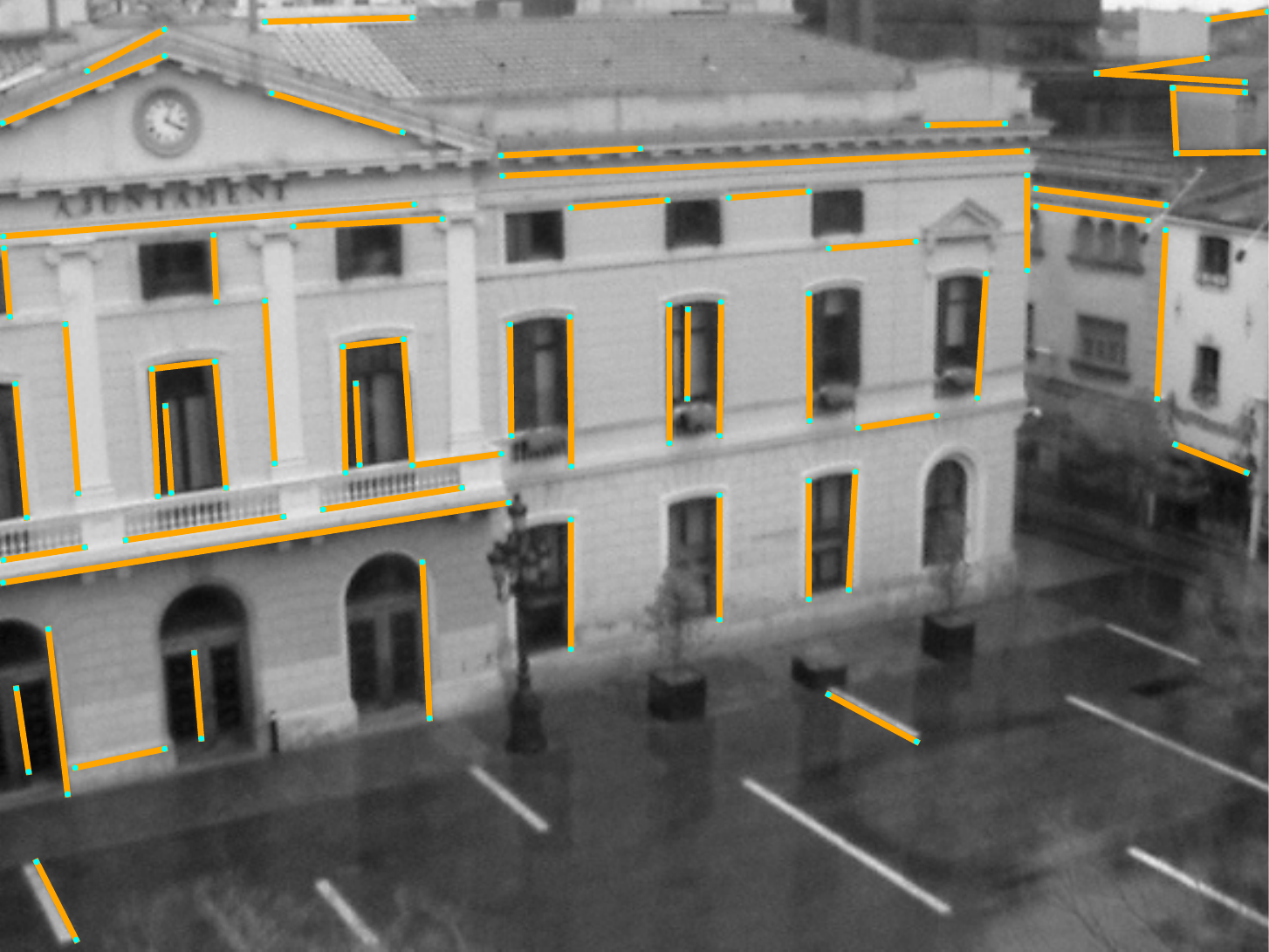} &
        \includegraphics[width=0.32\textwidth]{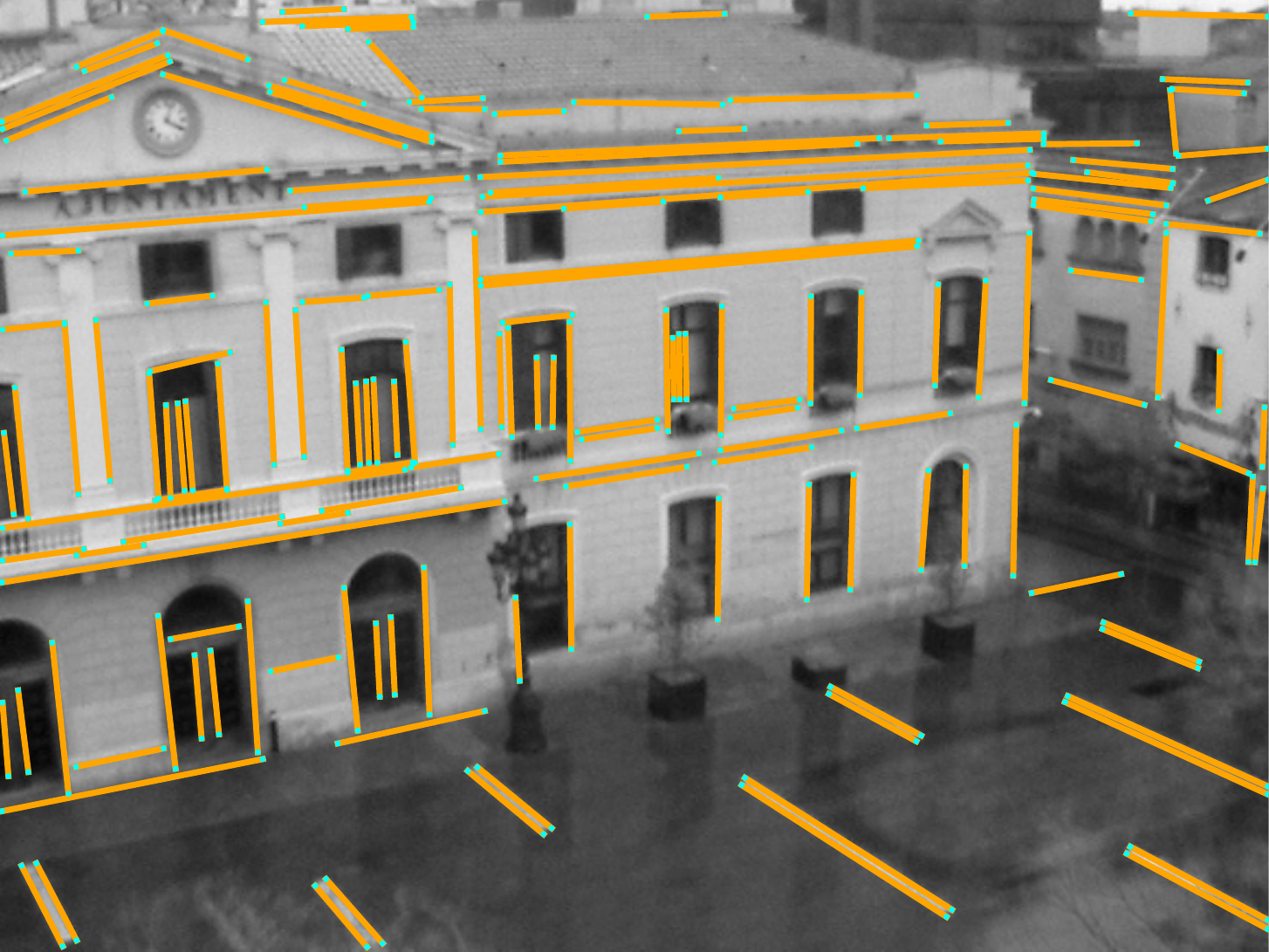} \\
    \end{tabular}
    \caption{\textbf{Re-training the HAWP detector~\cite{hawp} with the proposed pseudo ground truth lines.} It yields unsatisfactory lines compared to the DeepLSD approach, mainly because detecting line endpoints with a network prediction is challenging for high densities of line segments.}
    \label{fig:baseline_retraining}
\end{figure*}

%
In the main paper, we proposed an ablation study by re-training the HAWP~\cite{hawp} detector with our ground truth (GT) supervision. We provide here additional details and visualizations of this ablation.
Instead of taking our DeepLSD approach of predicting the distance and angle fields and then applying LSD on top of it, one could also extract lines from the ground truth distance and angle fields, and then use these lines to supervise any existing deep line detector.
Figure~\ref{fig:baseline_retraining} shows two examples of lines detected by the original HAWP, the re-trained version using our GT lines, and DeepLSD. 
The latter remains the most satisfactory one, and thus justifies our approach of leveraging traditional line detectors instead of end-to-end line detection. 
One reason for the lower quality of the re-trained HAWP is that predicting the position of endpoints with an additional attraction field is not suitable for generic lines, as there are often too many of them in most images. 
This approach works better for wireframe lines, which are sparser and require less accuracy.

\section{Additional Visual Localization Results}
\label{sec:full_7scenes}
%
While the main paper focuses on the most challenging scene of the 7Scenes dataset~\cite{7scenes}, Stairs, we provide here the results of visual localization on the full dataset. As described in the main paper, we detect keypoints with SuperPoint~\cite{superpoint}, match them with SuperGlue~\cite{sarlin_2020_superglue}, and build on top of hloc~\cite{sarlin2019coarse,hloc} by adding line features and using them in the pose estimation. The lines are again matched with the SOLD2~\cite{Pautrat_Lin_2021_CVPR} line detector. Table~\ref{tab:full_7scenes} displays the results of several state-of-the-art line detectors in terms of translation and rotation errors, as well as pose accuracy at a 5 cm / 5 degree threshold. DeepLSD obtains the best translation error on all scenes, as well as the best metrics on the full dataset. It can be noted that the improvement with respect to previous methods is rather small, due to the fact that 7Scenes is already very saturated for visual localization. 

\begin{table*}[]
    \centering
    \scriptsize
    \setlength{\tabcolsep}{6pt}
    \begin{tabular}{ccccccc}
        \toprule
         & \makecell{Point-only\\SP~\cite{superpoint} + SG~\cite{sarlin_2020_superglue}} & LSD~\cite{von2008lsd} & SOLD2~\cite{Pautrat_Lin_2021_CVPR} & TP-LSD~\cite{huang2020tp} & HAWPv3~\cite{HAWP-journal} & DeepLSD \\
        \midrule
        Chess & \textbf{2.4} / 0.81 / \textbf{94.5} & \textbf{2.4} / 0.82 / 94.4 & \textbf{2.4} / 0.81 / 94.0 & \textbf{2.4} / \textbf{0.80} / 94.4 & \textbf{2.4} / \textbf{0.80} / \textbf{94.5} & \textbf{2.4} / 0.82 / \textbf{94.5} \\
        Fire & 1.9 / 0.76 / 96.4 & \textbf{1.7} / 0.73 / 96.5 & 1.8 / 0.76 / 95.9 & 1.8 / 0.76 / 95.8 & 1.9 / 0.77 / \textbf{97.1} & \textbf{1.7} / \textbf{0.70} / 96.7 \\
        Heads & 1.1 / 0.74 / 99.0 & 1.1 / 0.74 / 99.4 & 1.1 / 0.76 / 99.3 & 1.1 / \textbf{0.73} / \textbf{99.5} & 1.1 / 0.80 / 99.2 & \textbf{1.0} / \textbf{0.73} / \textbf{99.5} \\
        Office & 2.7 / 0.83 / 83.9 & \textbf{2.6} / \textbf{0.79} / 84.7 & 2.7 / 0.82 / 83.8 & \textbf{2.6} / 0.81 / 84.1 & 2.7 / 0.82 / 83.8 & \textbf{2.6} / 0.80 / \textbf{85.0} \\
        Pumpkin & 4.0 / 1.05 / 62.0 & 4.0 / 1.04 / 62.1 & 4.1 / 1.07 / 60.7 & 4.0 / 1.04 / \textbf{62.6} & 4.0 / 1.04 / 62.3 & \textbf{3.9} / \textbf{1.02} / 62.2 \\
        Redkitchen & 3.3 / \textbf{1.12} / 72.5 & \textbf{3.2} / 1.14 / 73.2 & \textbf{3.2} / \textbf{1.12} / \textbf{73.5} & \textbf{3.2} / \textbf{1.12} / 73.2 & 3.3 / 1.13 / 73.0 & \textbf{3.2} / 1.13 / 73.4 \\
        Stairs & 4.7 / 1.25 / 53.4 & 3.4 / 0.94 / 73.2 & 3.5 / 0.96 / 71.5 & 3.4 / 0.93 / 72.1 & 3.4 / 0.98 / 74.2 & \textbf{3.1} / \textbf{0.85} / \textbf{76.6} \\
        \midrule
        Total & 2.9 / 0.94 / 80.2 & \textbf{2.6} / 0.89 / 83.4 & 2.7 / 0.90 / 82.7 & \textbf{2.6} / 0.88 / 83.1 & 2.7 / 0.91 / 83.4 & \textbf{2.6} / \textbf{0.86} / \textbf{84.0} \\
        \bottomrule
    \end{tabular}
    \caption{\textbf{Visual localization on the 7Scenes dataset.} We report the translation error (in cm) / rotation error (in deg) / pose accuracy at a 5 cm / 5 deg threshold (in \%) for the 7 scenes and the average score across all scenes.}
    \label{tab:full_7scenes}
\end{table*}

\section{Vanishing Point Estimation}
\label{sec:vp_estimation}
%
Another common application for line segments is the vanishing point (VP) estimation task. 
Given the line segments extracted by all the baselines and our method, we apply multi-model fitting with Progressive-X~\cite{Barath_2019_ICCV} to find an unconstrained number of (not necessarily orthogonal) VPs. 
A minimal set of 2 lines provides a VP candidate, and its consistency with the other lines is evaluated under the $d_{\text{VP}}$ metric~\cite{Tardif_2009}. This distance is computed as the average orthogonal distance between the endpoints of a line segment and the infinite line going from the VP to the midpoint of the segment.
Based on the inlier lines, we do a weighted least squares of the distance of all inliers to the VP, using the line length as weight. 
We tune the parameters of the model fitting algorithm for each method on a validation set.

We consider two benchmarks for vanishing point estimation. YorkUrbanDB~\cite{yorkurban} pictures 102 images (51 for validation and 51 for test) of urban scenes. 
It offers 2 or 3 ground truth VPs per image, ground truth lines, and the association between VPs and lines. 
Additionally, we consider the extended set of VPs proposed in YUD+~\cite{kluger2020consac}, which labels up to 8 VPs per image. 
The second dataset is adapted from the NYU Depth dataset V2~\cite{silberman_2012} by~\cite{kluger2020consac}, consisting of 1449 images (we keep the last 49 for parameter tuning), each labelled with 1 to 8 VPs.

We consider three metrics. \emph{VP consistency} counts the percentage of ground truth lines that are within a given threshold of the predicted VPs~\cite{Tardif_2009}. 
Each set of ground truth lines is associated to a single predicted VP and each VP can be associated with at most one set of lines. 
We only show this metric for YorkUrbanDB as NYU does not have manually labelled lines. 
\emph{VP error} measures how precise the estimated VPs are in 3D. 
It is the angular error between the directions in 3D of the ground truth VPs and the predicted ones. 
We perform again a 1:1 matching to optimally assign the predicted VPs to the ground truth ones. 
For each experiment, we run the VP detection algorithm 20 times and report the median results.
\emph{AUC} represents the Area Under the Curve (AUC) of the recall curve of the VPs, as described in \cite{kluger2020consac}. We show the average AUC and its standard deviation over 5 runs.

\begin{figure}
    \centering
    \includegraphics[width=0.75\columnwidth]{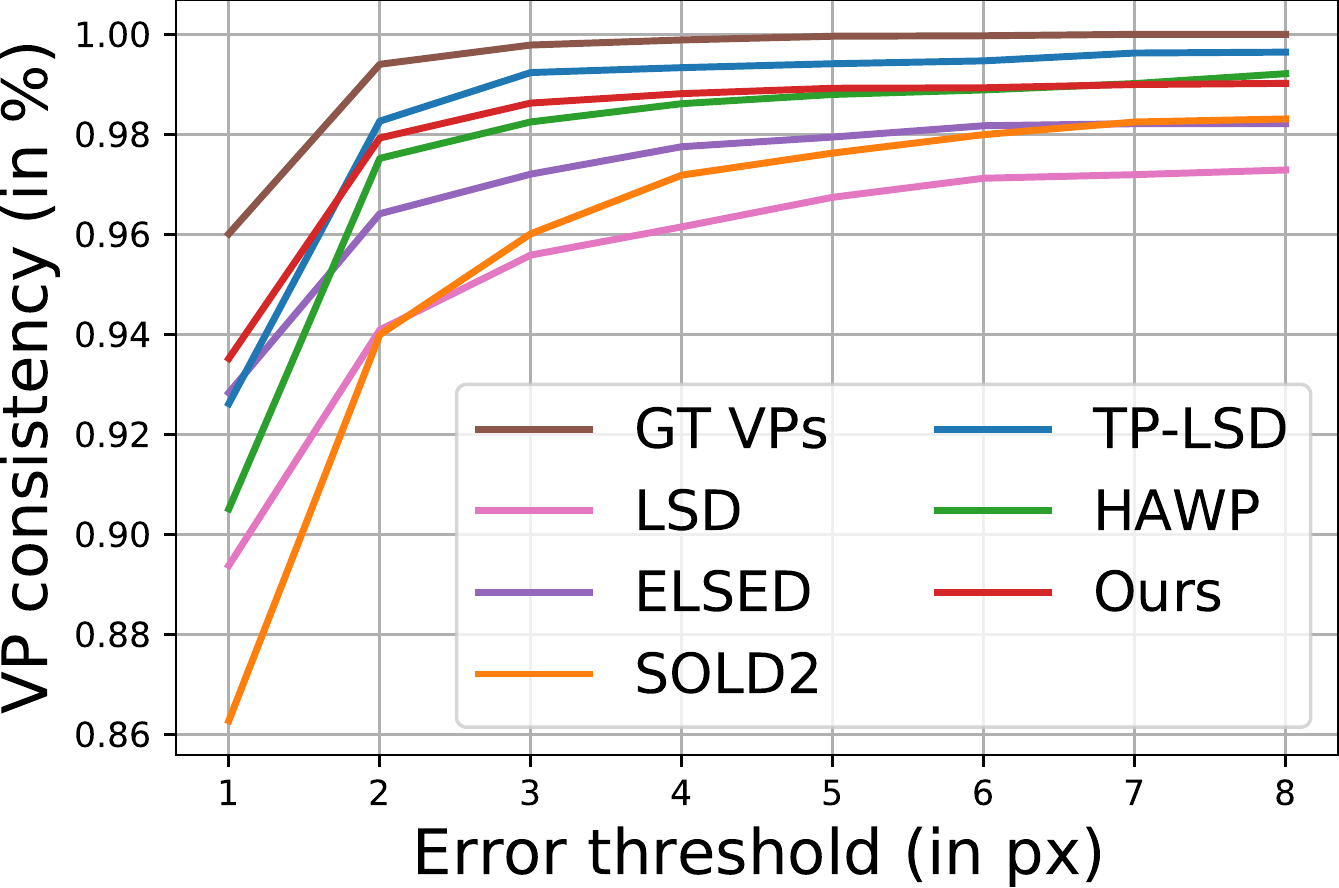}
    \caption{\textbf{VP consistency on the York Urban dataset~\cite{yorkurban}.} \OURS \ ranks first on the 1 pixel threshold of VP consistency, meaning that it leads to the largest number of highly accurate VPs.}
    \label{fig:vp_consistency}
\end{figure}

\begin{table}[]
    \centering
    \scriptsize
    \setlength{\tabcolsep}{4pt}
    \renewcommand{\arraystretch}{1.2}
    \begin{tabular}{lcccc}
         \toprule
          & \multicolumn{2}{c}{YUD+~\cite{yorkurban}} & \multicolumn{2}{c}{NYU-VP\cite{silberman_2012,kluger2020consac}} \\ \cmidrule{2-3} \cmidrule{4-5}
          & VP error $\downarrow$ & AUC $\uparrow$ & VP error $\downarrow$ & AUC $\uparrow$ \\
         \midrule
         LSD~\cite{von2008lsd} & 2.05 & 82.9 (5.3) & 3.29 & 68.6 (6.3) \\
         ELSED~\cite{suarez2021elsed} & 1.88 & 81.9 (6.0) & \fst{3.24} & 68.3 (6.6) \\
         HAWP~\cite{hawp} & 1.76 & 84.2 (4.2) & 3.35 & 68.0 (5.7) \\
         TP-LSD~\cite{huang2020tp} & 1.73 & 85.1 (5.0) & 3.35 & 68.0 (4.5) \\
         SOLD2~\cite{Pautrat_Lin_2021_CVPR} & 2.59 & 75.4 (6.4) & 4.46 & 56.9 (7.6) \\
         DeepLSD (Ours) & \fst{1.63} & \textbf{85.6} (3.6) & \fst{3.24} & \textbf{69.1} (6.2) \\
         \bottomrule
    \end{tabular}
    \caption{\textbf{VP estimation on York Urban~\cite{yorkurban} and NYU-VP~\cite{silberman_2012,kluger2020consac}.} We compare \OURS \ with other baselines in terms of median VP error and average recall AUC (and standard deviation). \OURS \ obtains the best performance overall.}
    \label{tab:vp_estimation}
\end{table}

Results are shown in Figure~\ref{fig:vp_consistency} and Table~\ref{tab:vp_estimation}.
The wireframe methods TP-LSD~\cite{huang2020tp} and HAWP~\cite{hawp} are particularly good for vanishing point estimation, as they only detect structural lines, which are usually the only relevant ones for VP estimation.
However, when evaluated on the more challenging and non-Manhattan scenes of NYU-VP, the handcrafted line detectors provide the best accuracy since they can detect all types of lines. 
Our proposed \OURS~outperforms all baselines in terms of VP error and AUC, and obtains the most consistent lines with the GT VPs at small thresholds in Figure~\ref{fig:vp_consistency}.

\begin{figure}[tb]
    \centering
    \begin{minipage}{0.55\columnwidth}
        \includegraphics[width=\columnwidth]{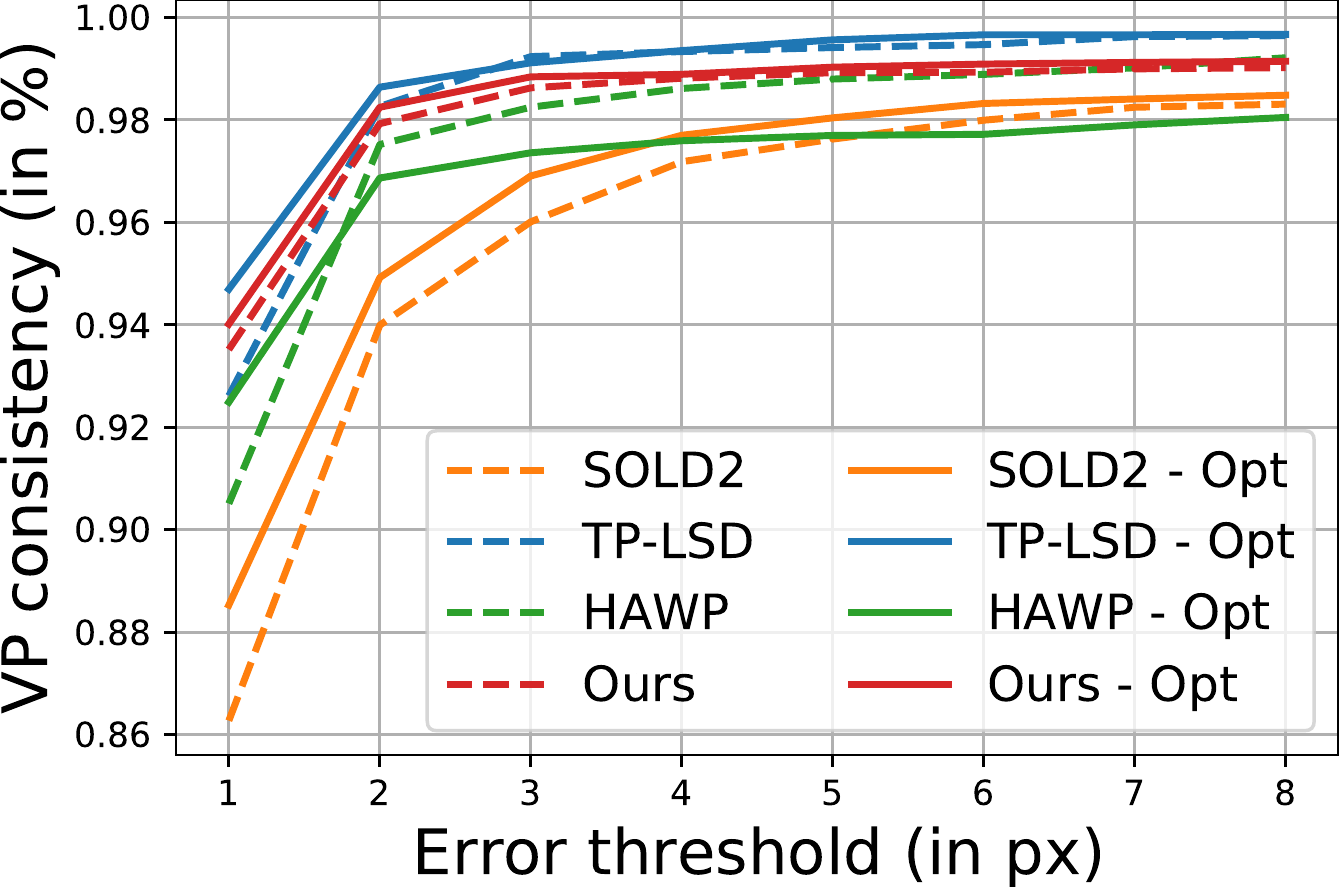}
    \end{minipage}
    \hspace{0.2cm}
    \begin{minipage}{0.4\columnwidth}
        \scriptsize
        \setlength{\tabcolsep}{0pt}
        \renewcommand{\arraystretch}{1.15}
        \begin{tabular}{lc}
             \toprule
              & \multirow{2}{*}{\makecell{VP error $\downarrow$}} \\
              & \\
             \midrule
             HAWP~\cite{hawp} & \fst{1.76} \\
             HAWP~\cite{hawp} - Opt & 1.78 \\
             \midrule
             TP-LSD~\cite{huang2020tp} & 1.73 \\
             TP-LSD~\cite{huang2020tp} - Opt & \fst{1.59} \\
             \midrule
             SOLD2~\cite{Pautrat_Lin_2021_CVPR} & 2.59 \\
             SOLD2~\cite{Pautrat_Lin_2021_CVPR} - Opt & \fst{2.28} \\
             \midrule
             DeepLSD (Ours) & 1.63 \\
             DeepLSD (Ours) - Opt & \fst{1.59} \\
             \bottomrule
        \end{tabular}
    \end{minipage}
    \caption{\textbf{Effect of the line refinement on VP estimation on YorkUrbanDB~\cite{yorkurban,kluger2020consac}.} The line optimization improves the VP consistency and error of most deep methods.}
    \label{fig:vp_estimation_refinement}
\end{figure}

We additionally study the effect of refinement on the VP estimation task in Figure~\ref{fig:vp_estimation_refinement}. 
We show again the difference in VP consistency on the YorkUrbanDB dataset~\cite{yorkurban} and VP error on YUD+~\cite{kluger2020consac}, with the optimization objective including VPs. Except for HAWP, all methods benefit from the refinement, showing that our refinement can improve the lines as much as their associated VPs.

\section{Line 3D Reconstruction}  \label{sec:3d_reconstruction}
\begin{figure*}[tb]
    \centering
    \scriptsize
    \newcommand{\szp}{0.24}
    \setlength{\tabcolsep}{3pt}
    \begin{tabular}{cccc}
        \multicolumn{4}{c}{ai\_001\_001} \\
        \includegraphics[width=\szp\textwidth]{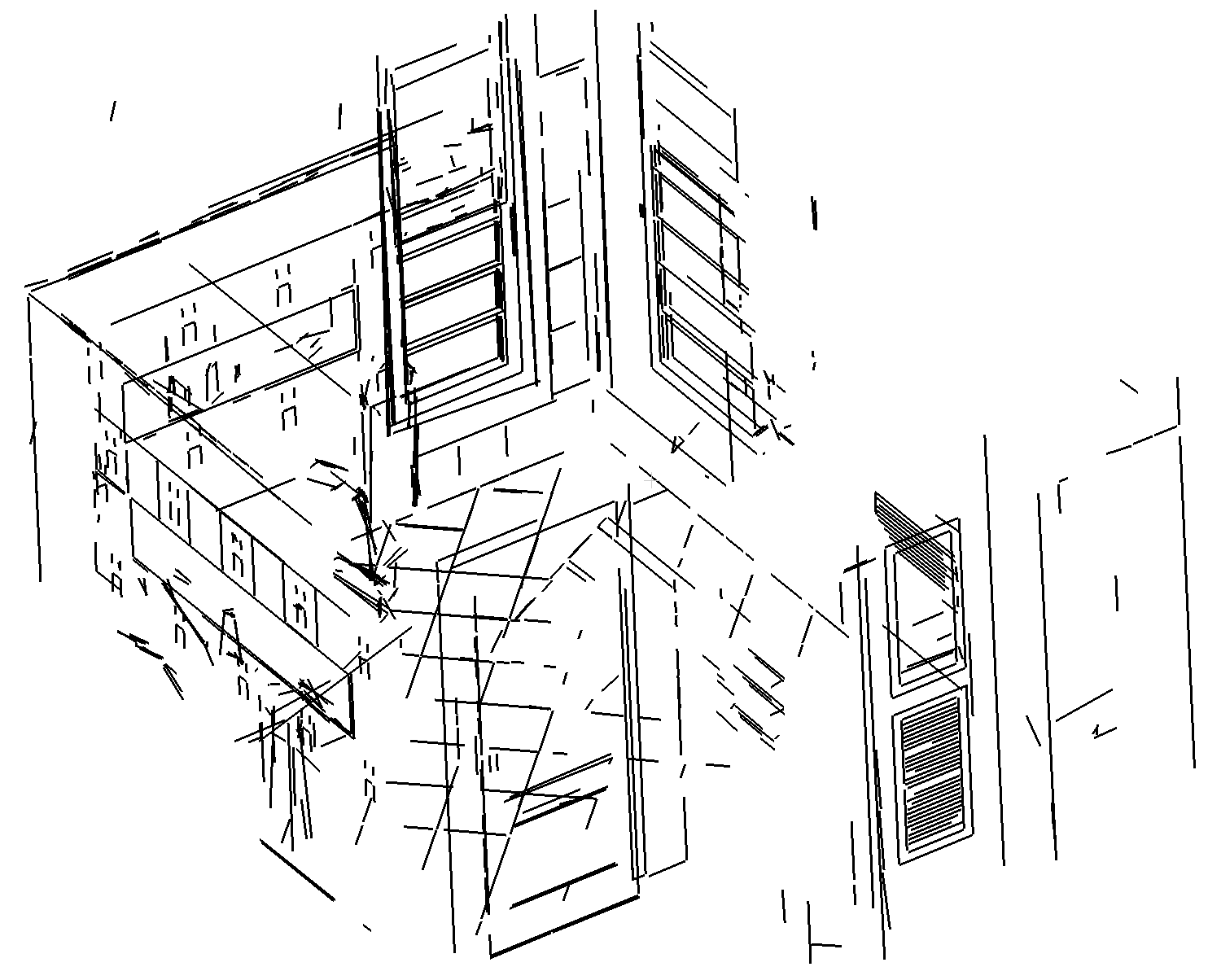} &
        \includegraphics[width=\szp\textwidth]{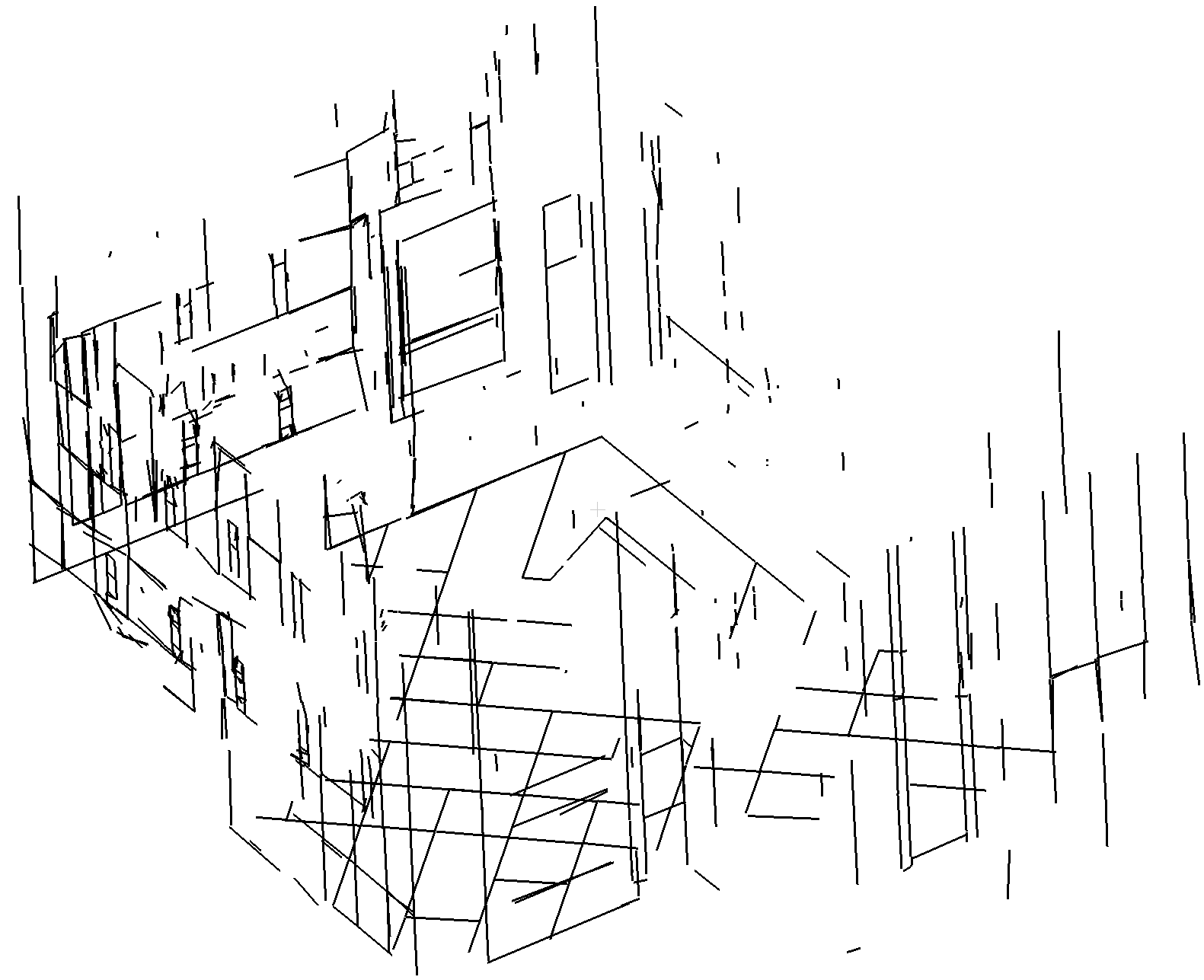} &
        \includegraphics[width=\szp\textwidth]{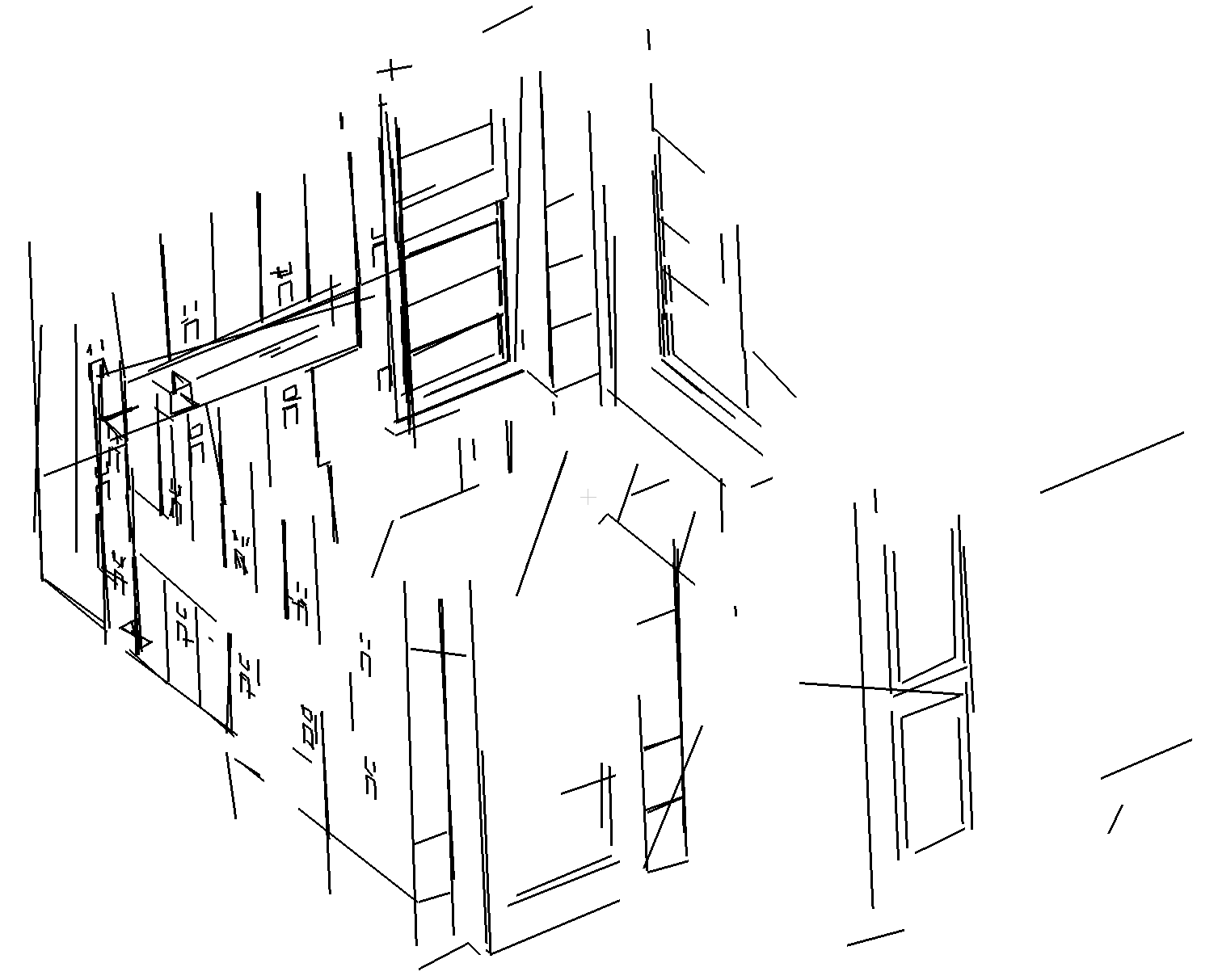} &
        \includegraphics[width=\szp\textwidth]{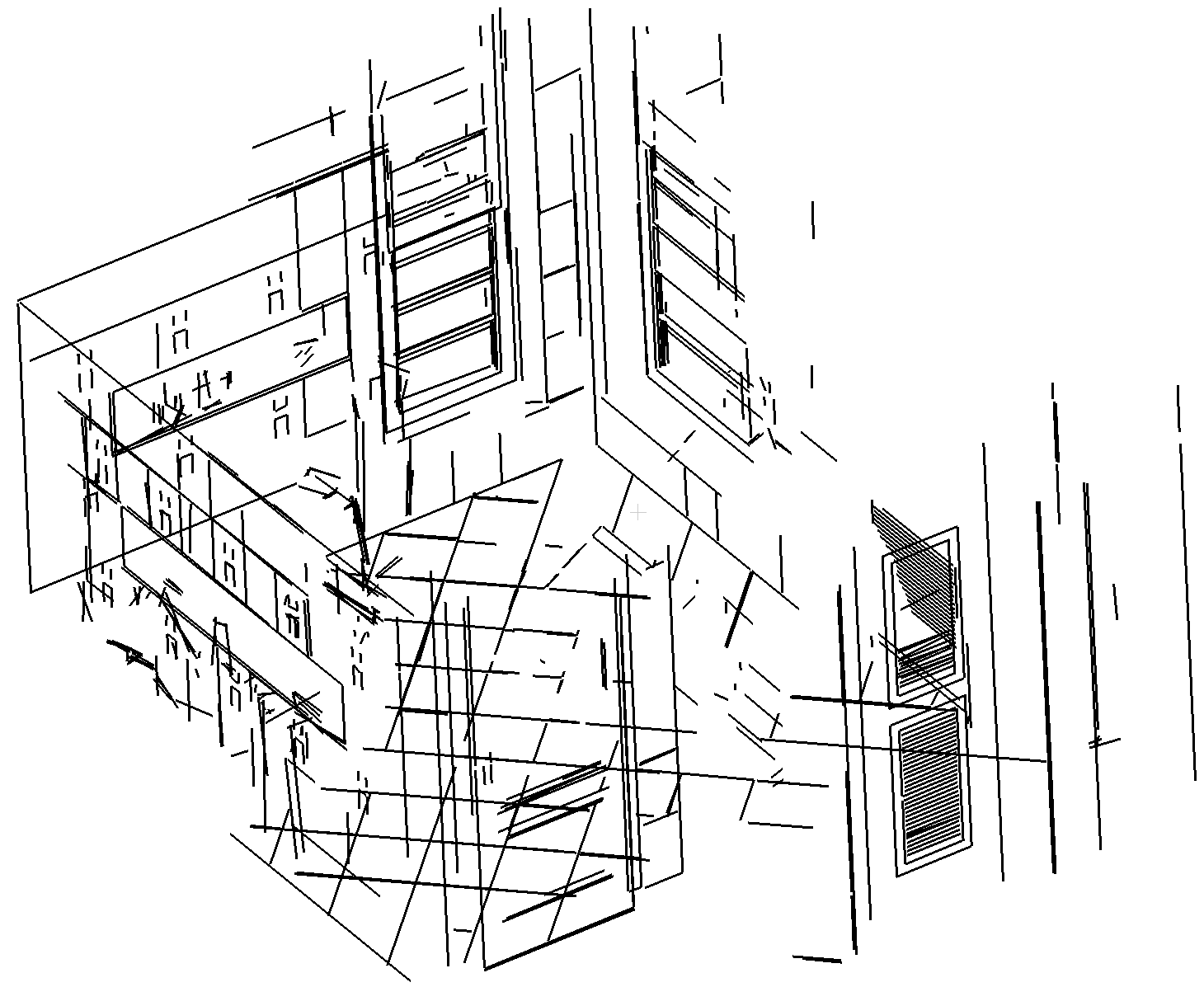} \\
        LSD~\cite{von2008lsd} & SOLD2~\cite{Pautrat_Lin_2021_CVPR} & TP-LSD~\cite{huang2020tp} & DeepLSD (Ours) \\
        \noalign{\vskip 3mm}  \multicolumn{4}{c}{ai\_001\_002} \\
        \includegraphics[width=\szp\textwidth]{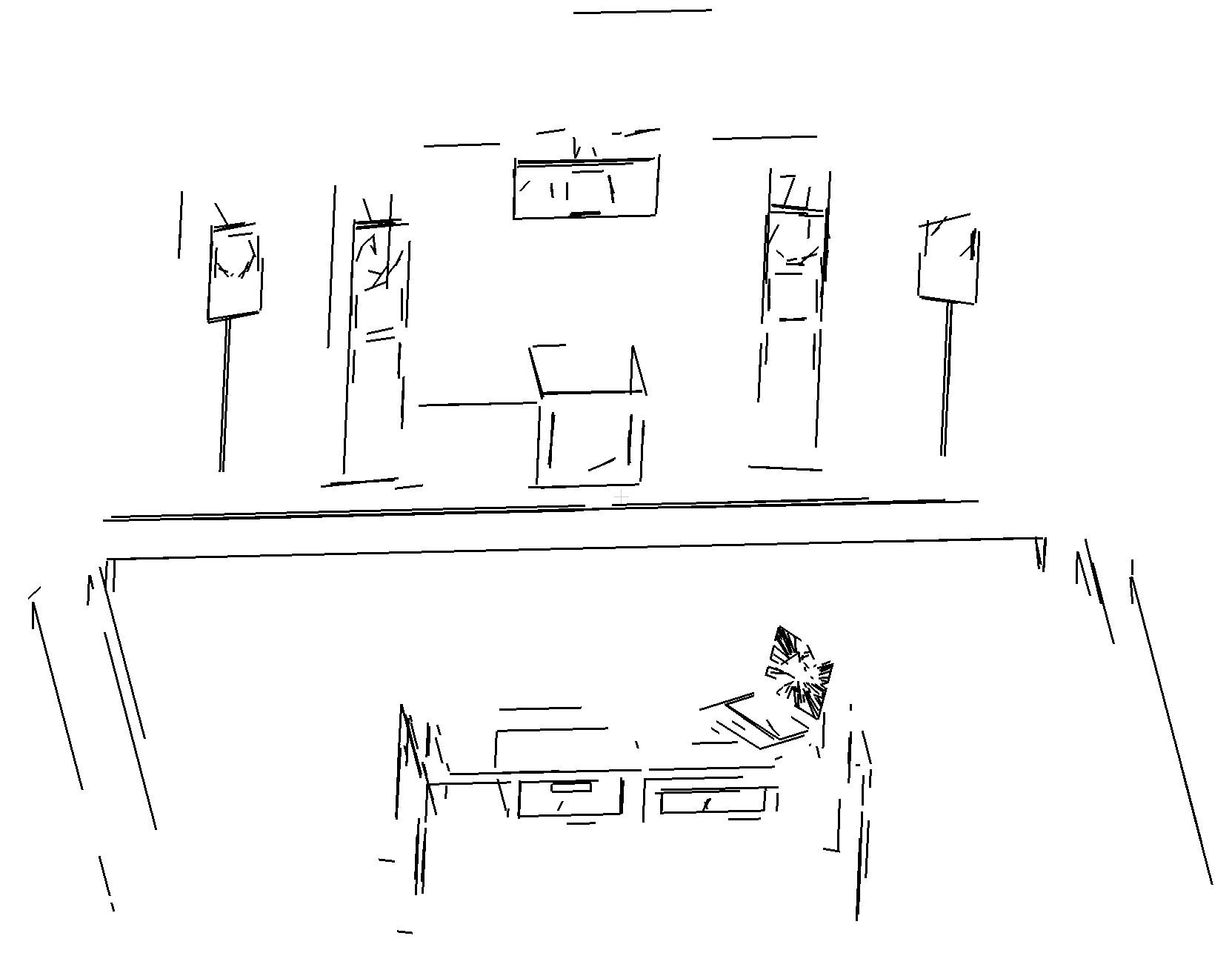} &
        \includegraphics[width=\szp\textwidth]{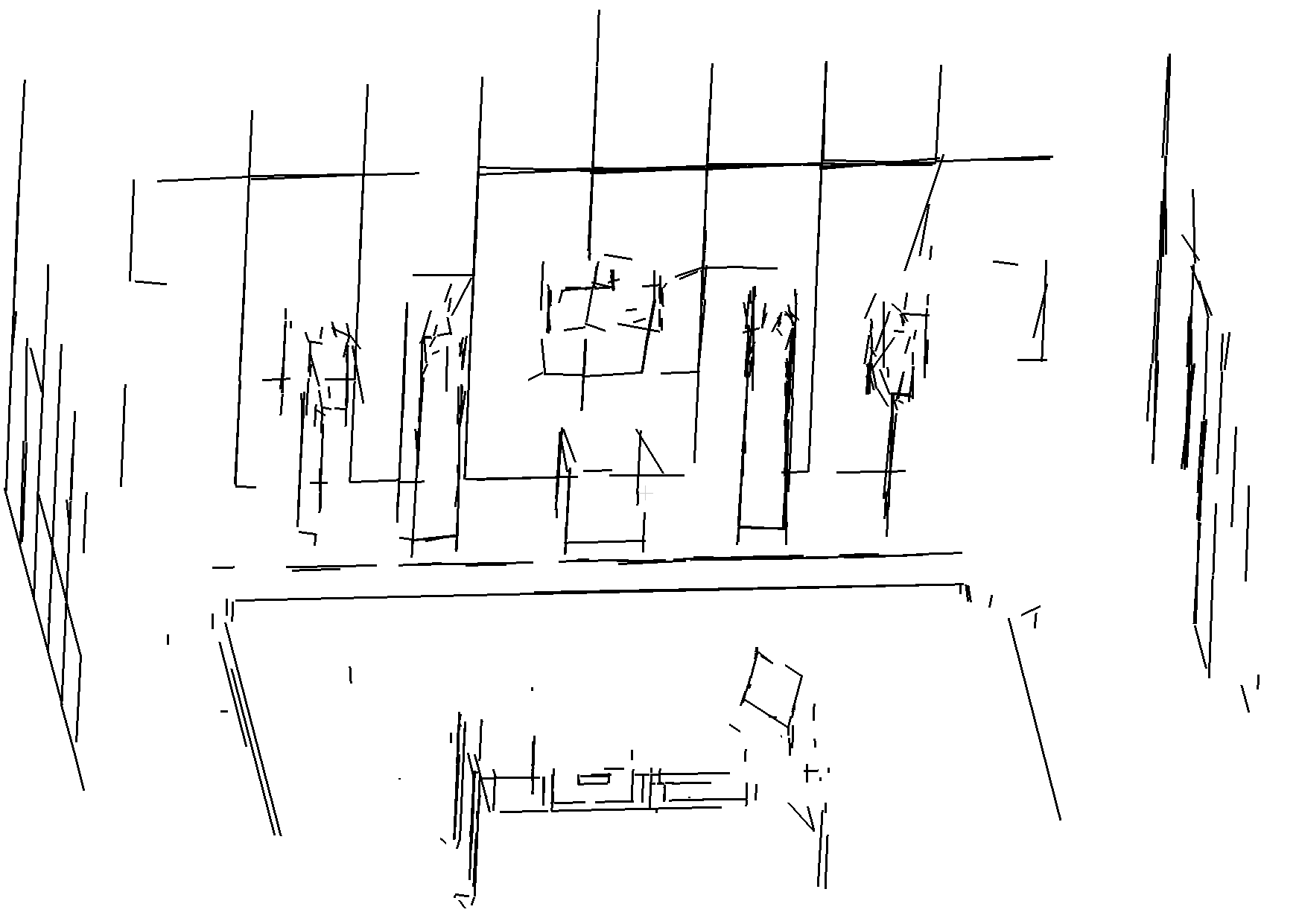} &
        \includegraphics[width=\szp\textwidth]{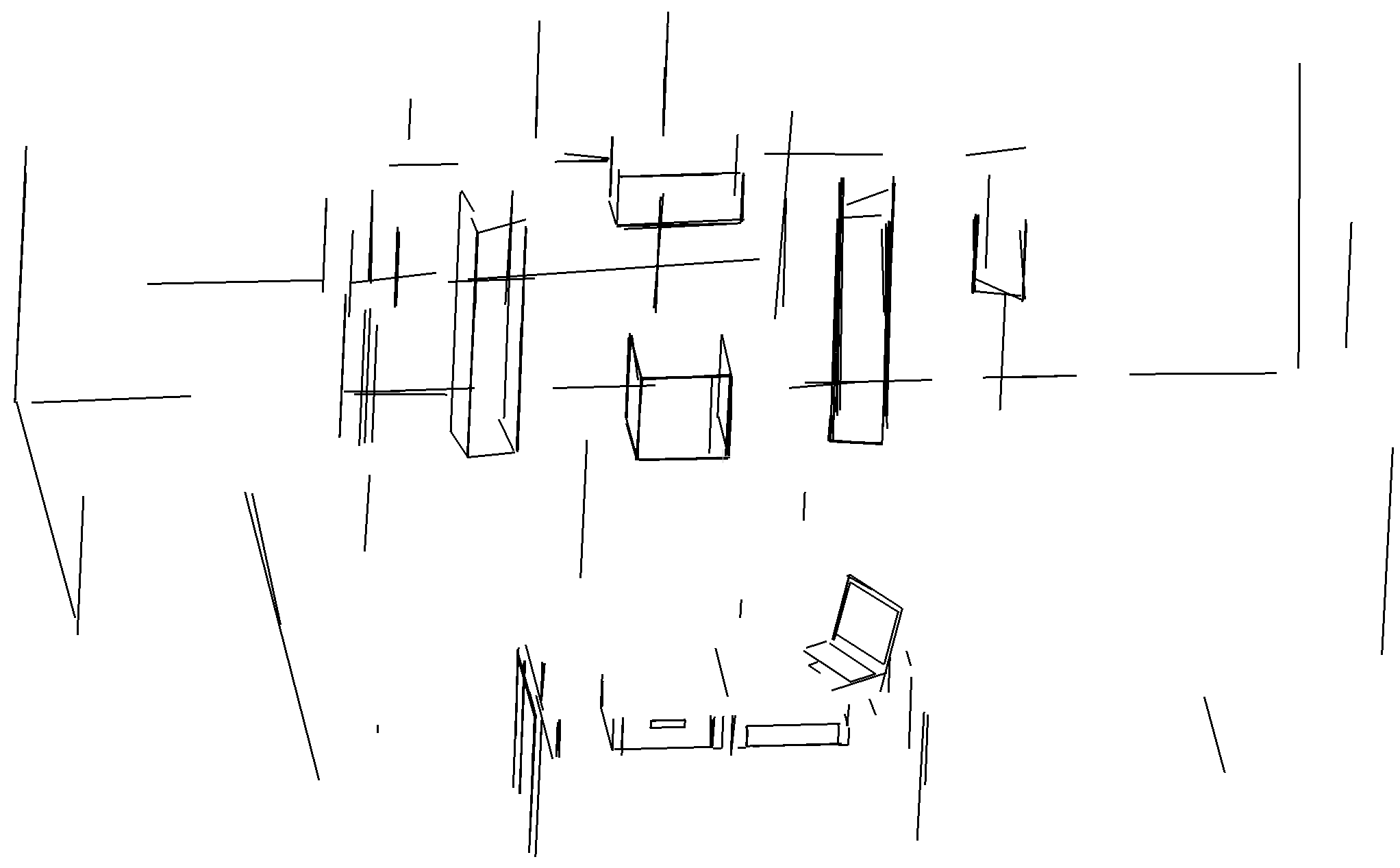} &
        \includegraphics[width=\szp\textwidth]{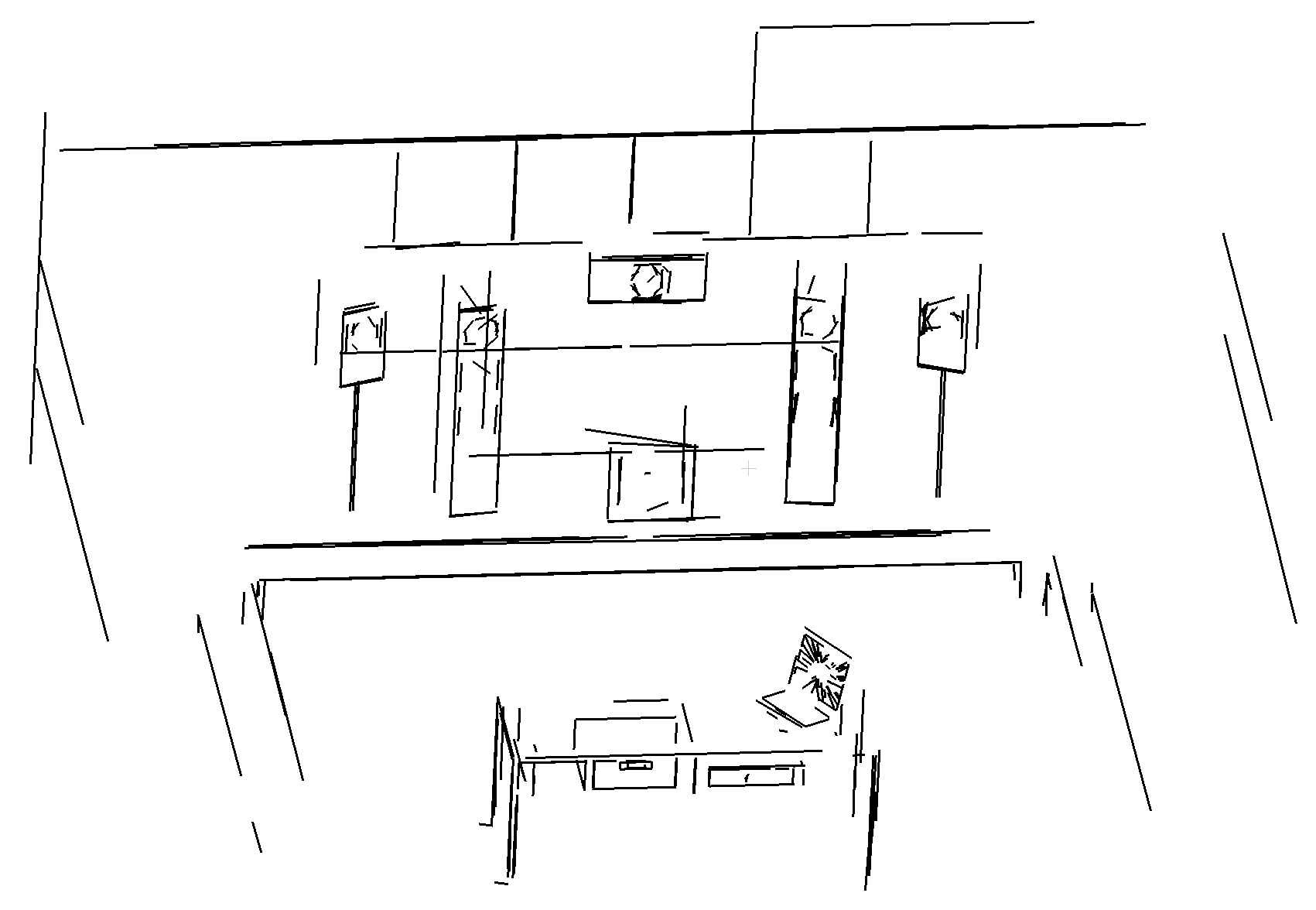} \\
        LSD~\cite{von2008lsd} & SOLD2~\cite{Pautrat_Lin_2021_CVPR} & TP-LSD~\cite{huang2020tp} & DeepLSD (Ours) \\
        \noalign{\vskip 3mm}  \multicolumn{4}{c}{ai\_001\_003} \\
        \includegraphics[width=\szp\textwidth]{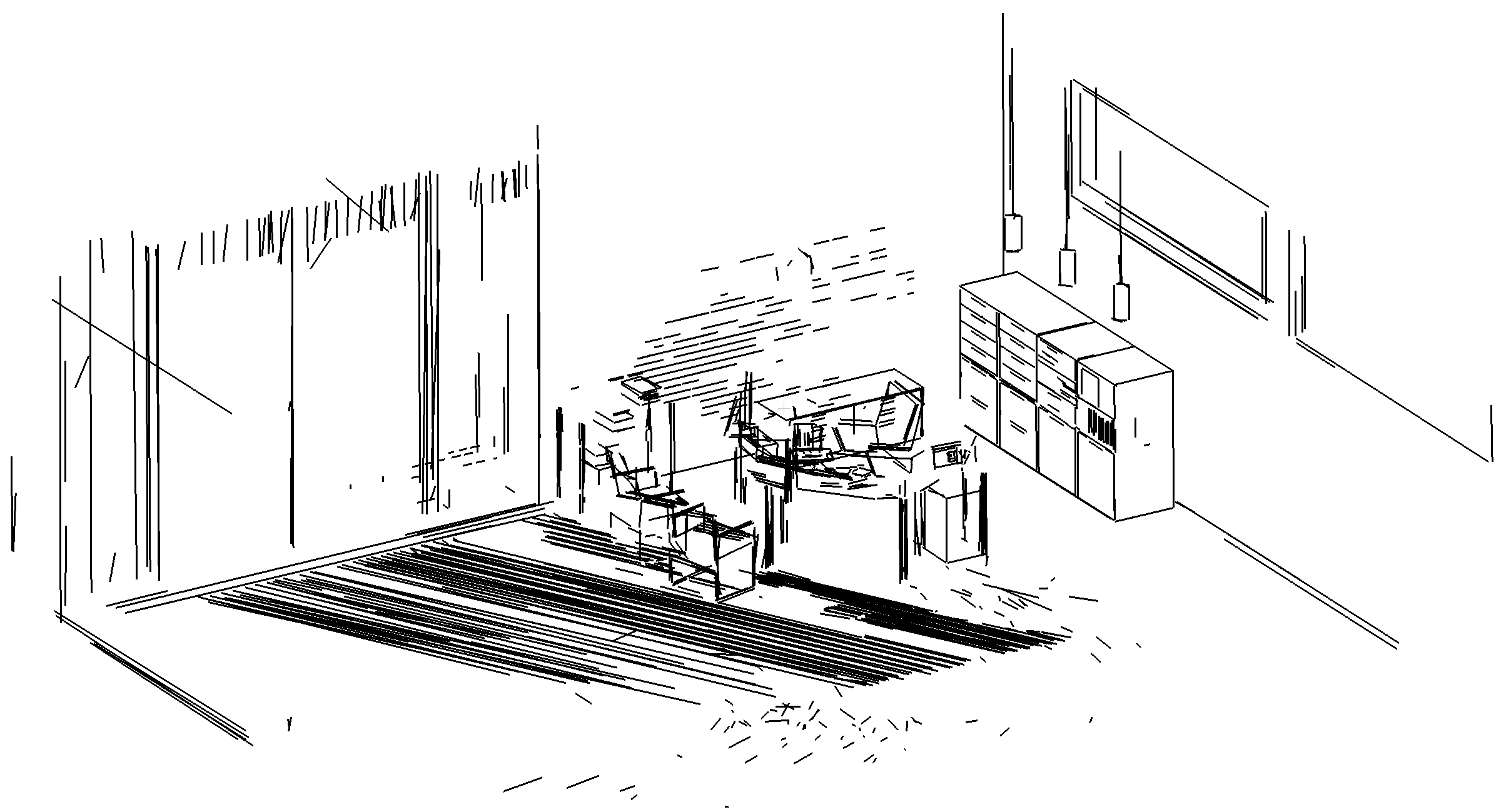} &
        \includegraphics[width=\szp\textwidth]{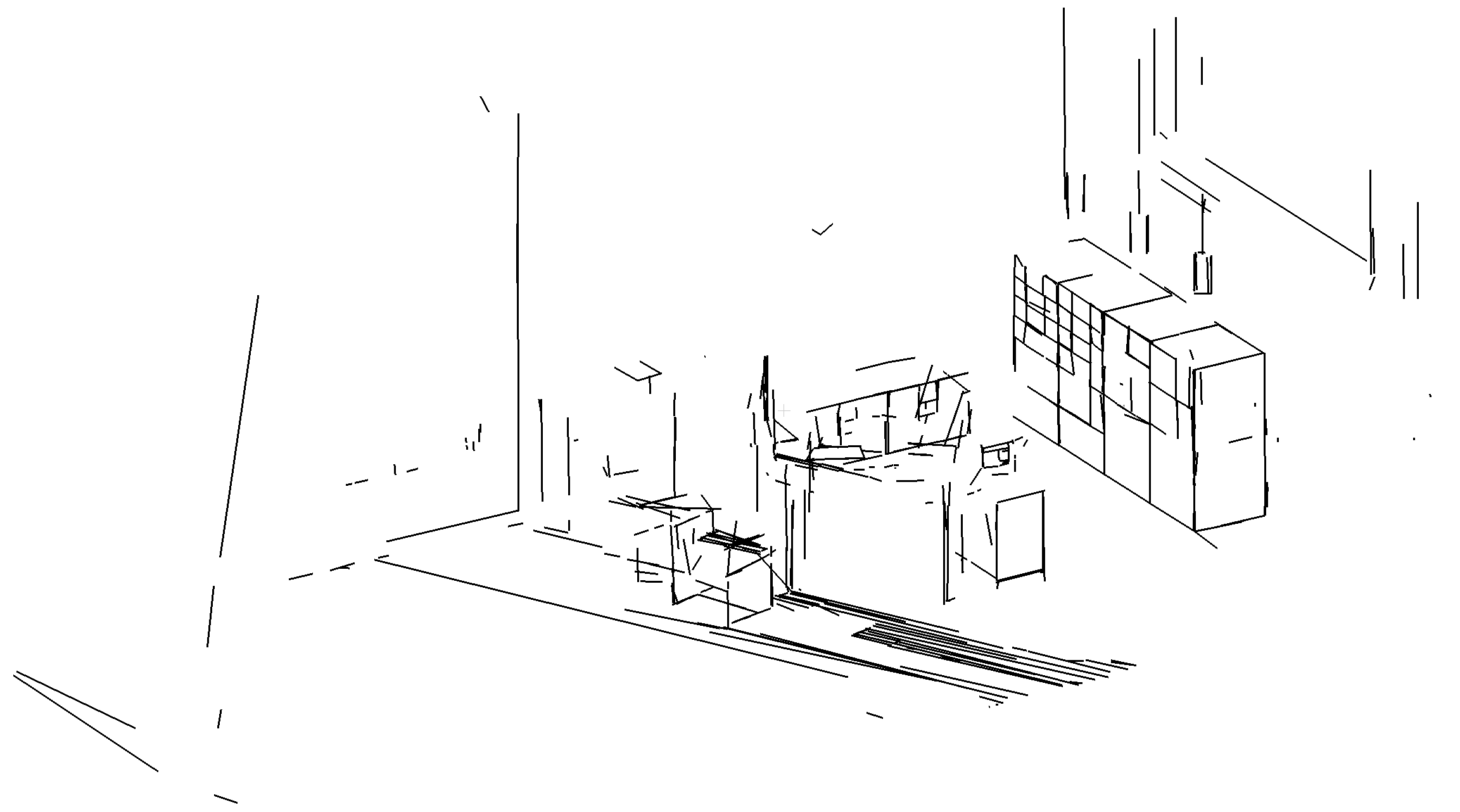} &
        \includegraphics[width=\szp\textwidth]{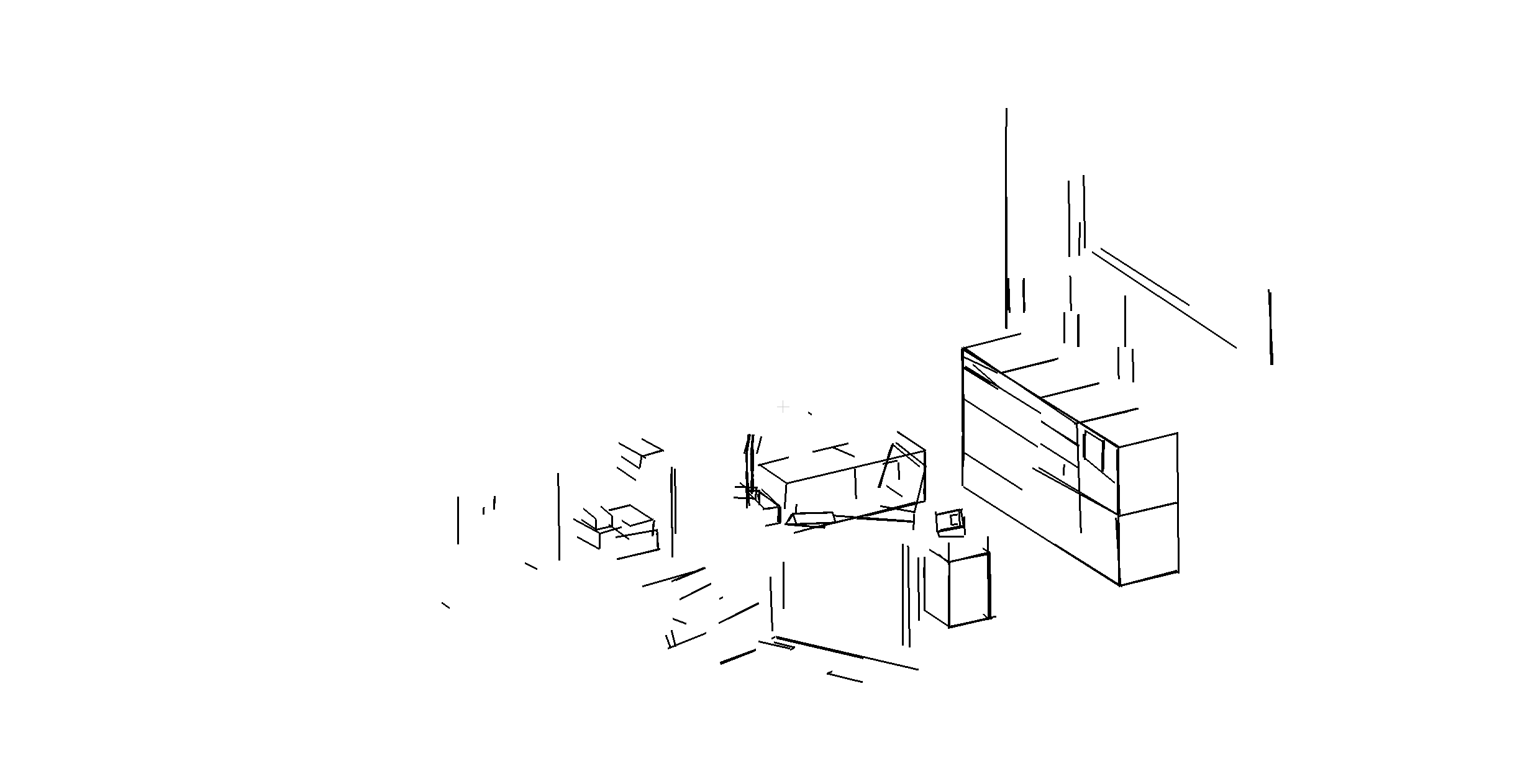} &
        \includegraphics[width=\szp\textwidth]{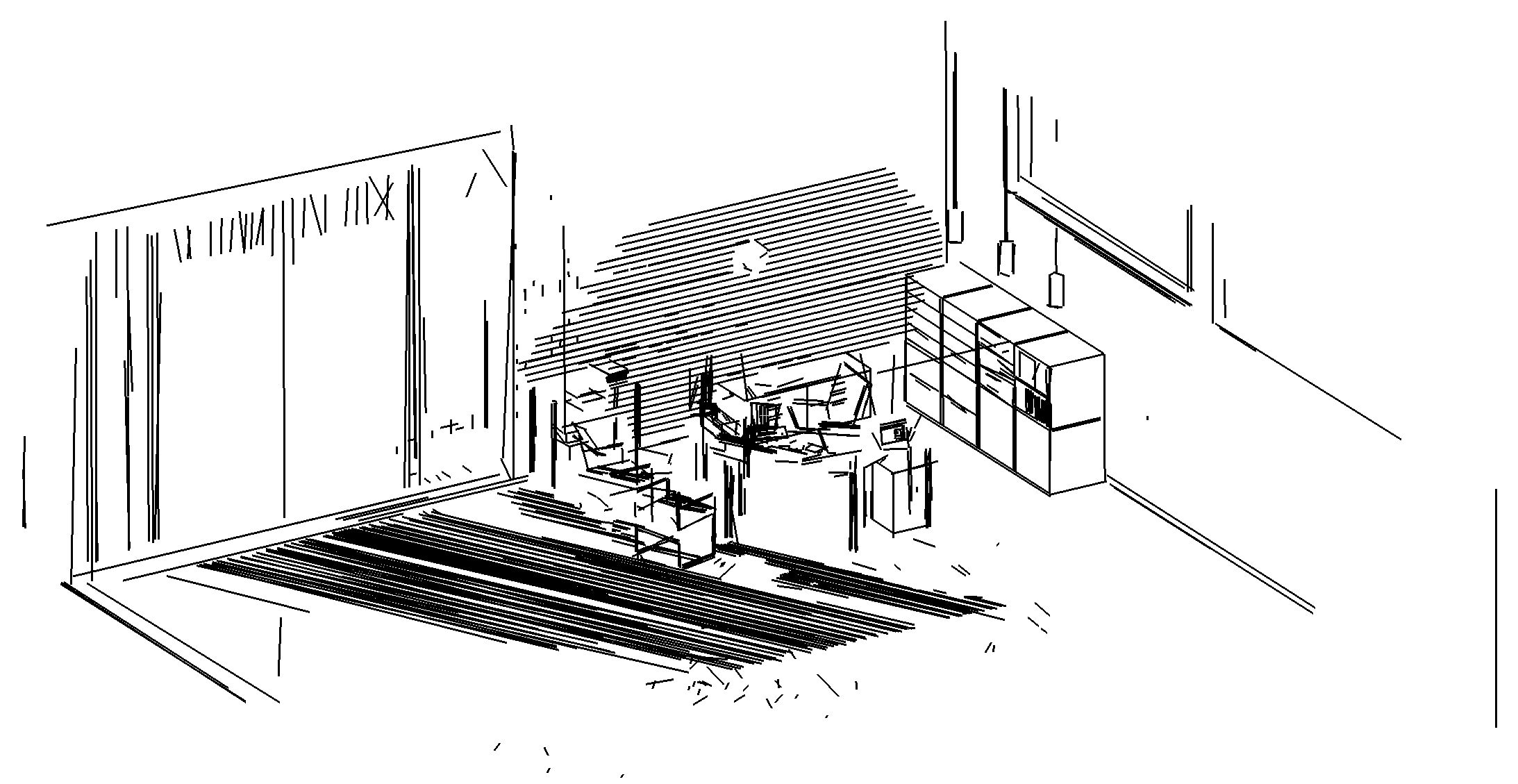} \\
        LSD~\cite{von2008lsd} & SOLD2~\cite{Pautrat_Lin_2021_CVPR} & TP-LSD~\cite{huang2020tp} & DeepLSD (Ours) \\
        \noalign{\vskip 3mm}  \multicolumn{4}{c}{ai\_001\_004} \\
        \includegraphics[width=\szp\textwidth]{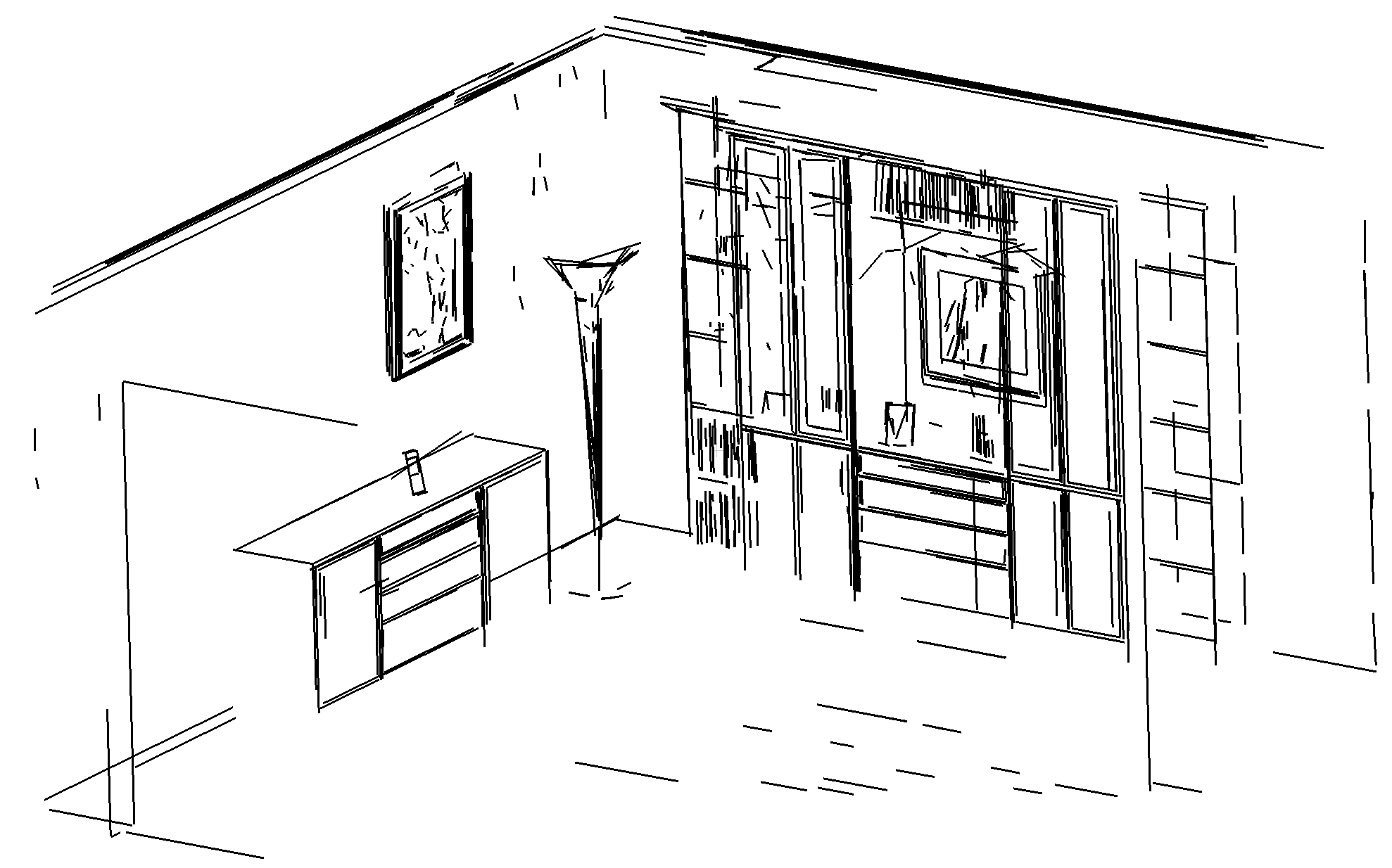} &
        \includegraphics[width=\szp\textwidth]{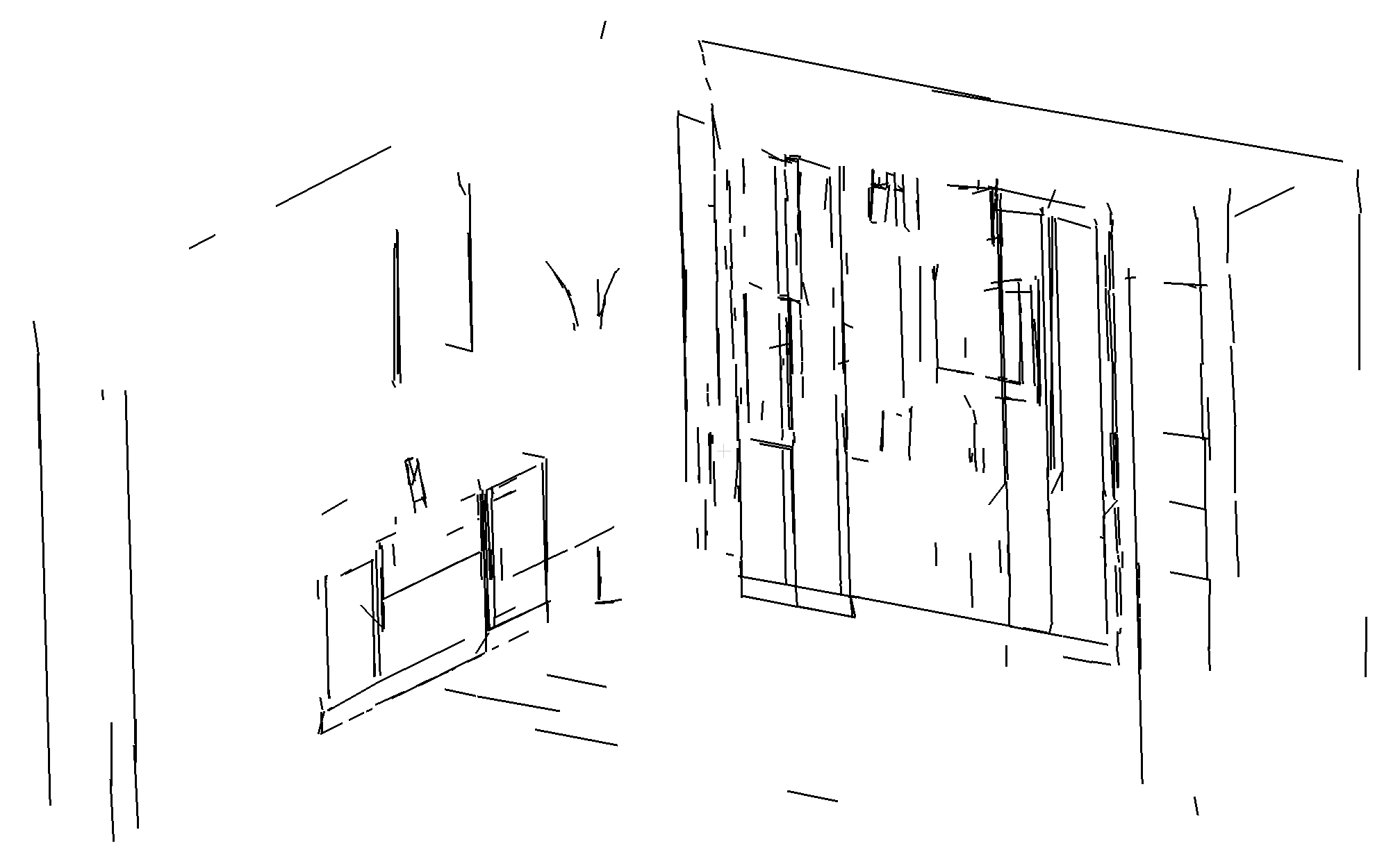} &
        \includegraphics[width=\szp\textwidth]{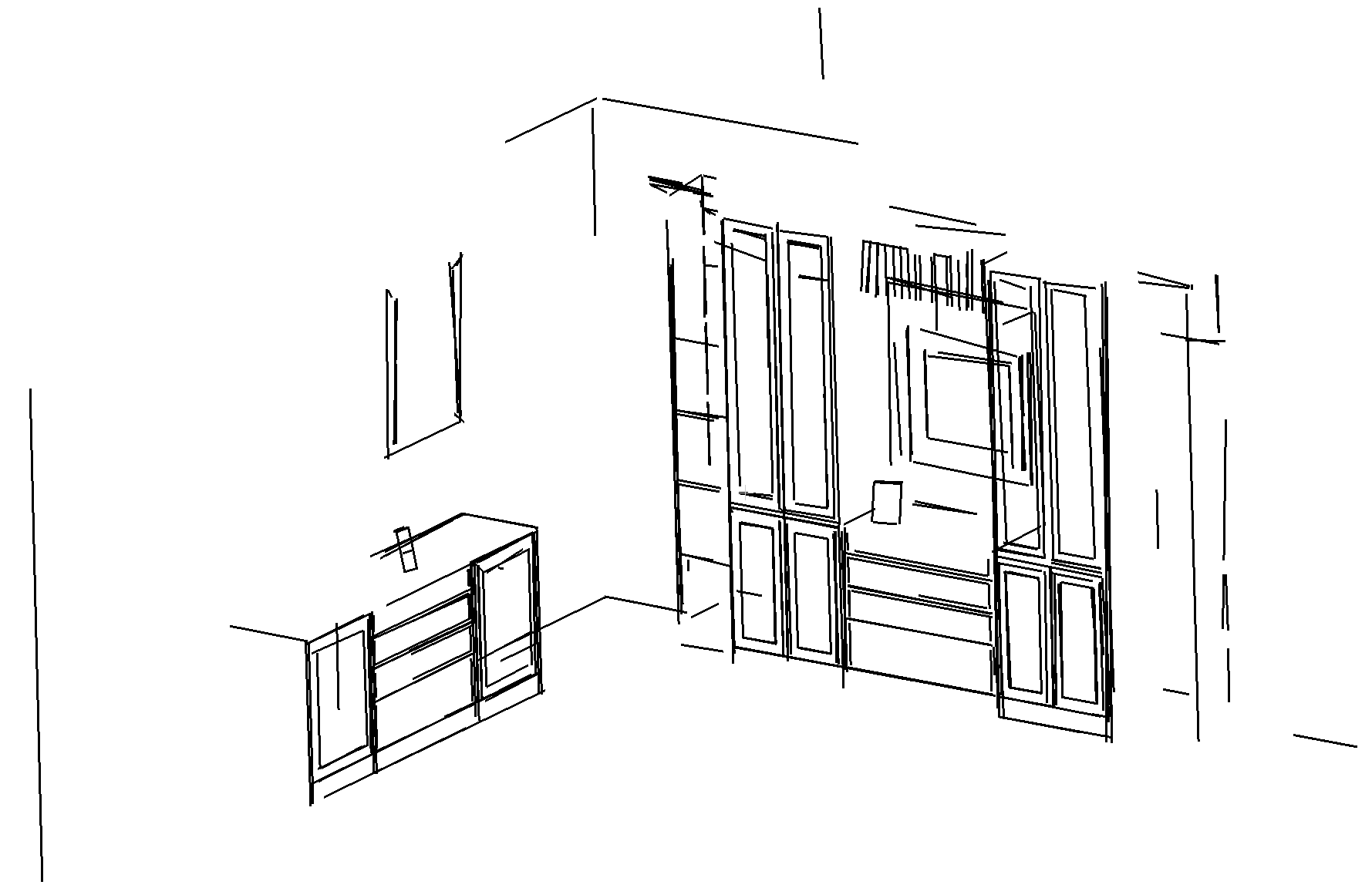} &
        \includegraphics[width=\szp\textwidth]{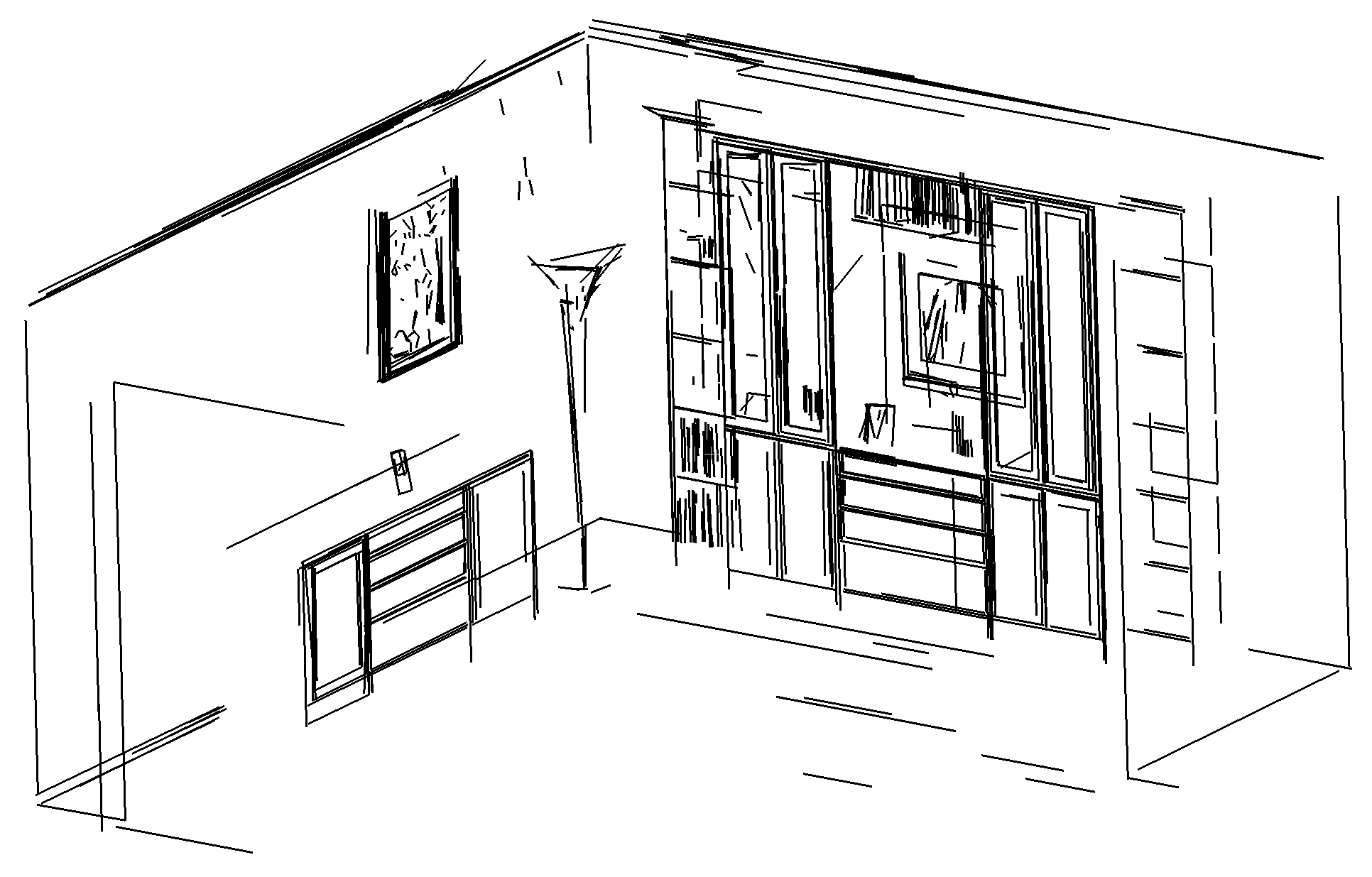} \\
        LSD~\cite{von2008lsd} & SOLD2~\cite{Pautrat_Lin_2021_CVPR} & TP-LSD~\cite{huang2020tp} & DeepLSD (Ours) \\[-5pt]
    \end{tabular}
    \caption{\textbf{Line 3D reconstruction on Hypersim~\cite{roberts_2021}.} We leverage the line 3D mapping software Line3D++~\cite{hofer_2017} on the first 4 scenes of Hypersim~\cite{roberts_2021}. DeepLSD produces more complete and accurate reconstructions than all baselines.}
    \label{fig:3d_reconstruction}
\end{figure*}

%
We show here in Figure~\ref{fig:3d_reconstruction} a qualitative comparison of the 3D line reconstructions of our lines and some baselines for the first 4 scenes of the Hypersim dataset~\cite{roberts_2021}.
TP-LSD~\cite{huang2020tp} can reconstruct fewer lines as it is trained on wireframe lines only and cannot recover subtle details of the scene.
While LSD~\cite{von2008lsd} is usually the traditional detector being used for 3D reconstruction~\cite{hofer_2017}, the reconstructions produced by DeepLSD are overall more complete and the lines are cleaner compared to the LSD reconstruction. In addition, LSD has a tendency to break segments on higher resolution images, while DeepLSD will detect longer and cleaner lines. 
Thus, it is easy to merge all lines of a track into a nice long 3D line for DeepLSD, while LSD will generate a collection of dissociated small segments along the 3D line.

\section{Limitations} \label{sec:limitations}
%
Even though DeepLSD can produce repeatable and accurate lines by taking advantage of the benefits of both traditional and learned methods, it still suffers from a few limitations:
\begin{itemize}
    \item The current approach of running a deep network, followed by handcrafted heuristics and line optimization is not fully differentiable. Making the full pipeline differentiable would mean making LSD differentiable, which is unclear how to do it. We plan to investigate this further in the future, as an end-to-end pipeline would certainly provide better training signals to the deep network processing the image.
    \item The generation of the pseudo ground truth lines is still limited by the performance of LSD~\cite{von2008lsd}. If a line is almost never detected by LSD during homography adaptation, it will most likely not be detected in the ground truth attraction field. Similarly, a noisy but repeatable line will be kept in the pseudo ground truth. One way to overcome this issue could be to leverage the trained DeepLSD to re-generate a new pseudo ground truth with less noise, as was done in SuperPoint~\cite{superpoint}.
    \item In spite of our efforts to make the pseudo ground truth as clean as possible, there is always a trade-off between detecting all low-contrast lines and avoiding to detect noisy lines in the background. For example, DeepLSD misses some good lines at the bottom right of the image in the 5th row of Figure~\ref{fig:supp_visualizations} and is also detecting some noisy lines in the sky of the image in the 7th row. We can influence this trade-off in two ways. First, by tuning the aggregation of the attraction field when generating the ground truth. We currently take the median value of the distance and angle fields, but one could also take a given percentile, to allow more or less outlier values. Second, one can enforce more or less constraints to the distance field for background areas. Enforcing a high distance field for pixels far away from the ground truth lines will reduce the number of noisy lines in the background, but will also ignore the lines with low contrast. The parameters proposed in this paper are the ones visually yielding the best trade-off between the two.
    \item Though the input image is processed through a deep network, there is still no proper semantic understanding of the detected lines, so that DeepLSD will detect any kind of lines. Depending on the application, one could imagine adding some semantic filtering in the ground truth generation to keep only a specific kind of lines (e.g. avoiding lines in the sky or on dynamic objects such as humans).
    \item The proposed line refinement is for now rather slow, especially when it is applied to other deep line detectors, as it requires running two networks. However, we believe that it is still valuable for applications that can run offline and that require high precision, such as for 3D reconstruction. Our current implementation can also certainly be optimized, and our network compressed to run on embedded devices, without sacrificing too much performance.
\end{itemize}

\section{Additional Visualizations}  \label{sec:visualizations}
\begin{figure*}[tb]
    \centering
    \scriptsize
    \setlength{\tabcolsep}{2pt}
    \begin{tabular}{ccccc}
        HAWP~\cite{hawp} & TP-LSD~\cite{huang2020tp} & SOLD2~\cite{Pautrat_Lin_2021_CVPR} & LSD~\cite{von2008lsd} & DeepLSD (Ours) \\
        \includegraphics[width=0.19\textwidth]{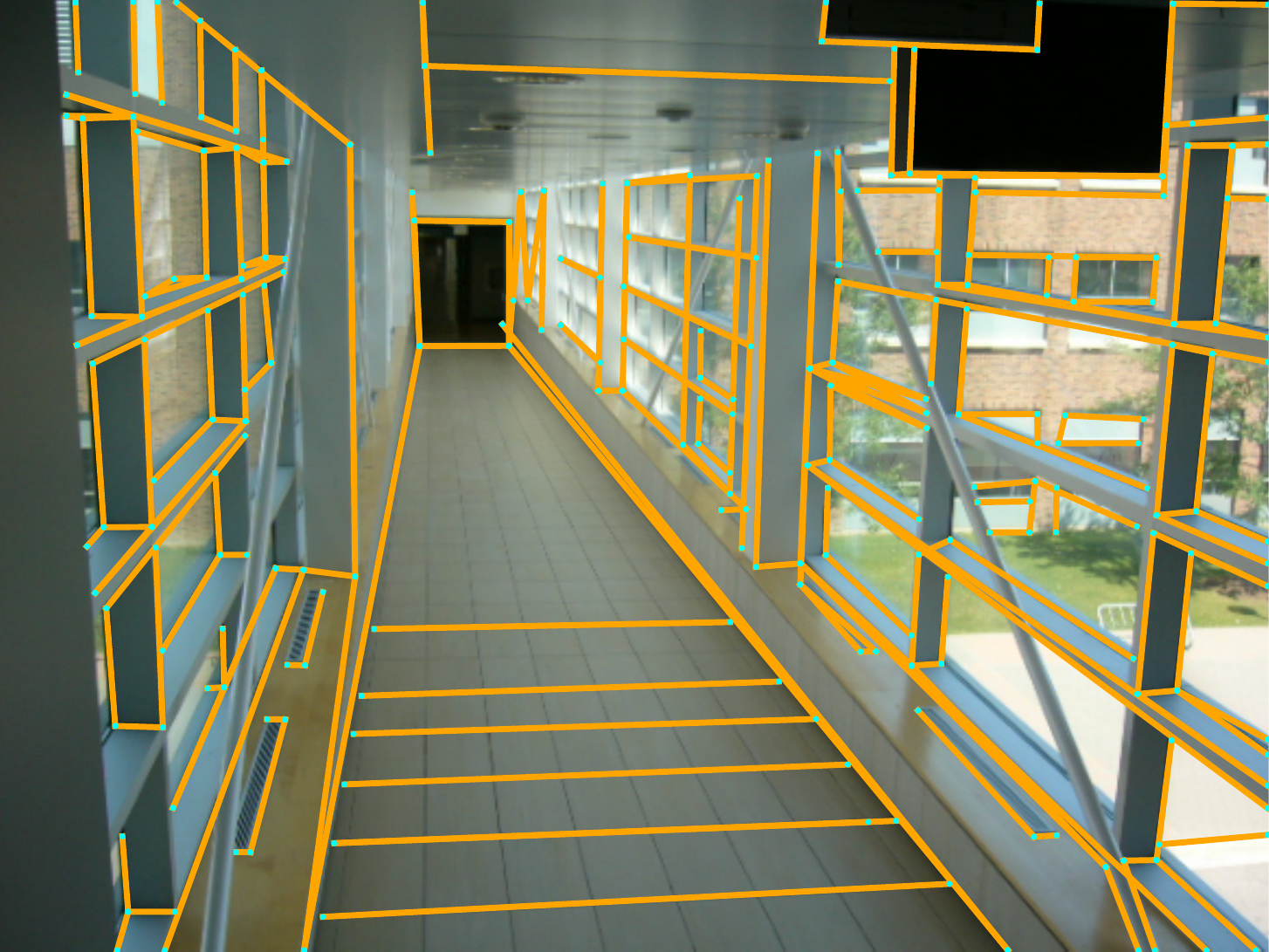}
        & \includegraphics[width=0.19\textwidth]{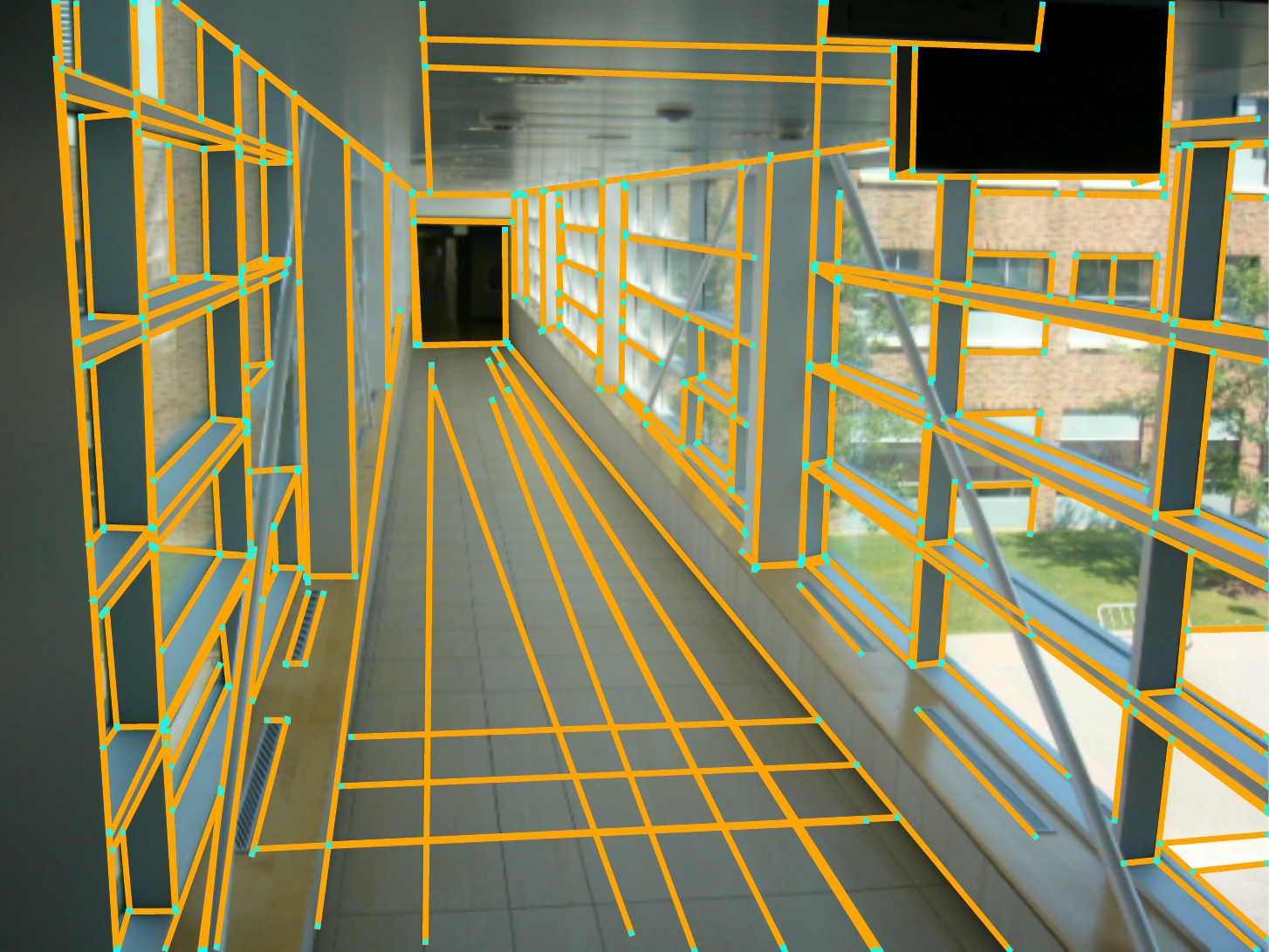}
        & \includegraphics[width=0.19\textwidth]{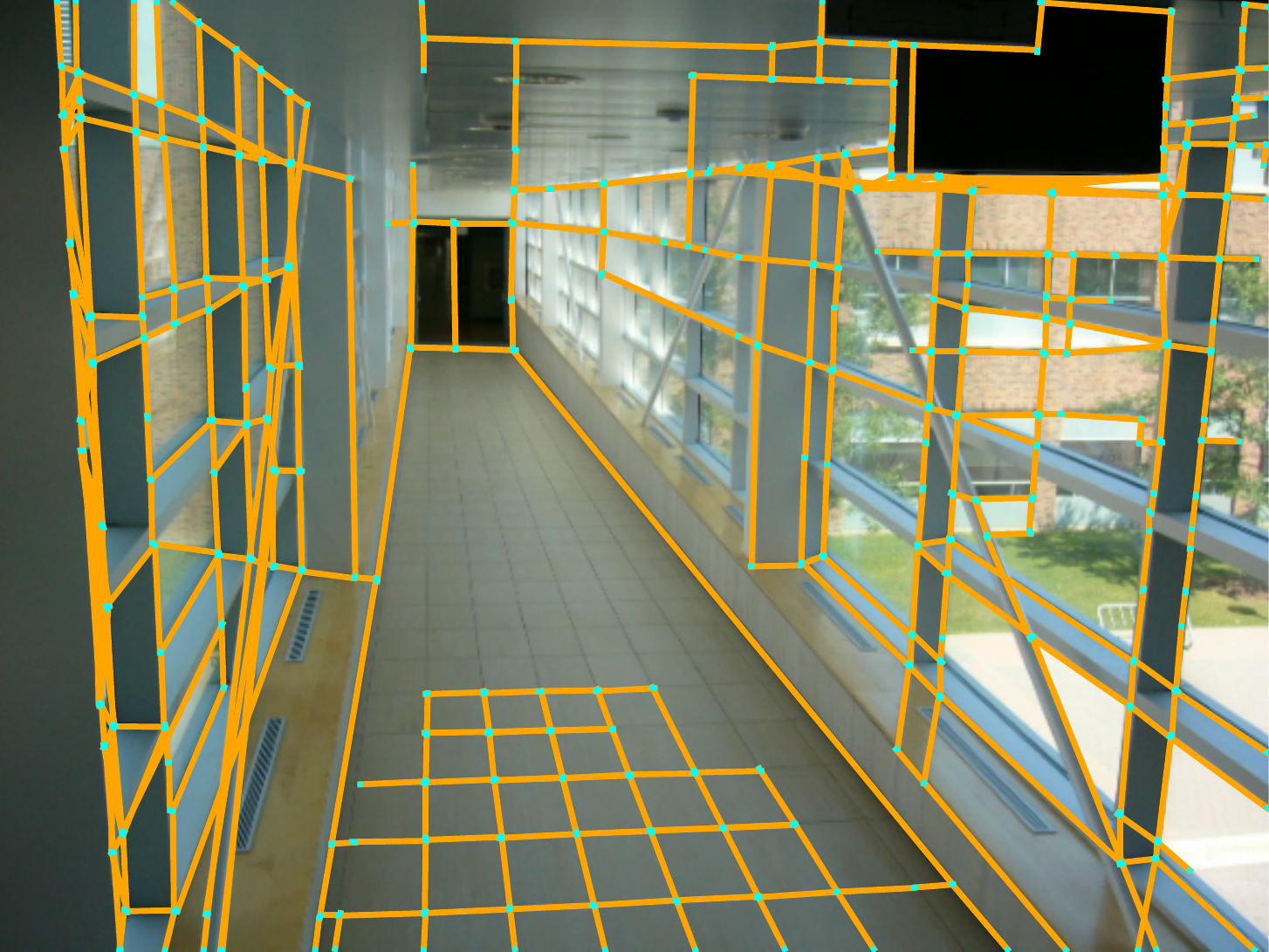}
        & \includegraphics[width=0.19\textwidth]{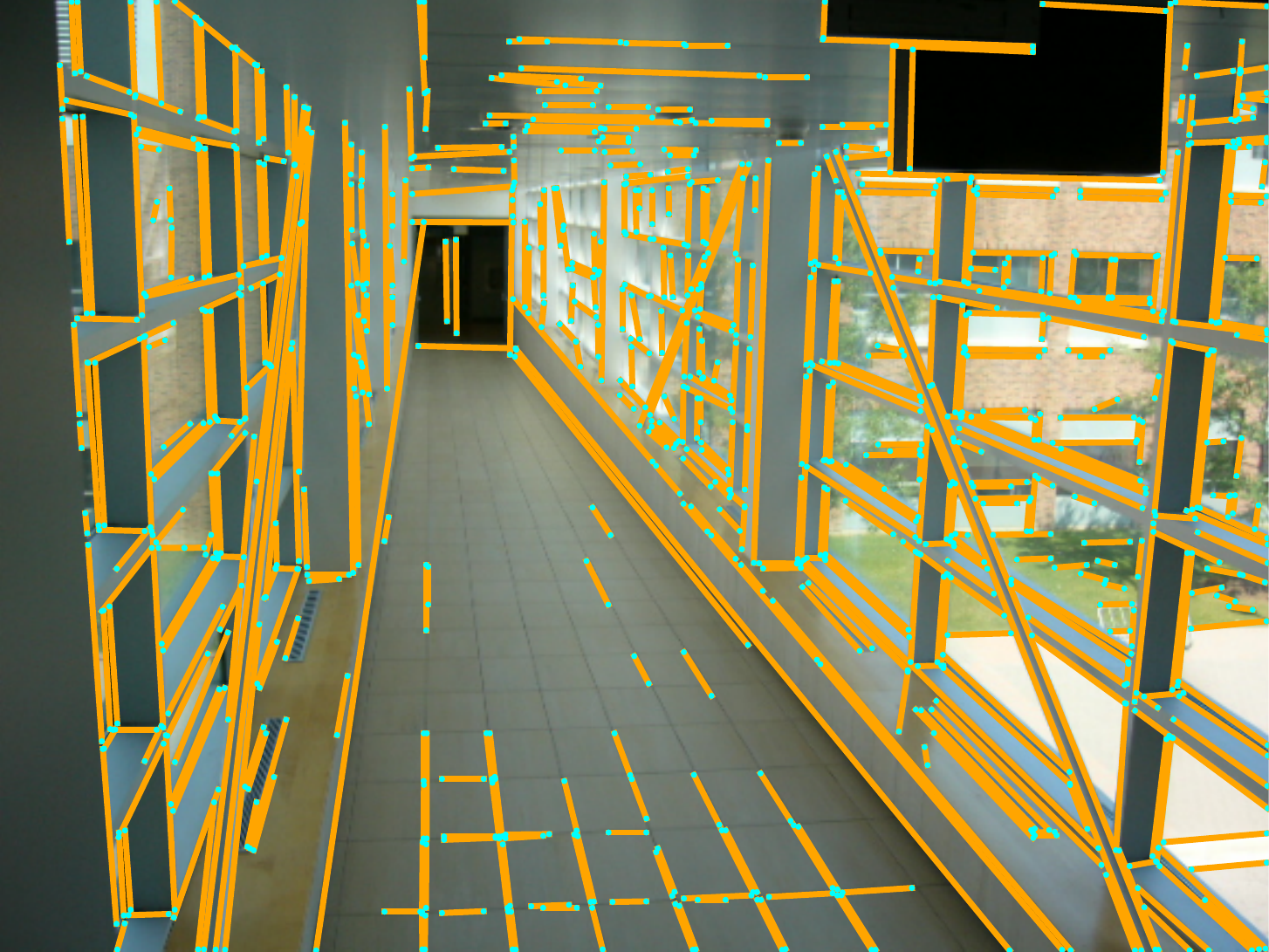}
        & \includegraphics[width=0.19\textwidth]{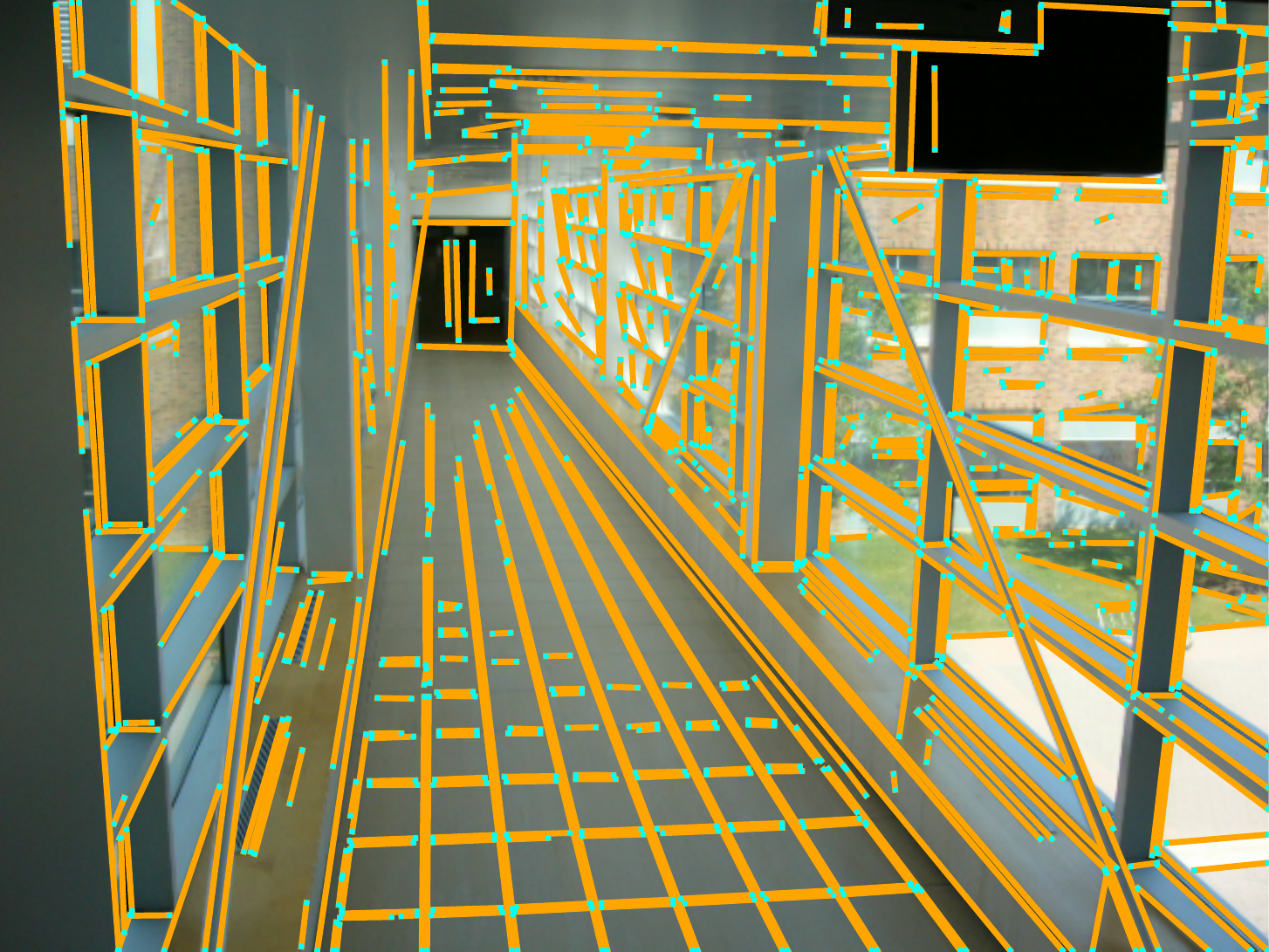} \\
        \includegraphics[width=0.19\textwidth]{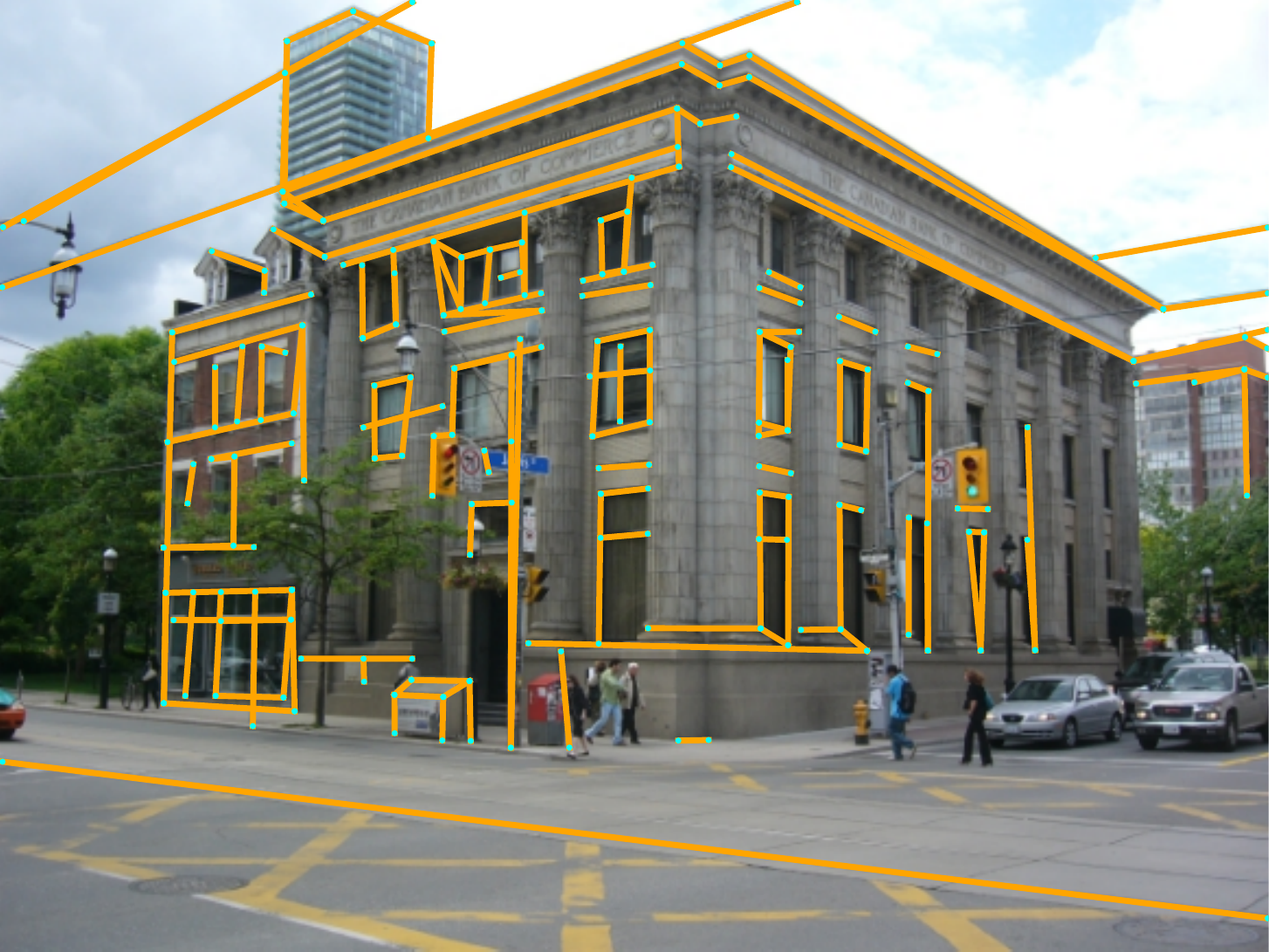}
        & \includegraphics[width=0.19\textwidth]{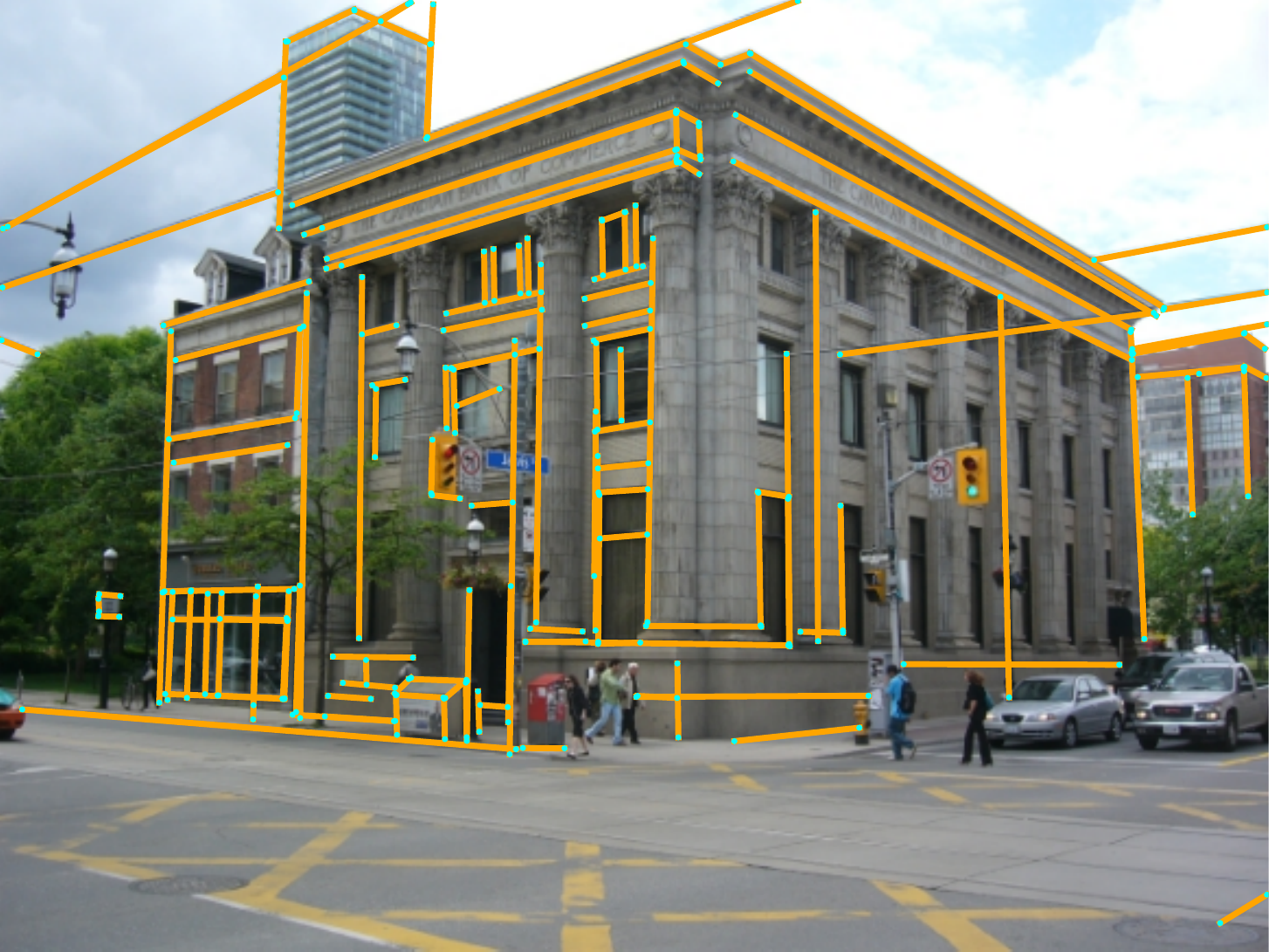}
        & \includegraphics[width=0.19\textwidth]{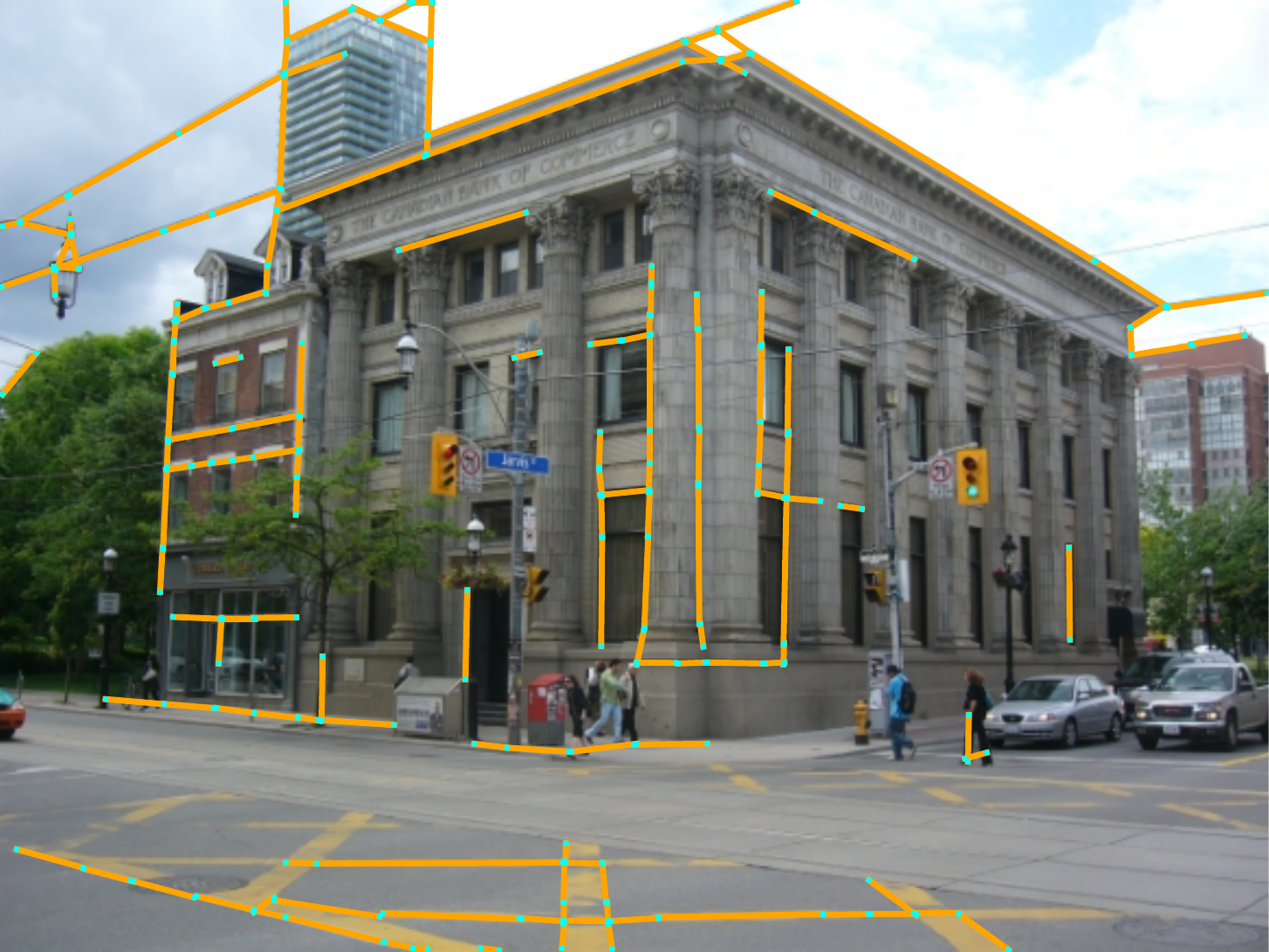}
        & \includegraphics[width=0.19\textwidth]{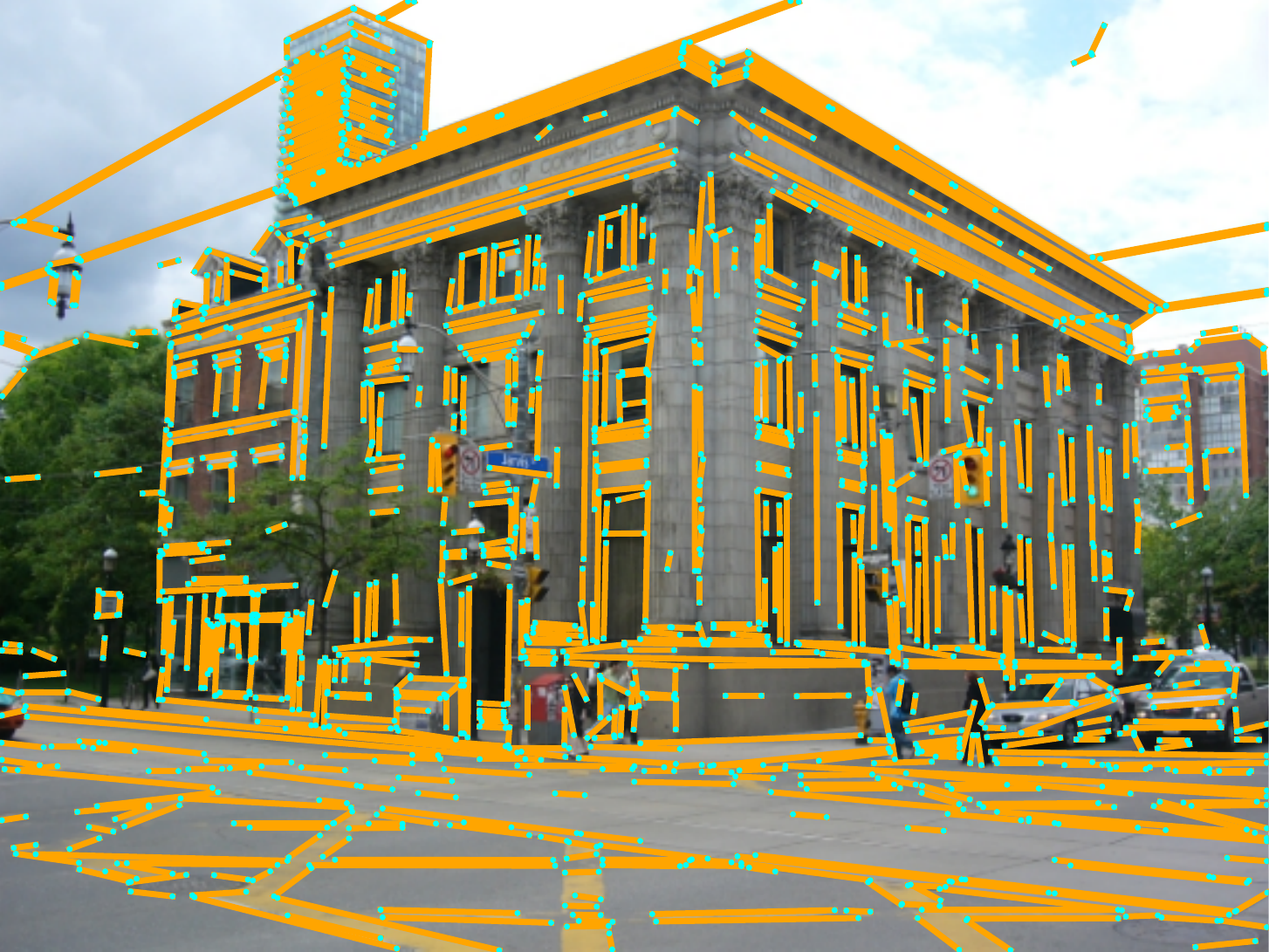}
        & \includegraphics[width=0.19\textwidth]{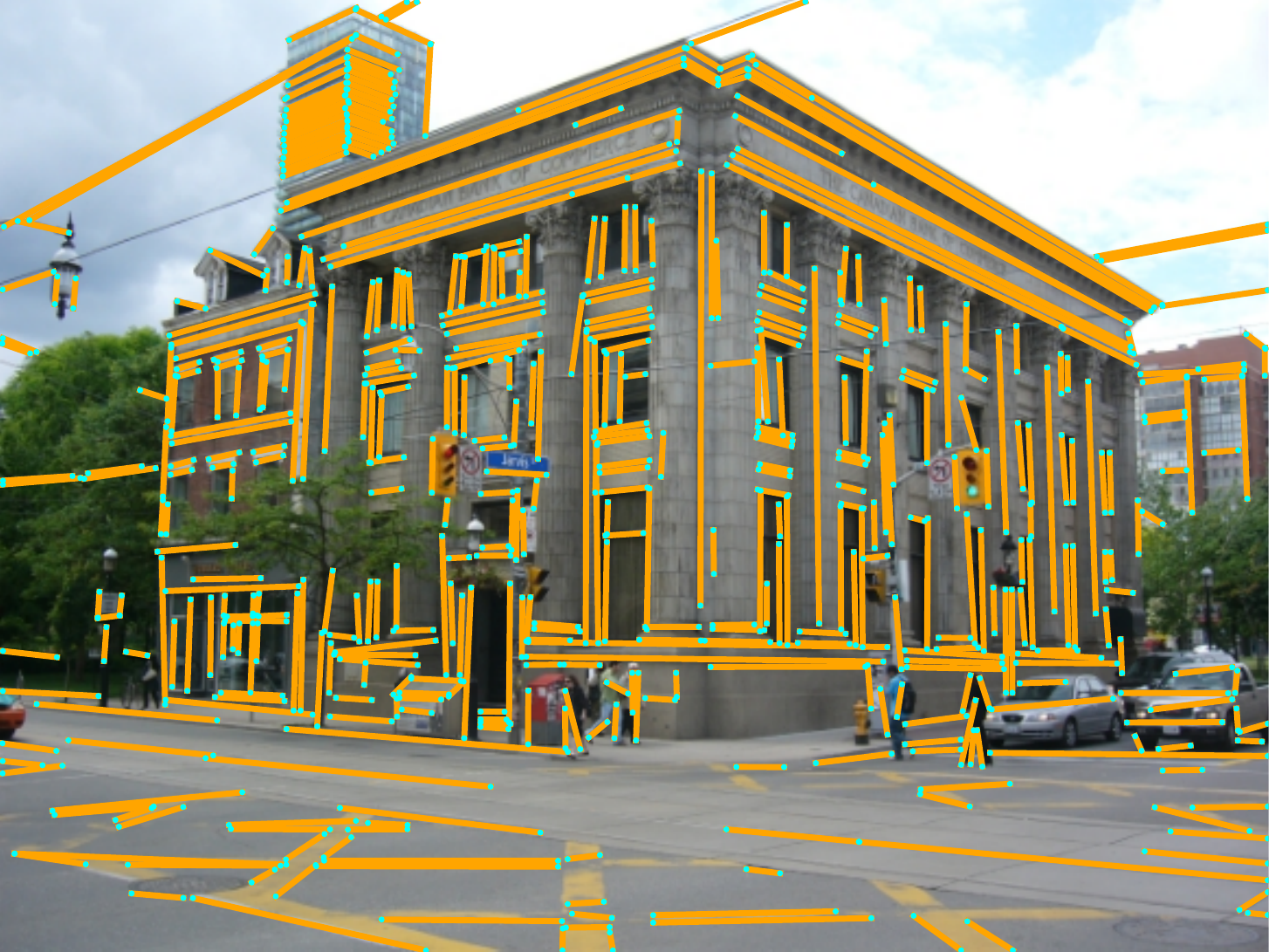} \\
        \includegraphics[width=0.19\textwidth]{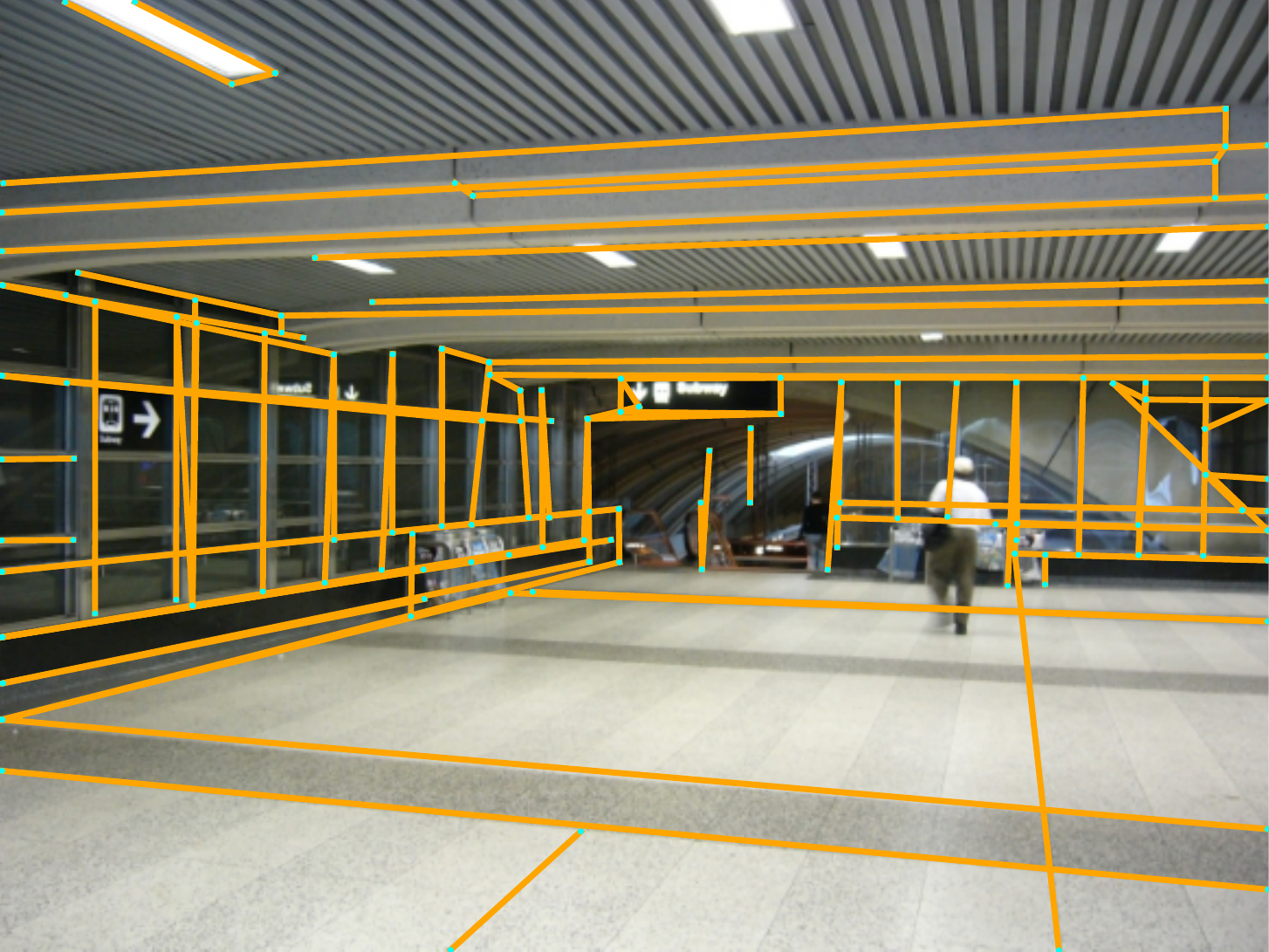}
        & \includegraphics[width=0.19\textwidth]{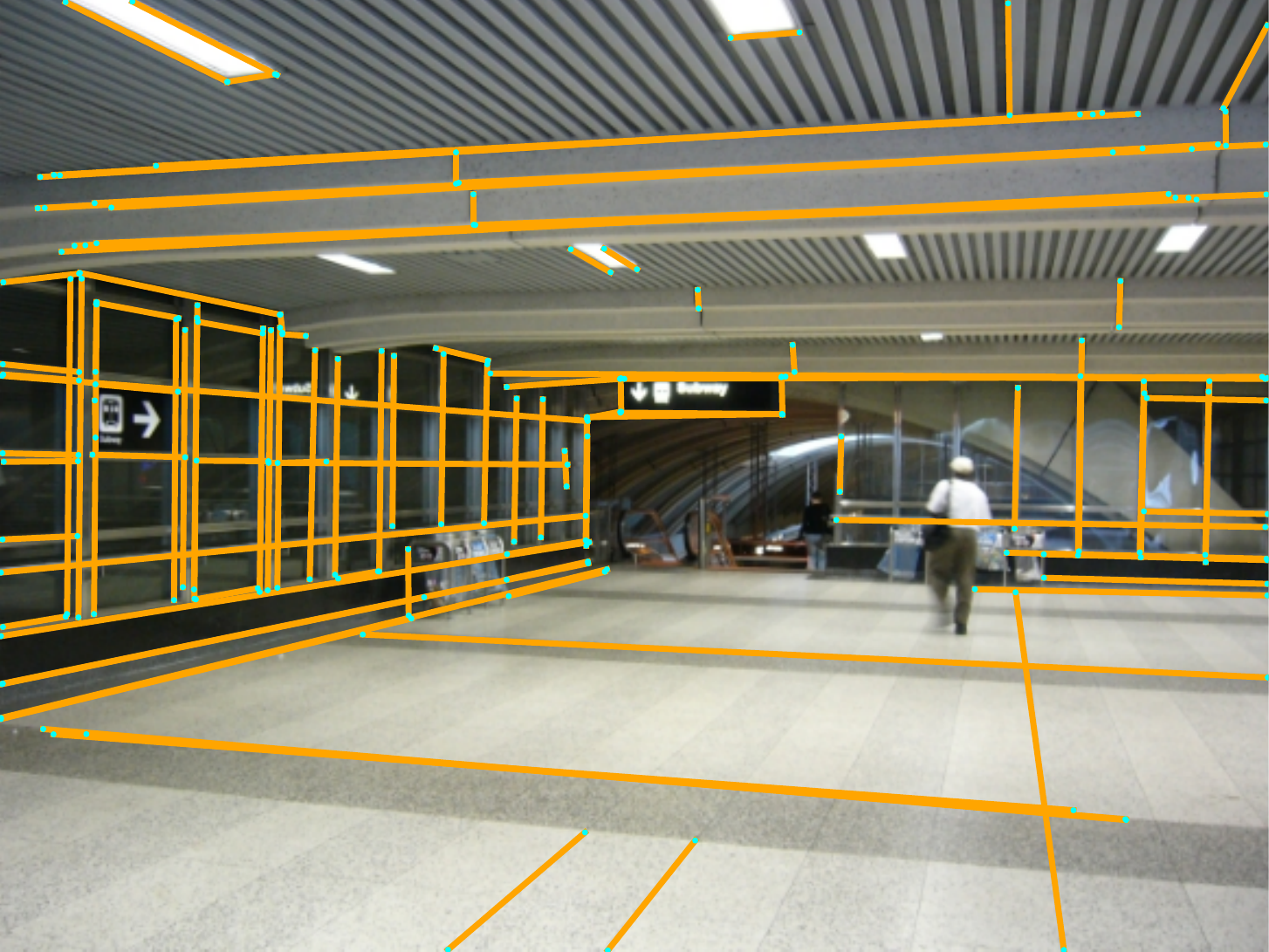}
        & \includegraphics[width=0.19\textwidth]{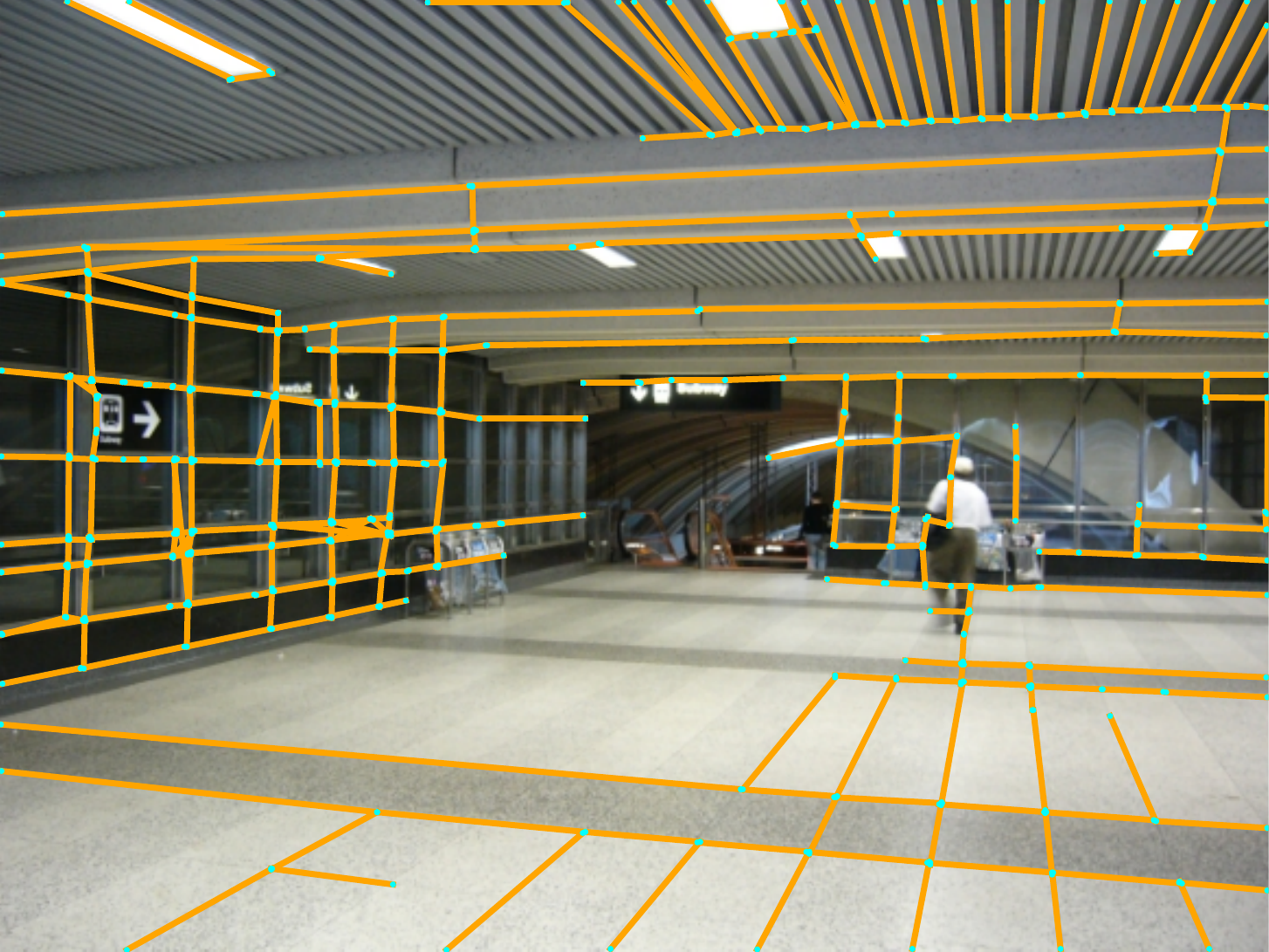}
        & \includegraphics[width=0.19\textwidth]{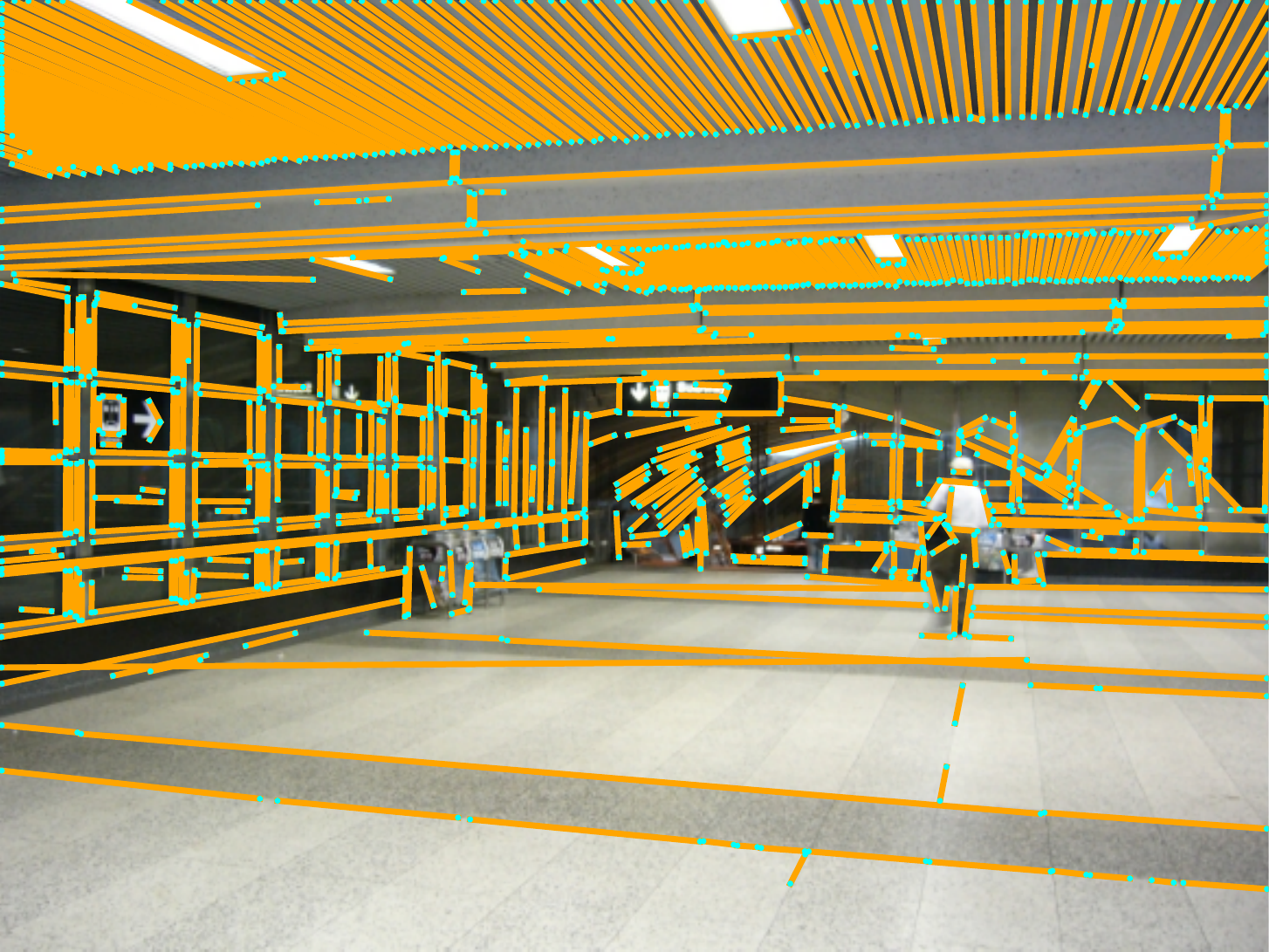}
        & \includegraphics[width=0.19\textwidth]{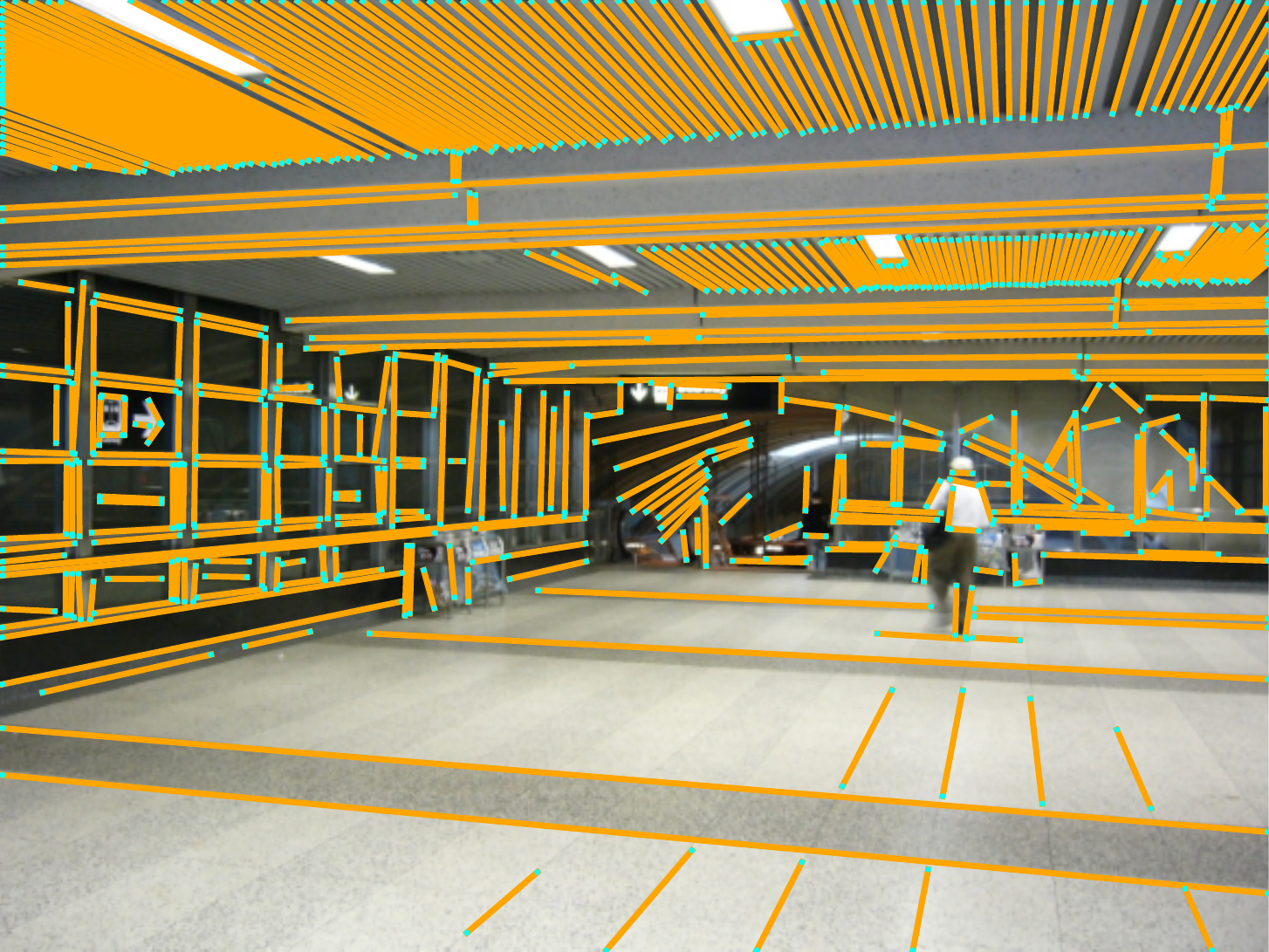} \\
        \includegraphics[width=0.19\textwidth]{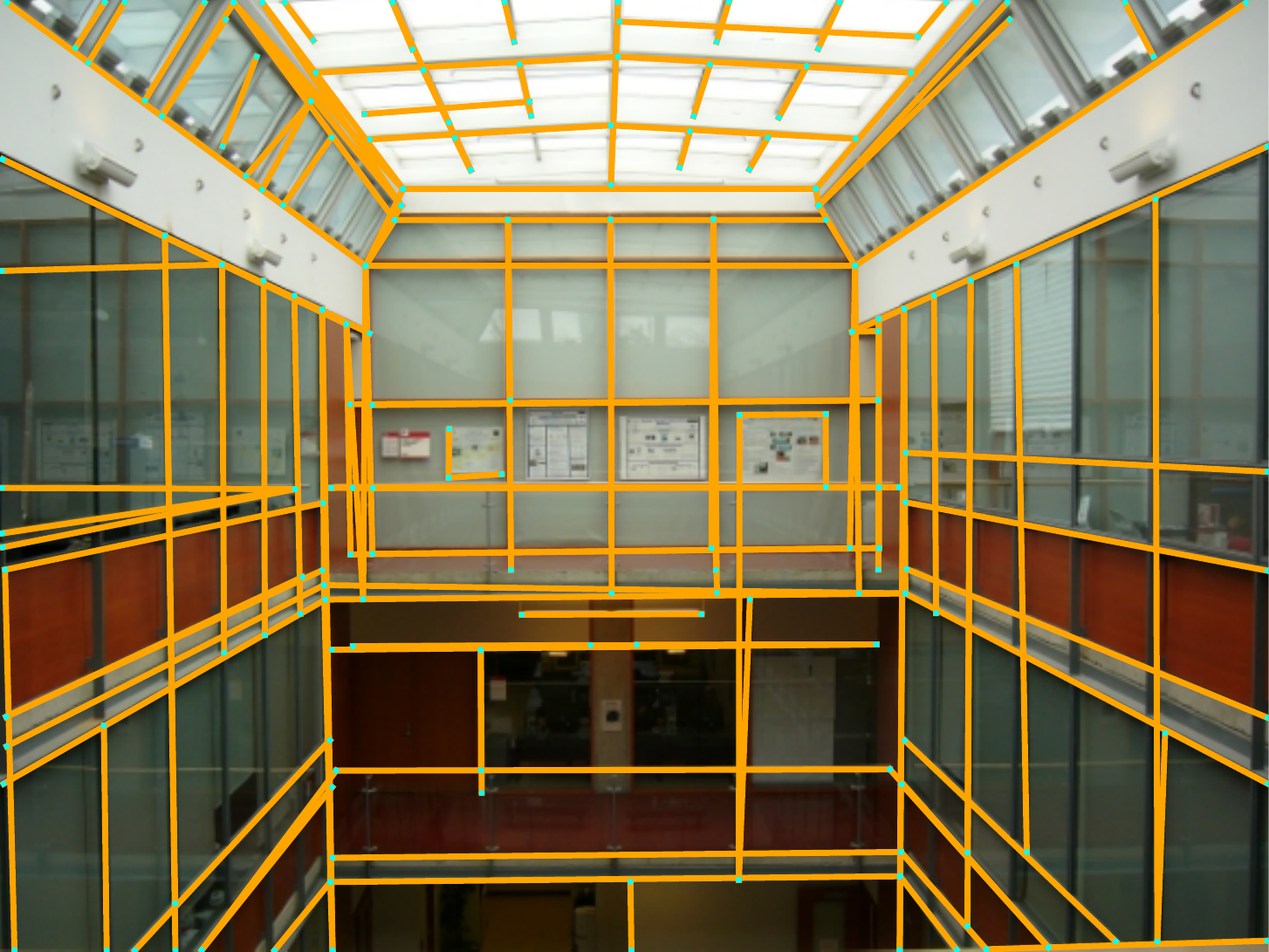}
        & \includegraphics[width=0.19\textwidth]{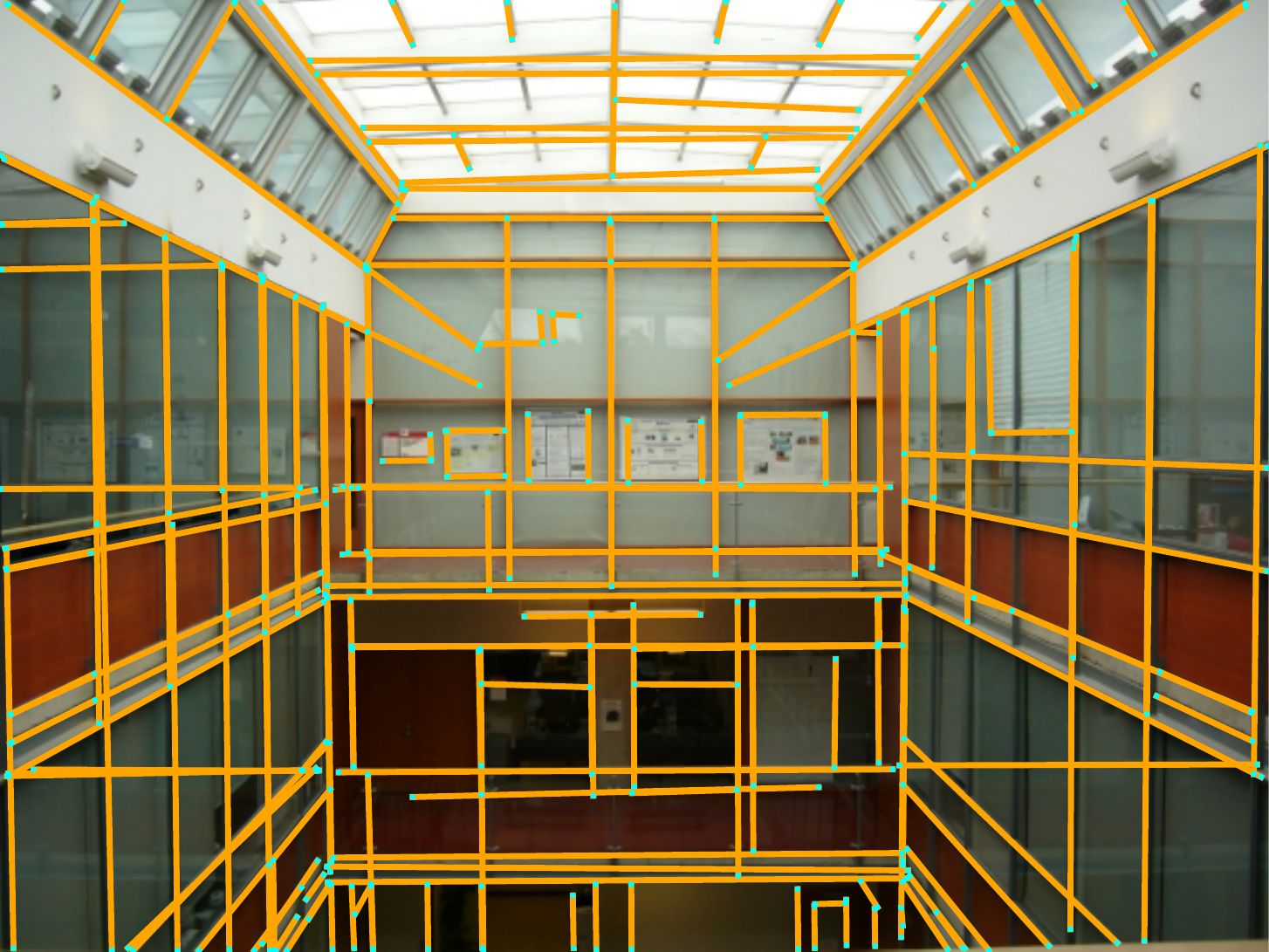}
        & \includegraphics[width=0.19\textwidth]{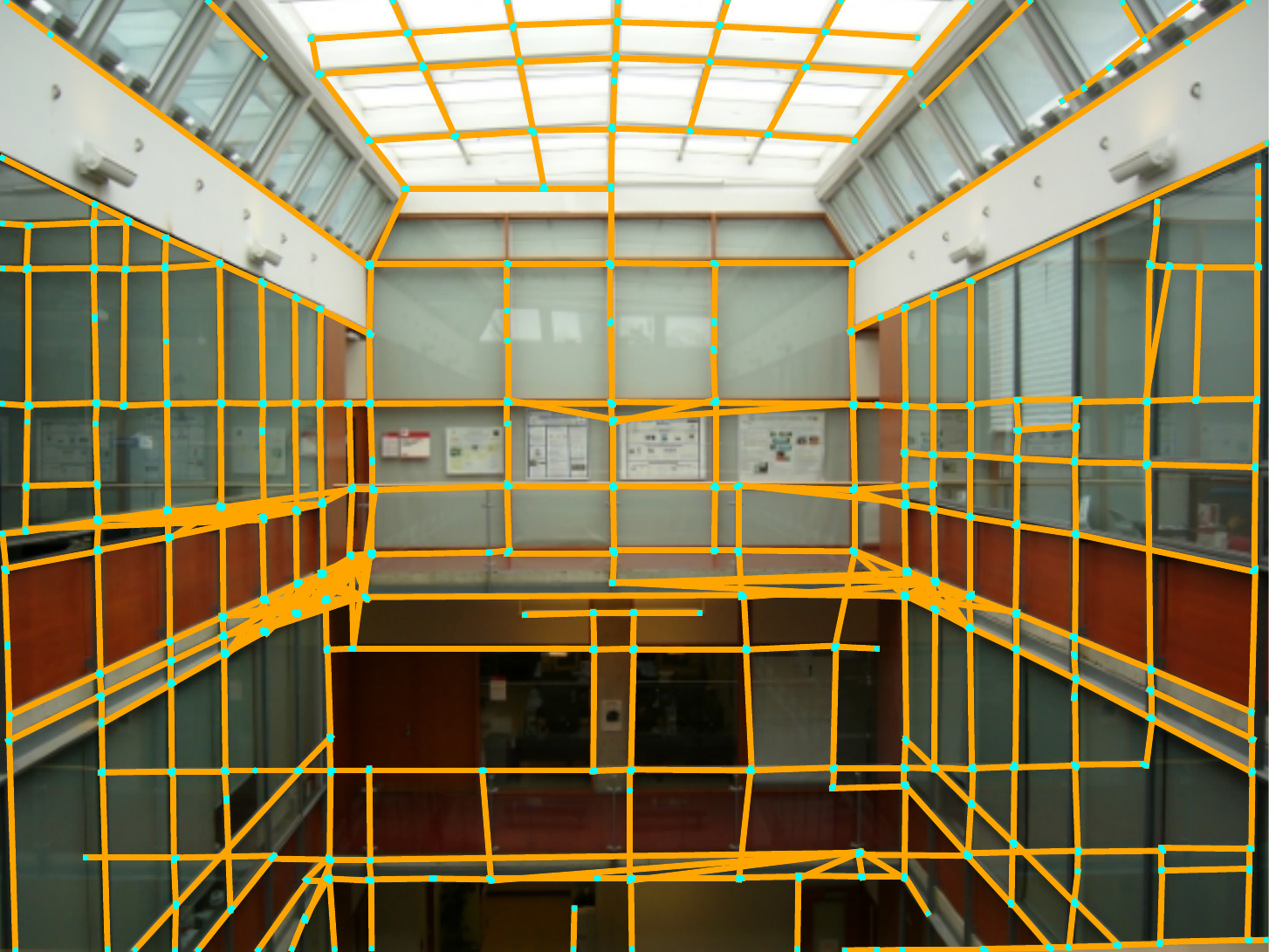}
        & \includegraphics[width=0.19\textwidth]{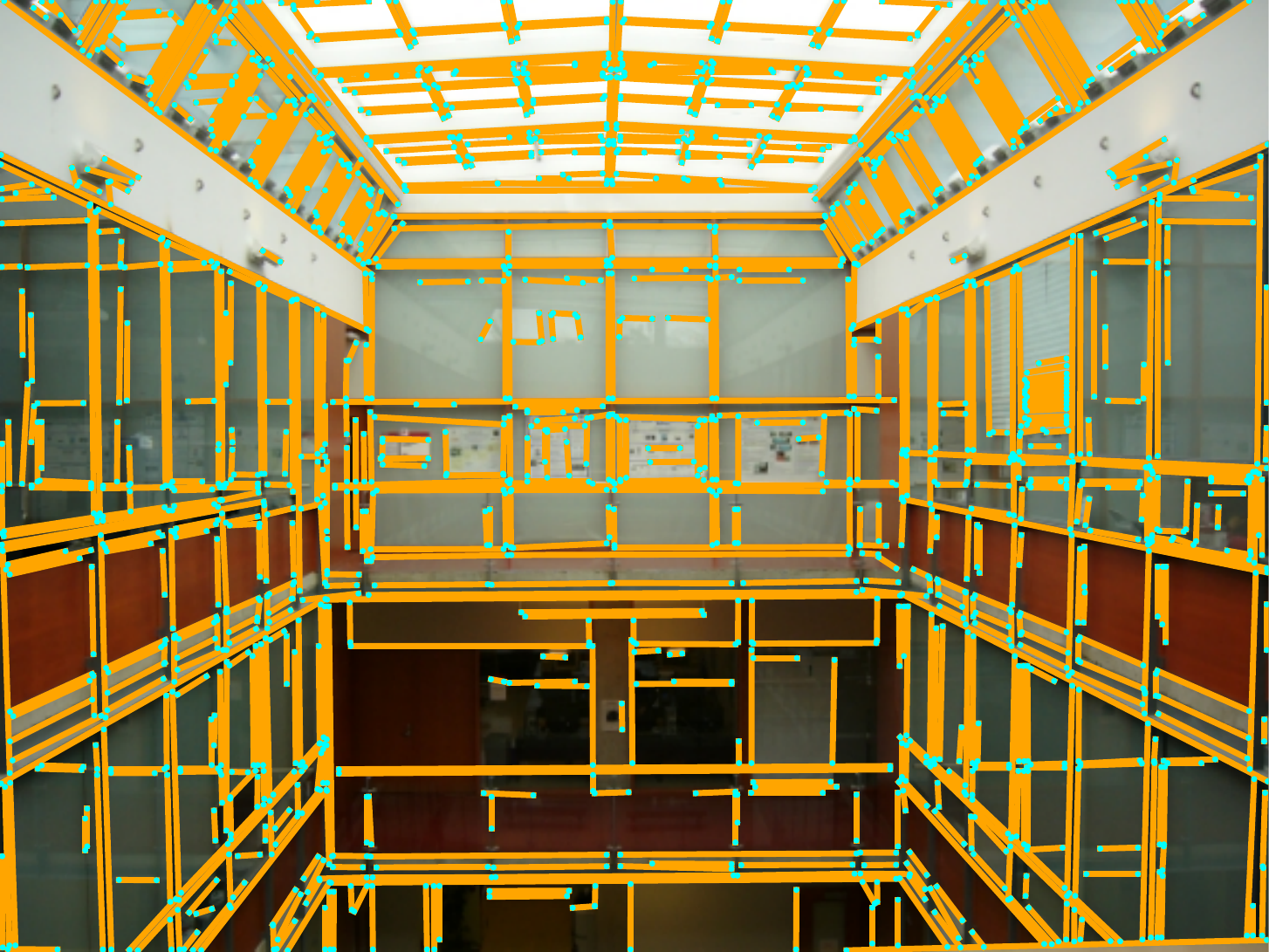}
        & \includegraphics[width=0.19\textwidth]{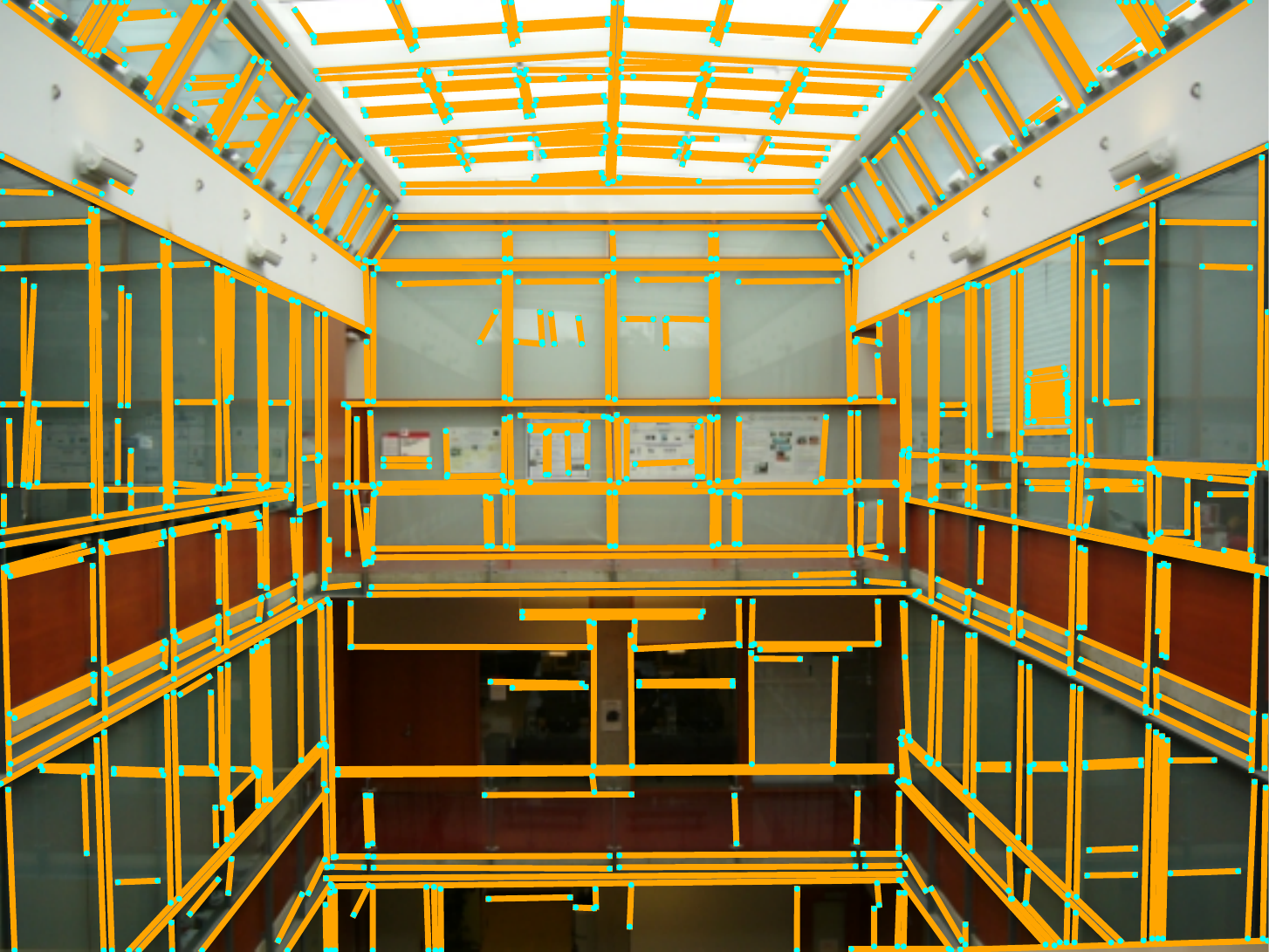} \\
        \includegraphics[width=0.19\textwidth]{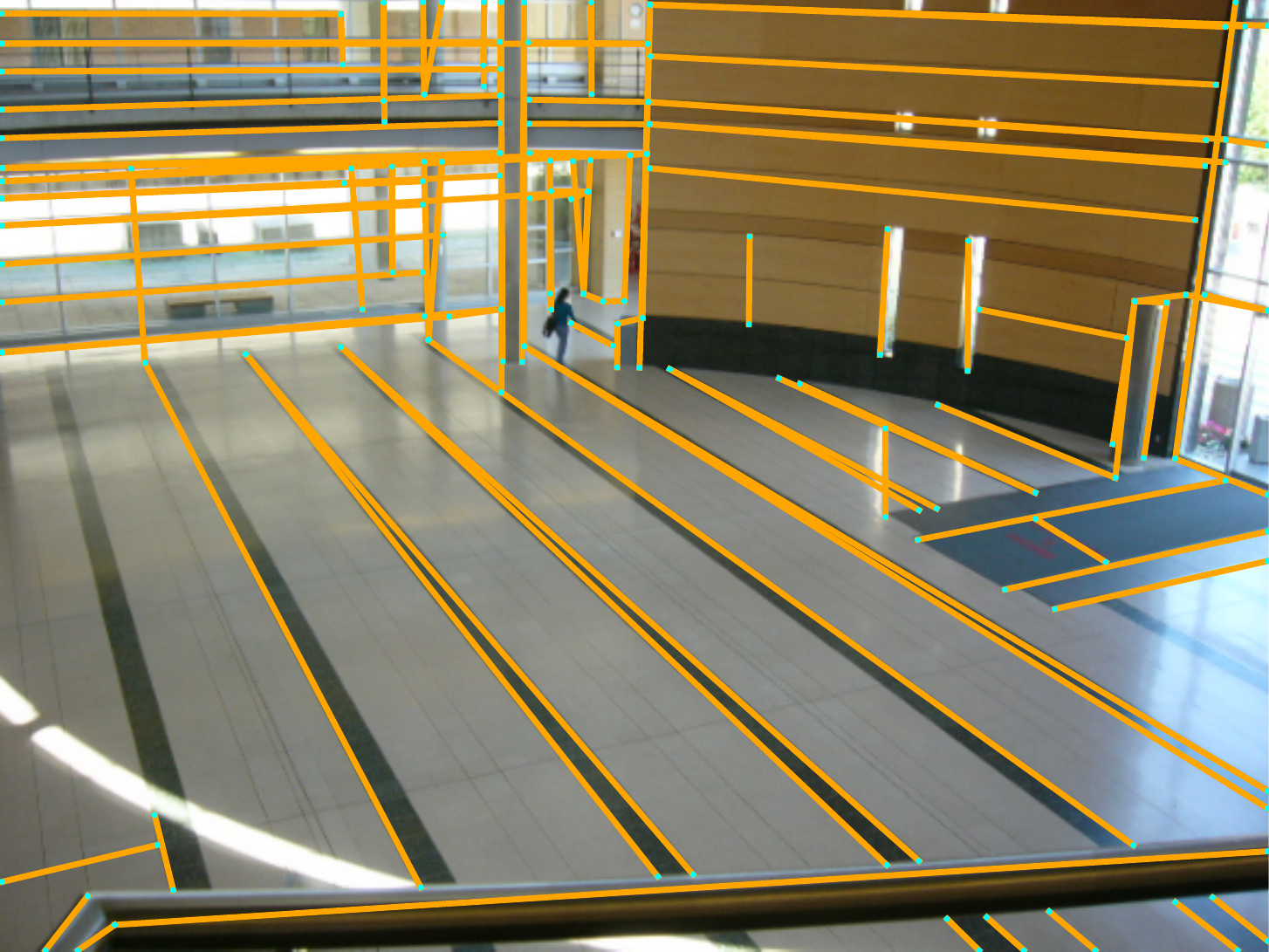}
        & \includegraphics[width=0.19\textwidth]{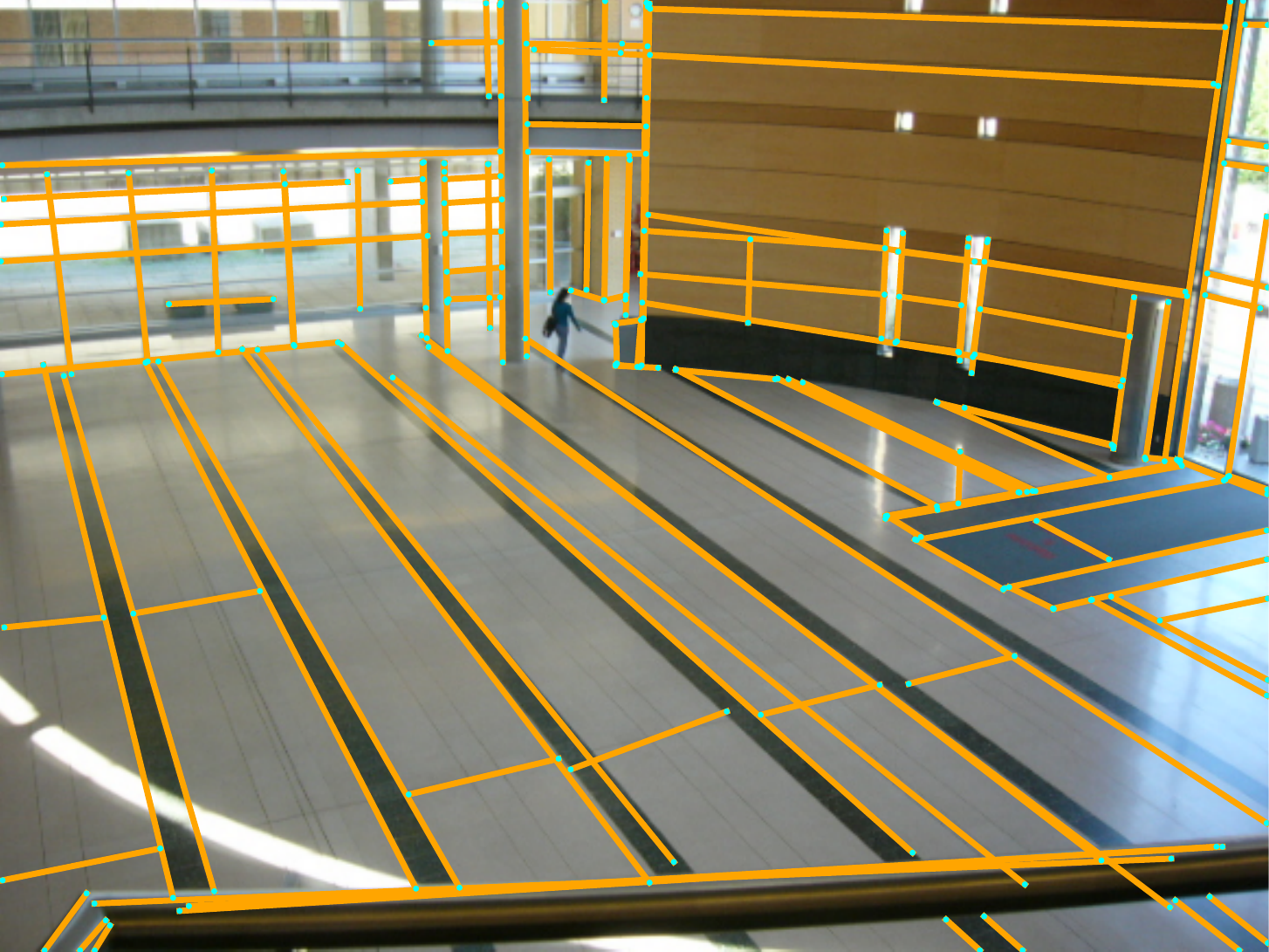}
        & \includegraphics[width=0.19\textwidth]{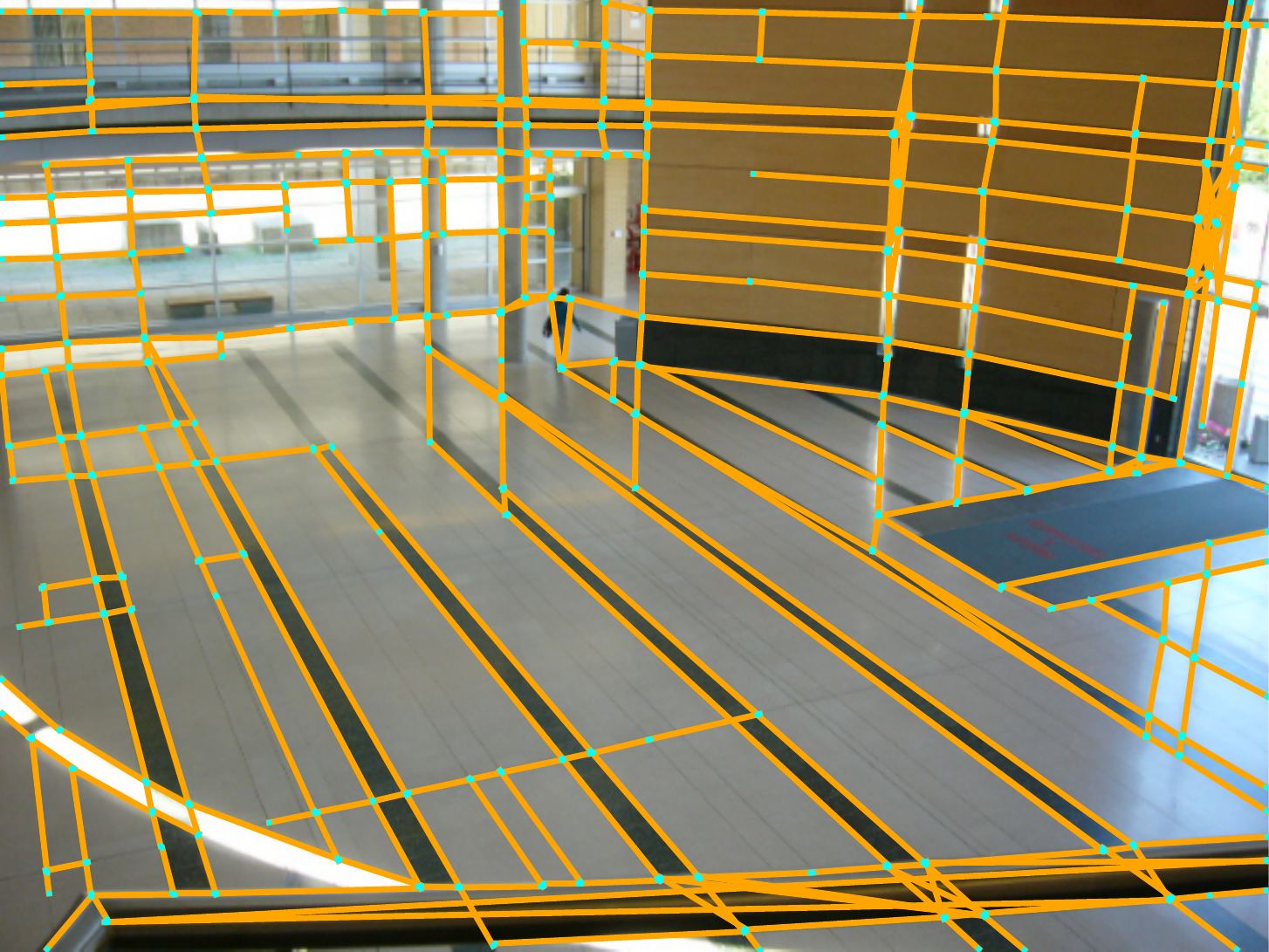}
        & \includegraphics[width=0.19\textwidth]{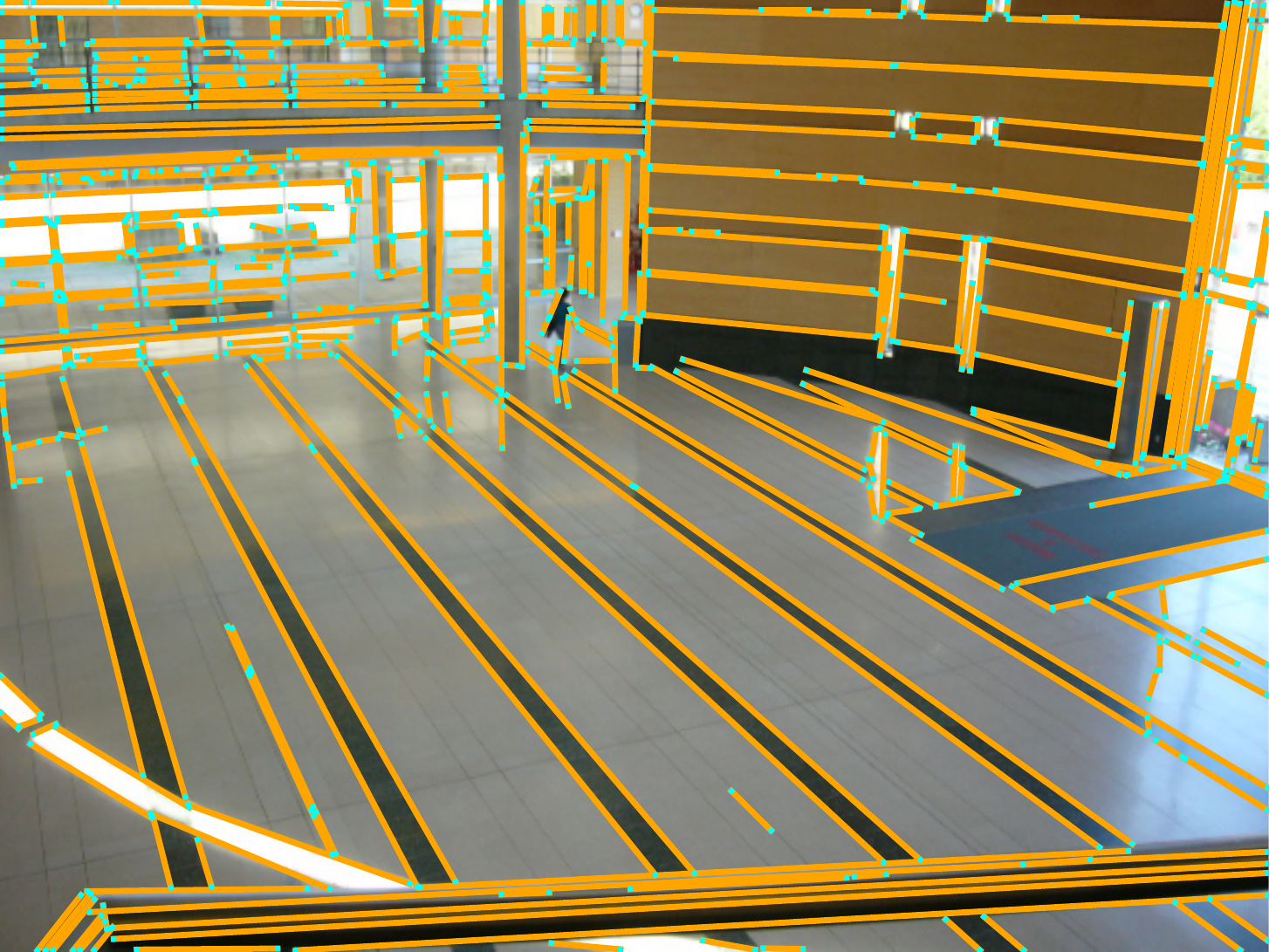}
        & \includegraphics[width=0.19\textwidth]{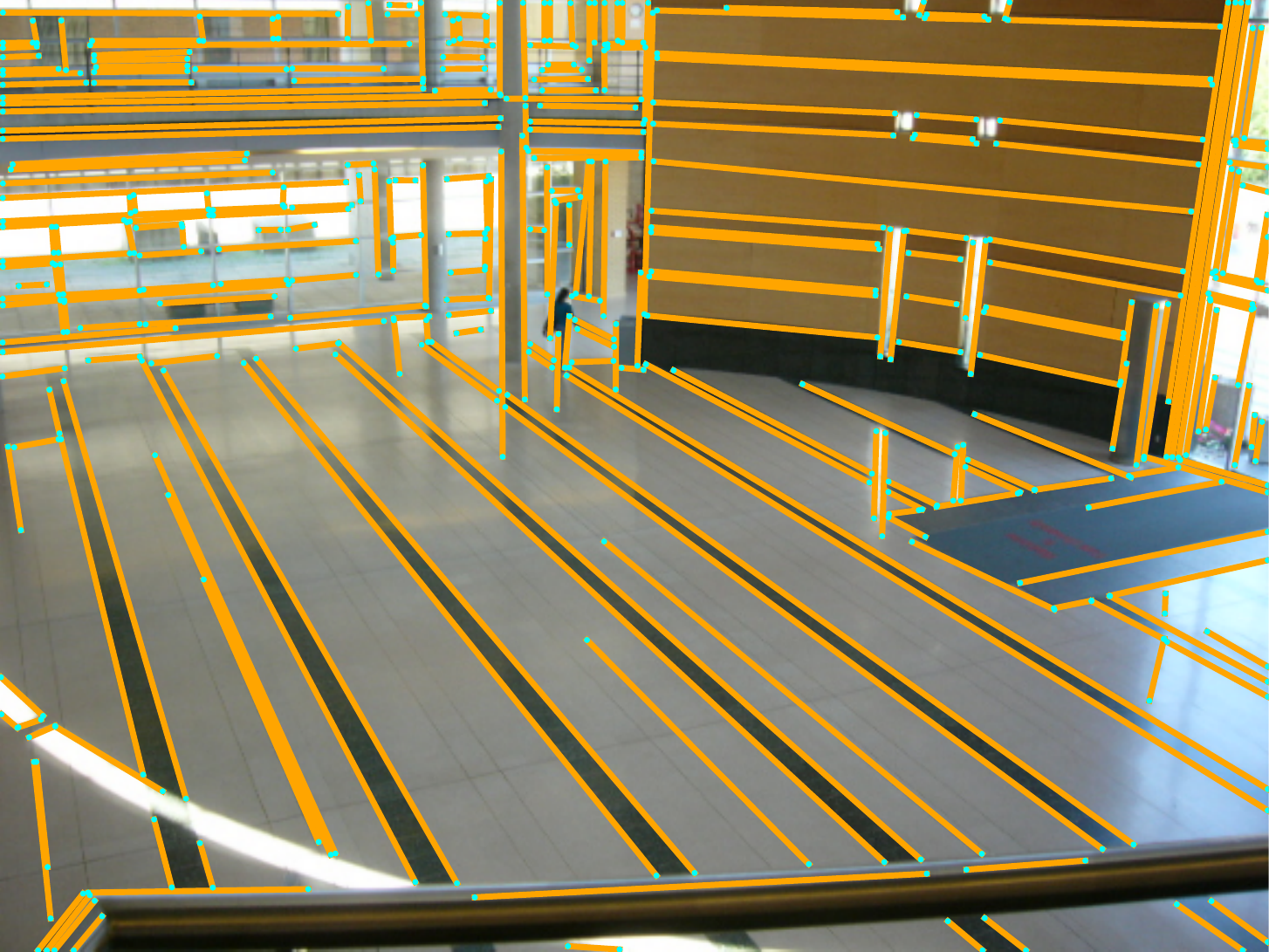} \\
        \includegraphics[width=0.19\textwidth]{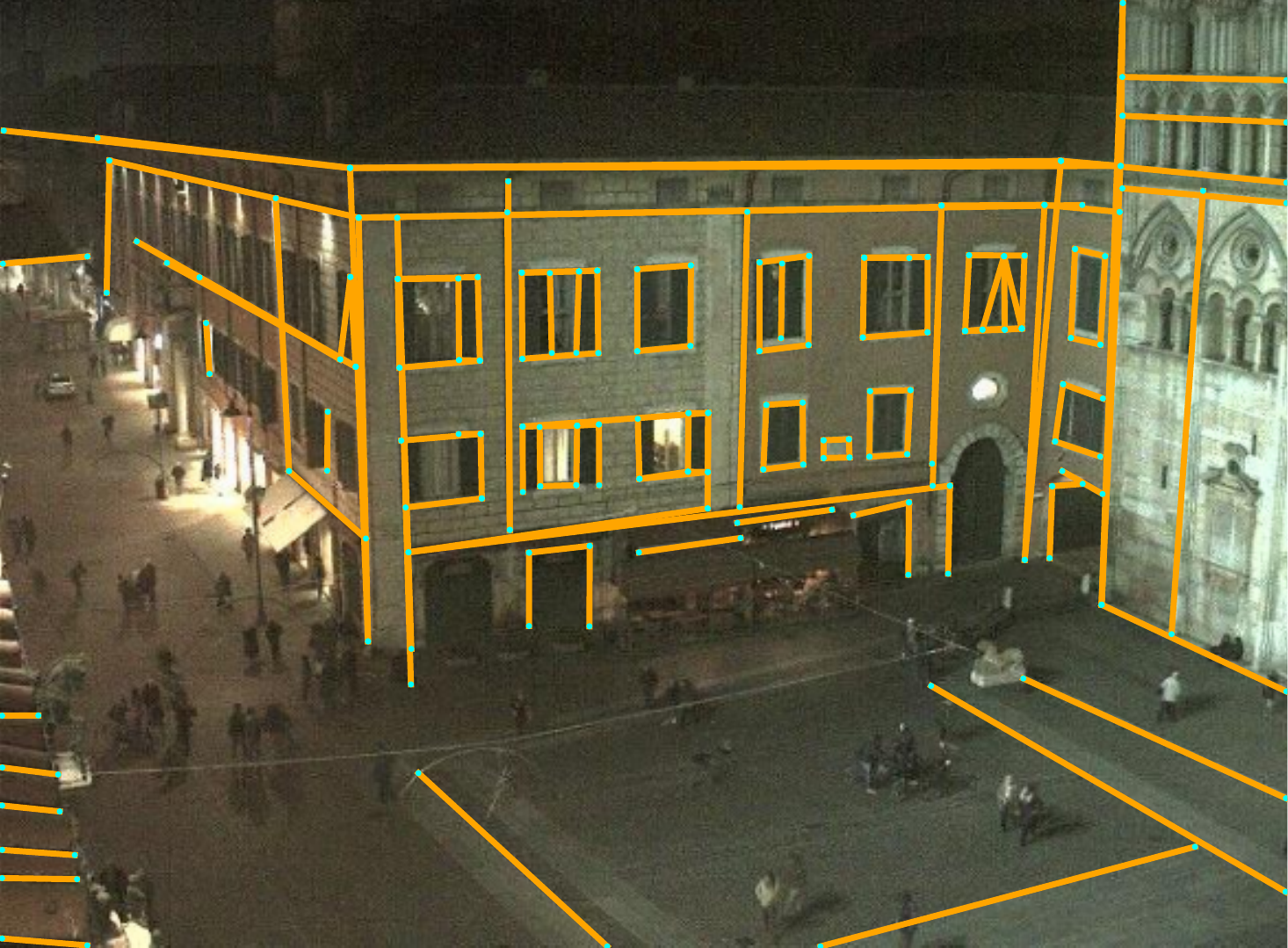}
        & \includegraphics[width=0.19\textwidth]{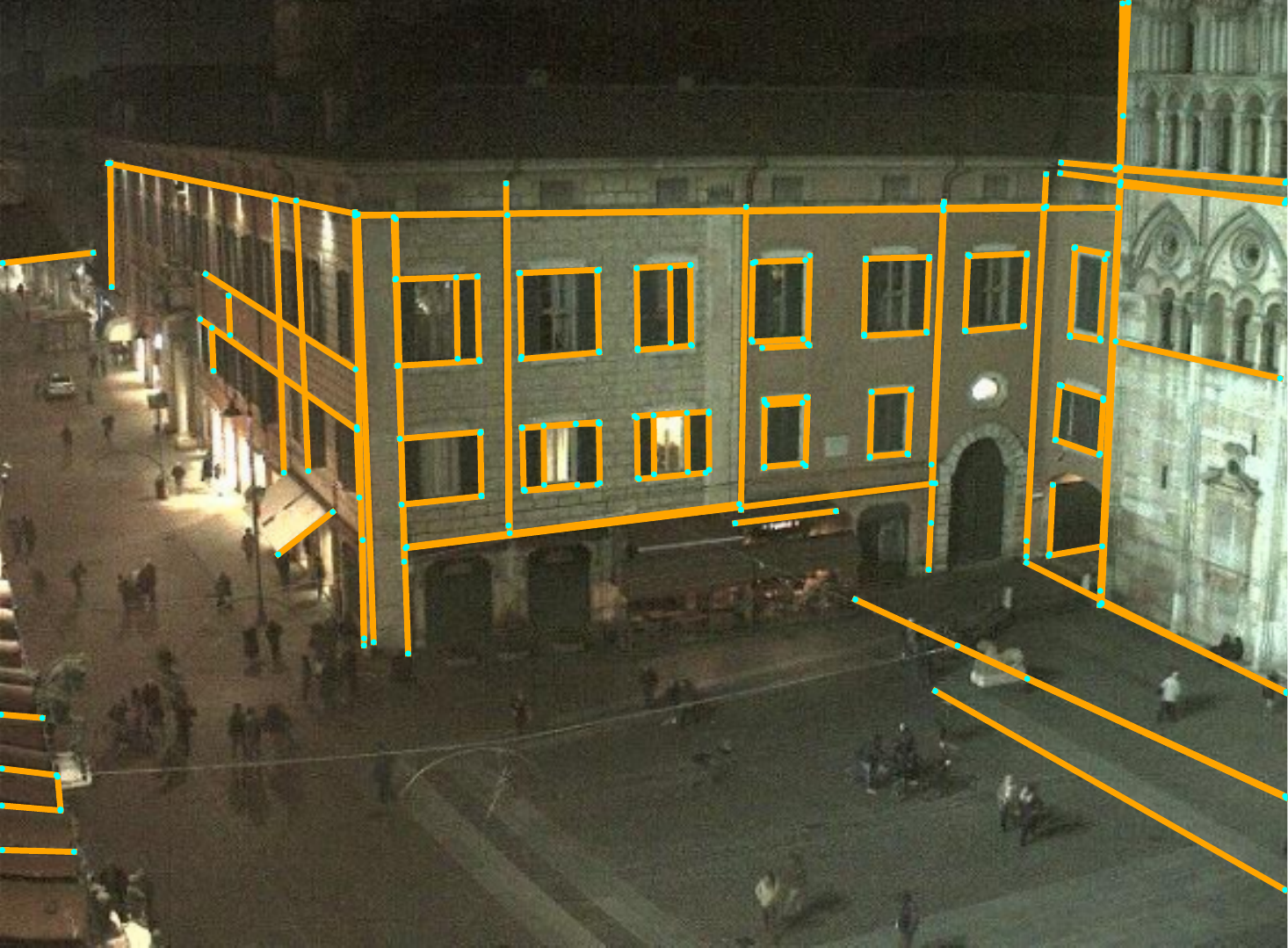}
        & \includegraphics[width=0.19\textwidth]{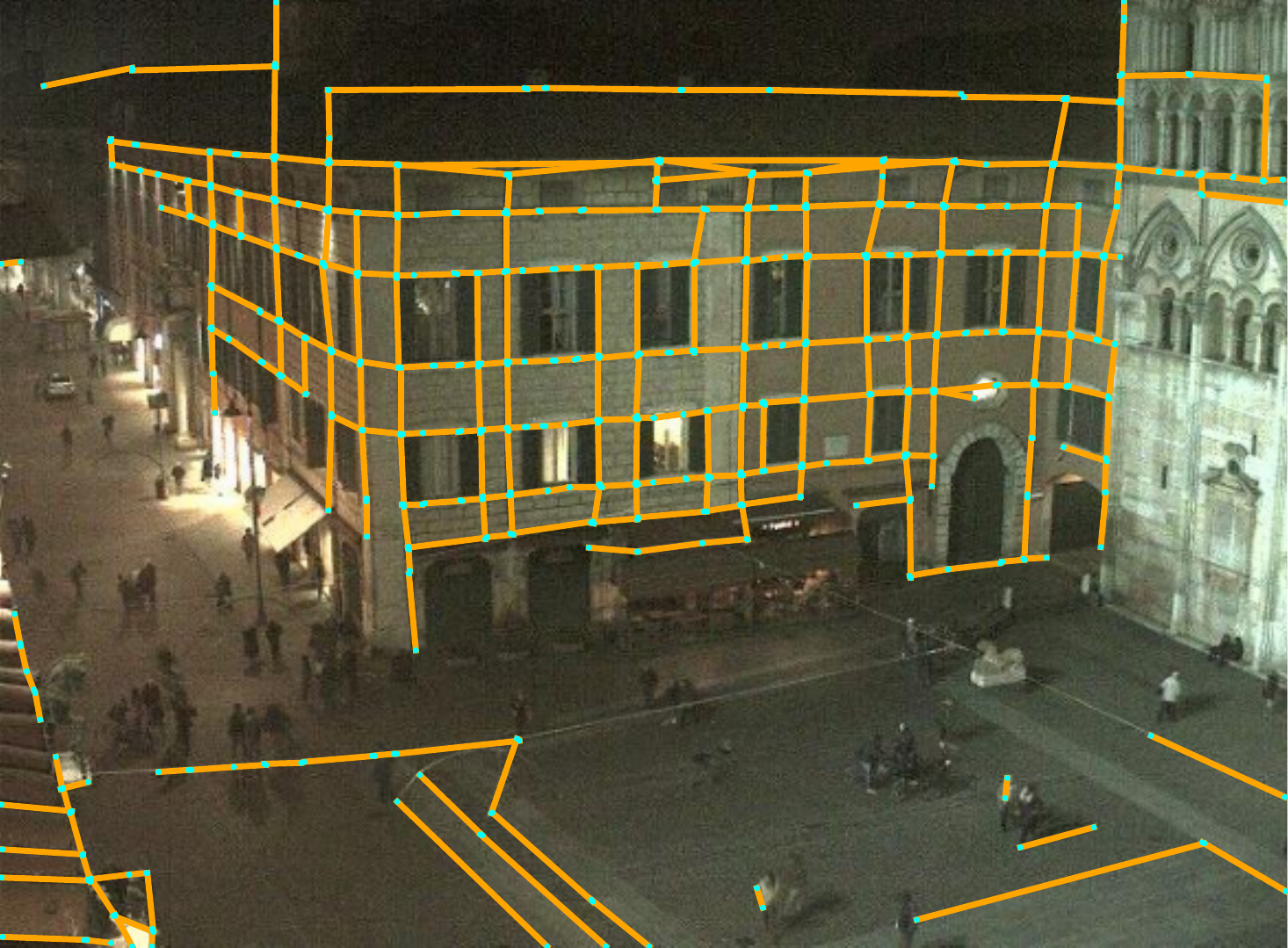}
        & \includegraphics[width=0.19\textwidth]{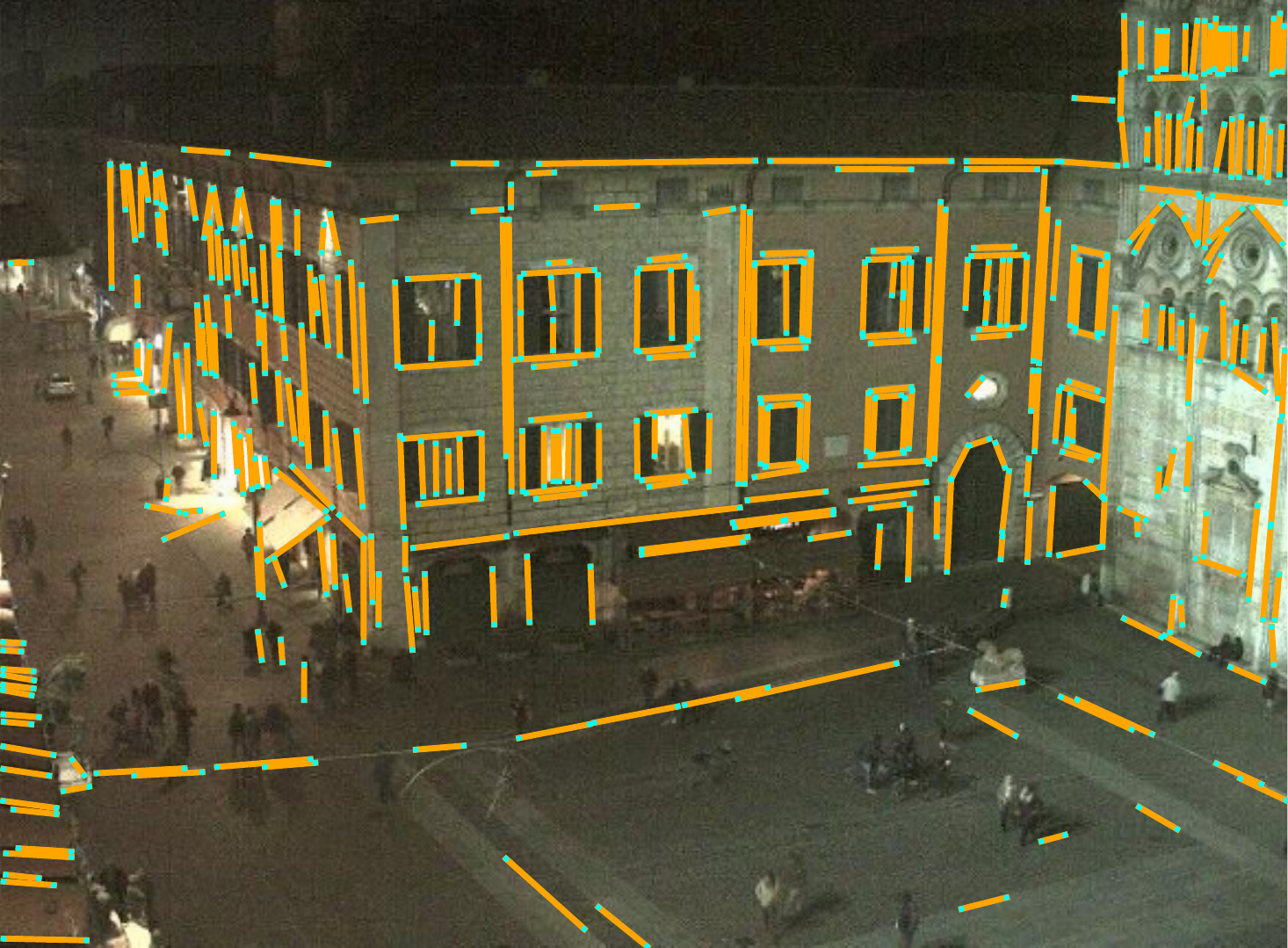}
        & \includegraphics[width=0.19\textwidth]{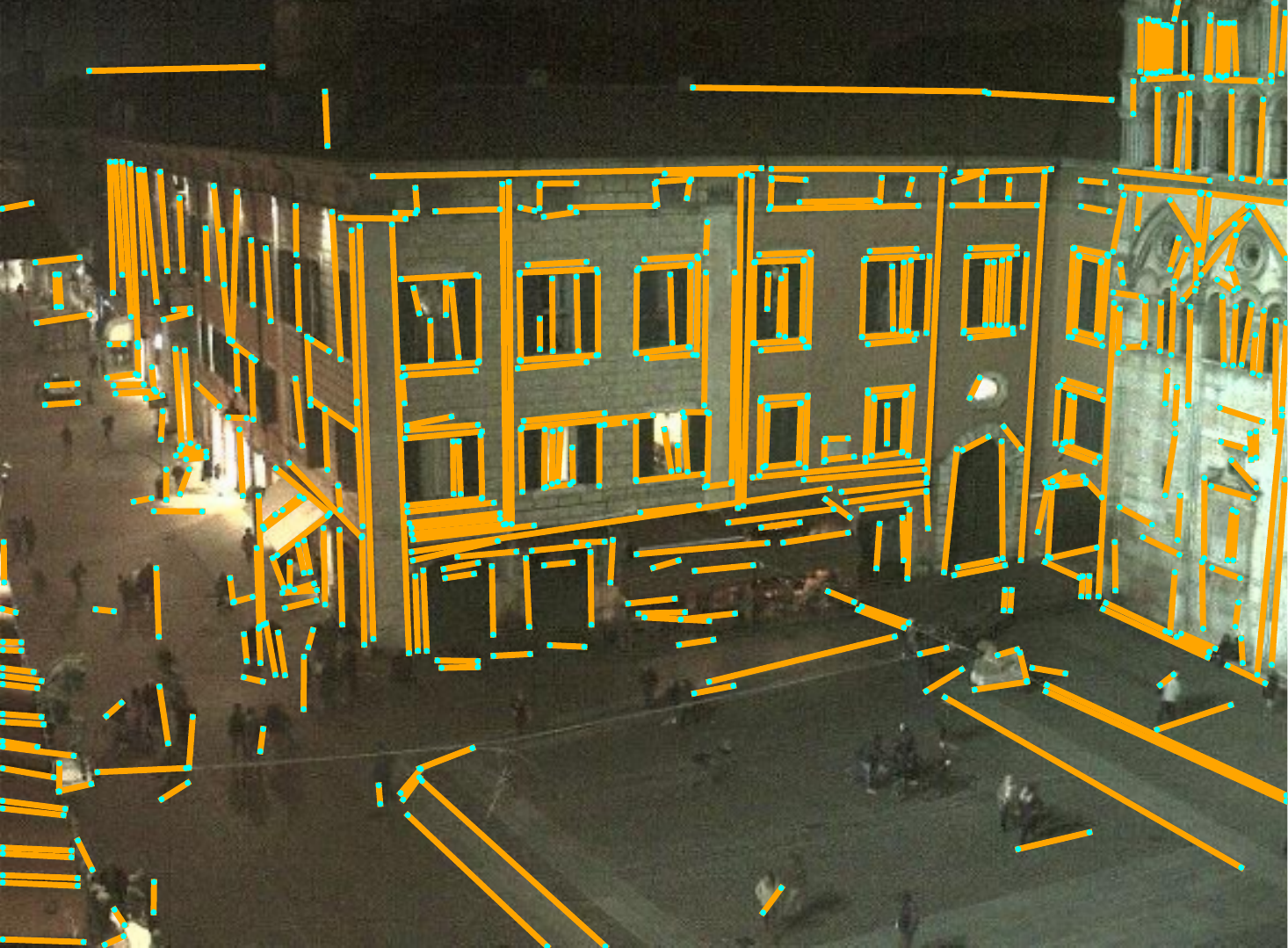} \\
        \includegraphics[width=0.19\textwidth]{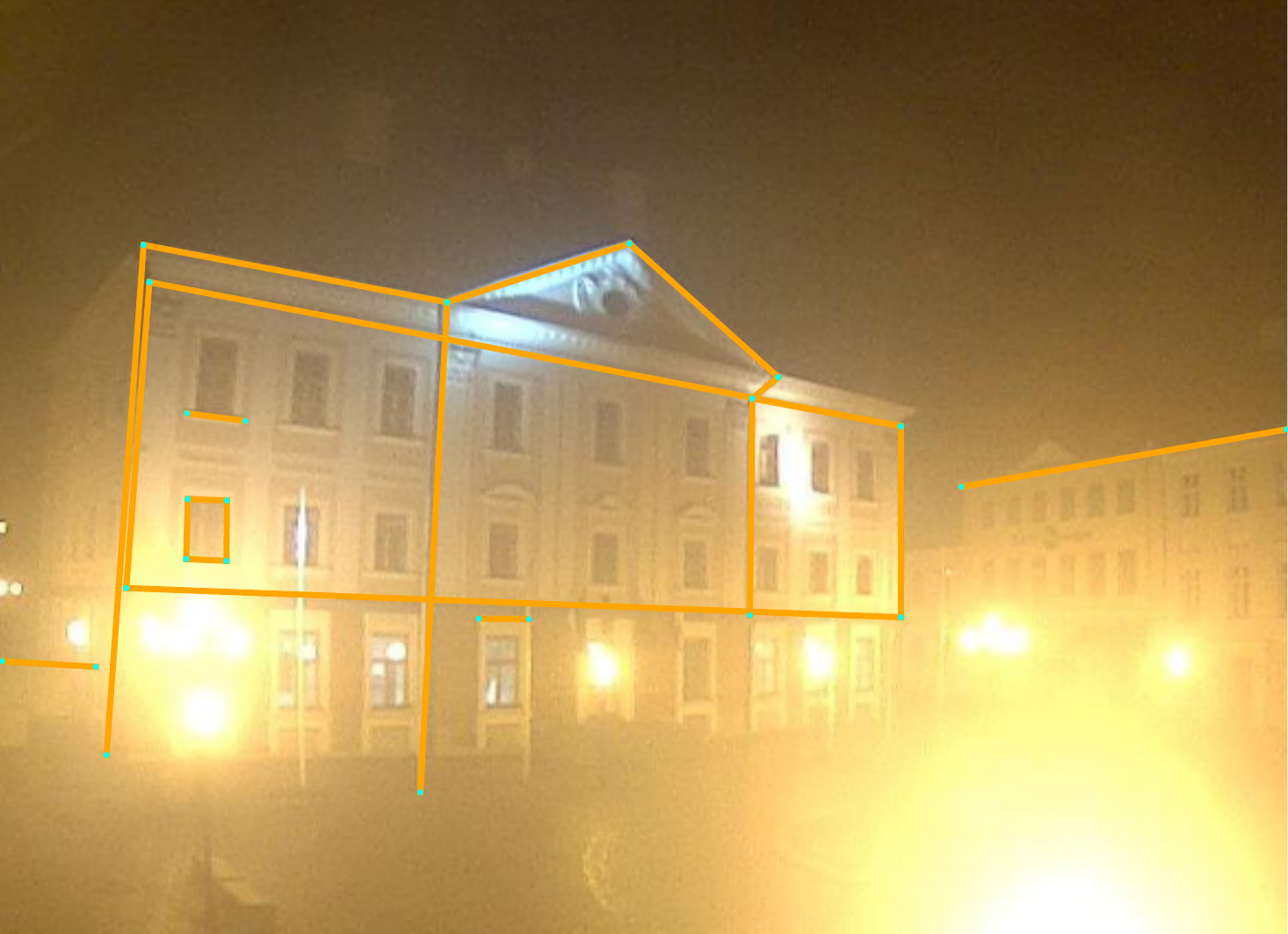}
        & \includegraphics[width=0.19\textwidth]{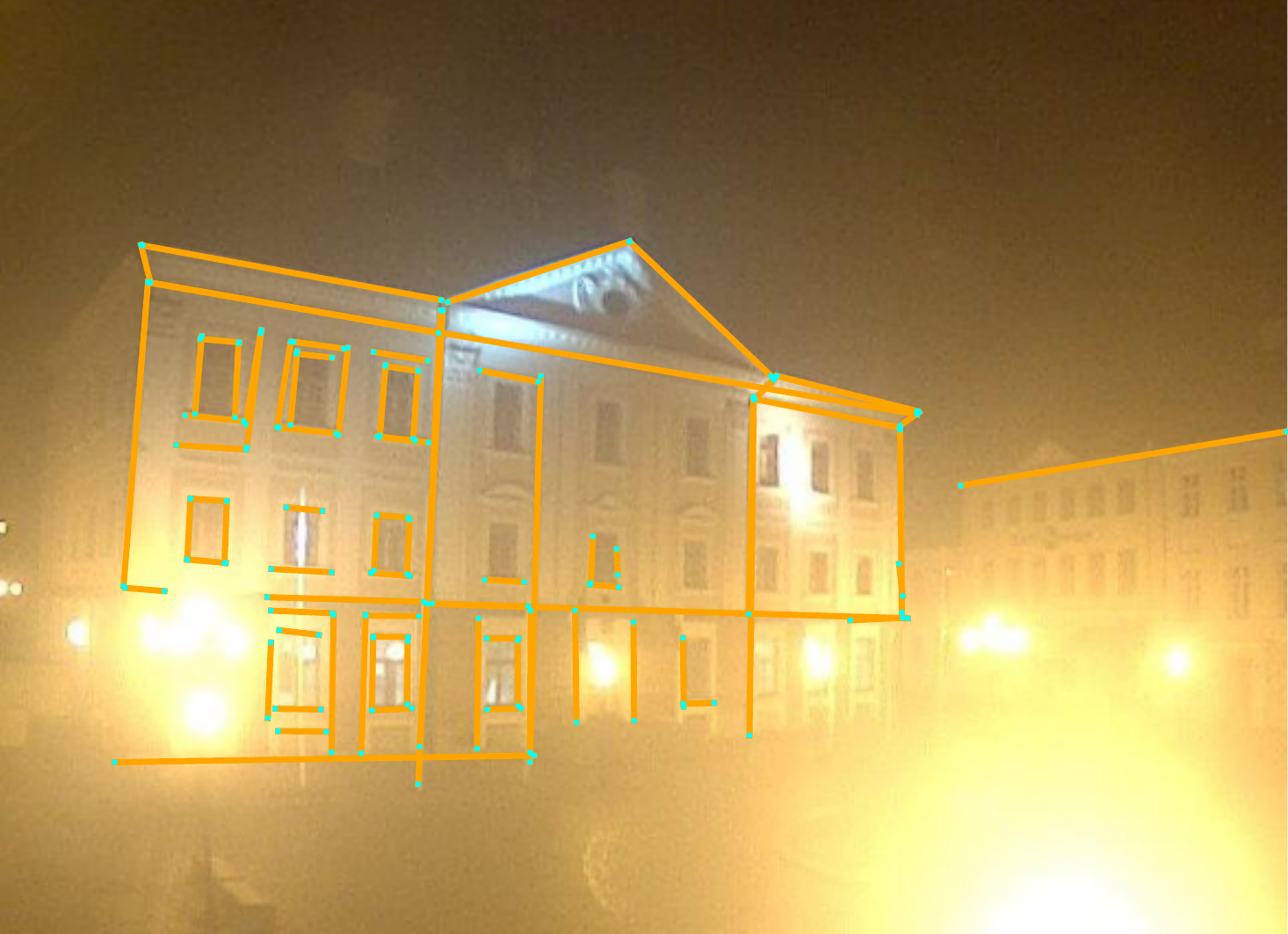}
        & \includegraphics[width=0.19\textwidth]{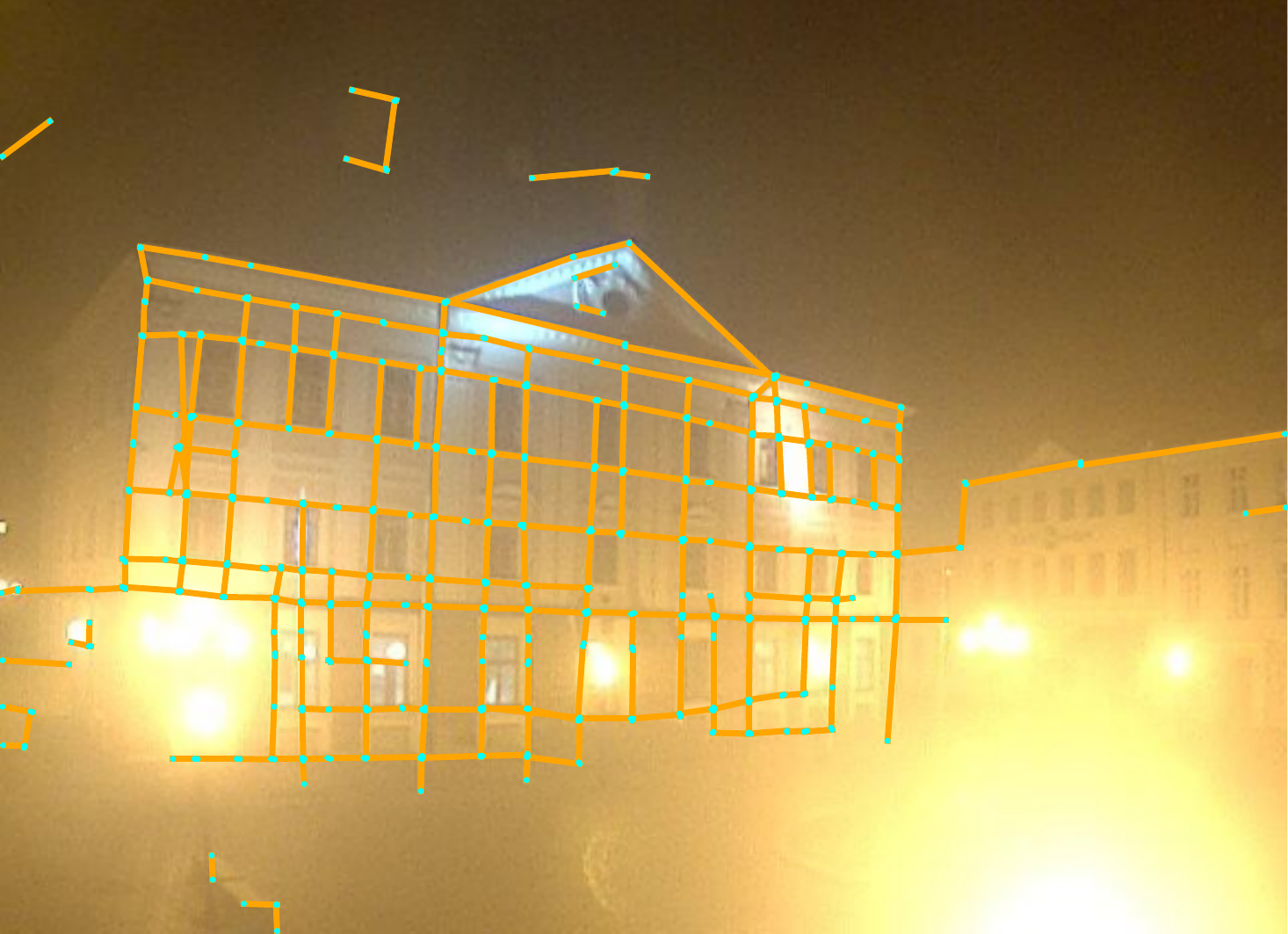}
        & \includegraphics[width=0.19\textwidth]{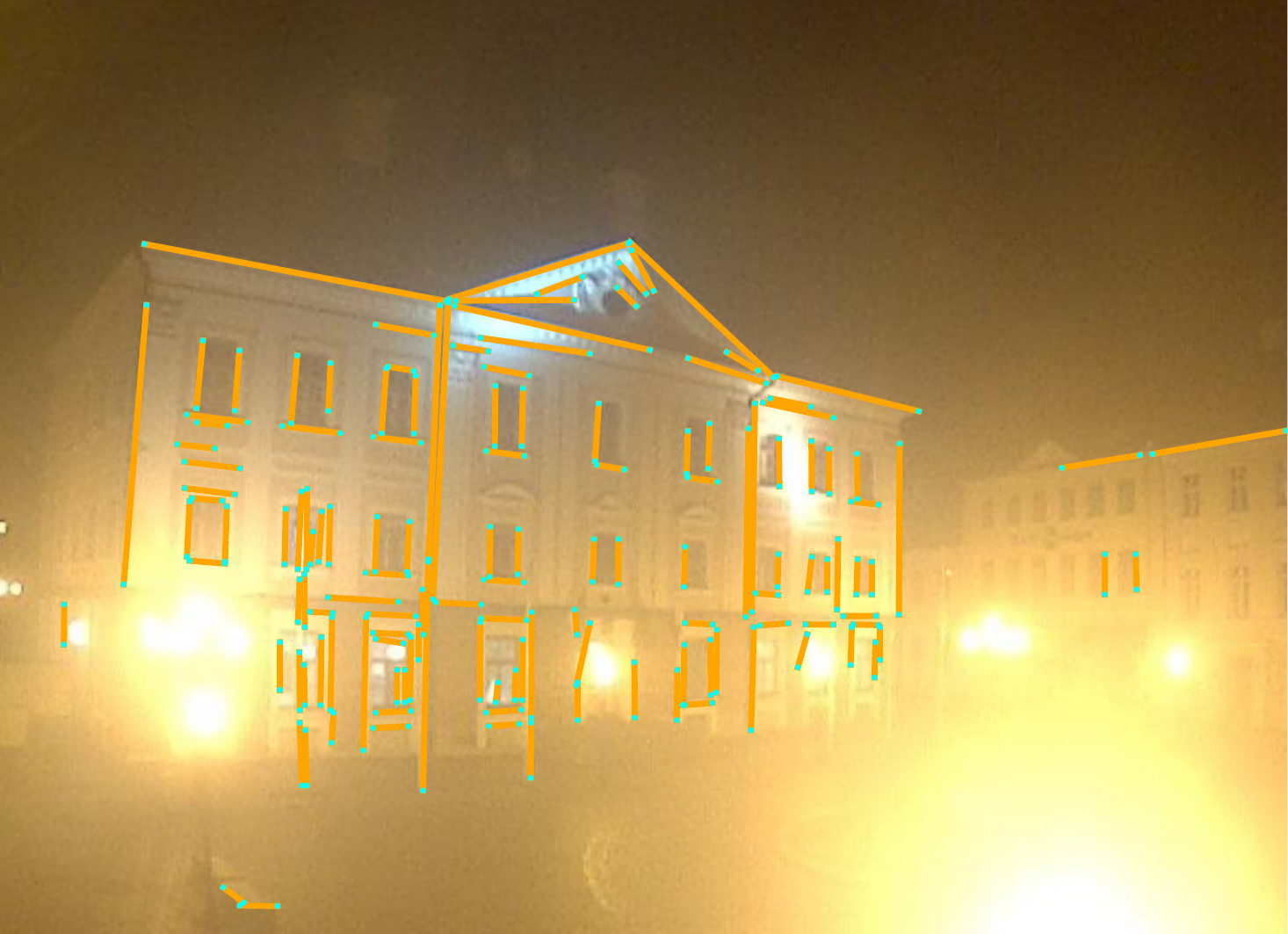}
        & \includegraphics[width=0.19\textwidth]{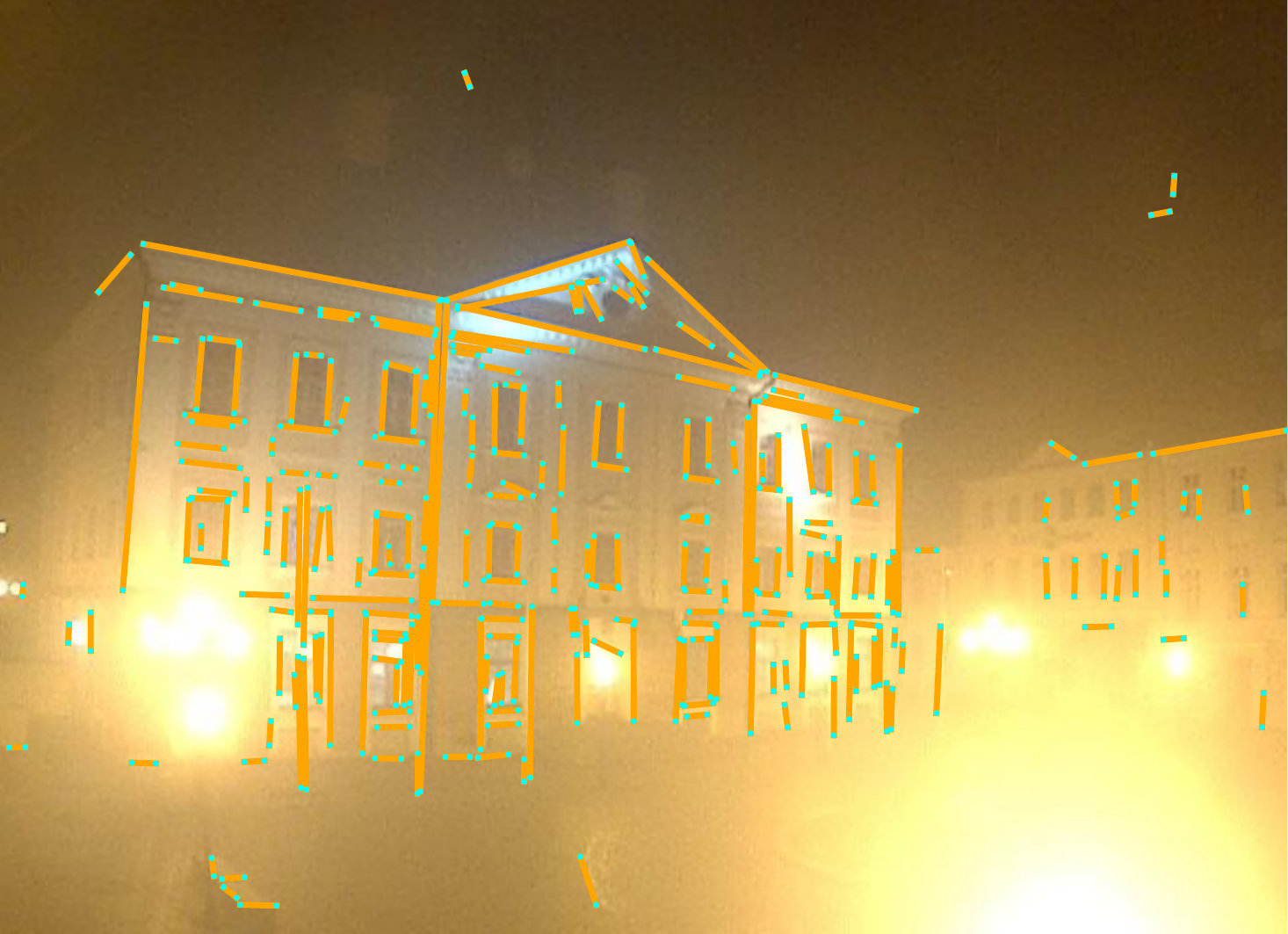} \\
        \includegraphics[width=0.19\textwidth]{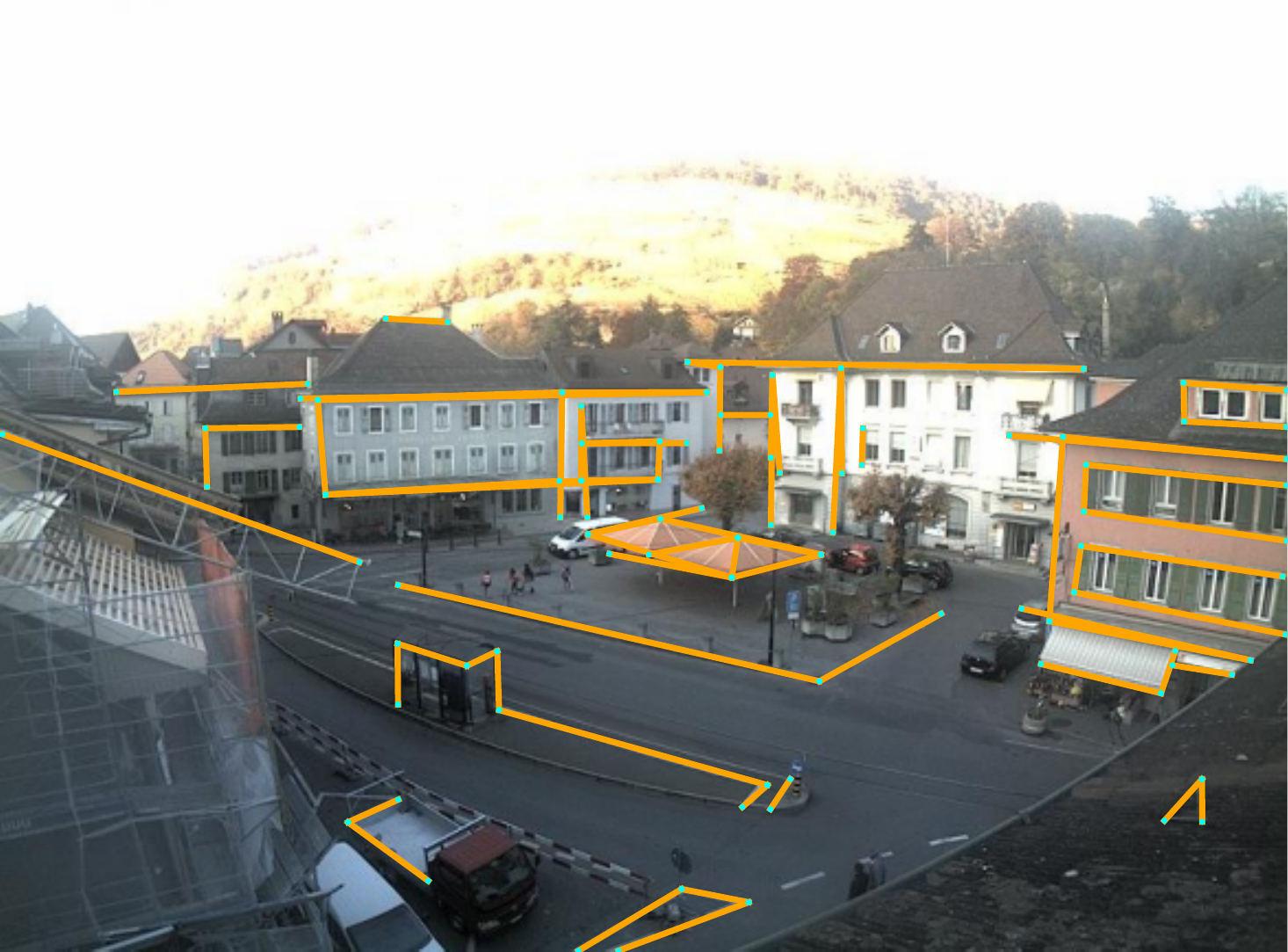}
        & \includegraphics[width=0.19\textwidth]{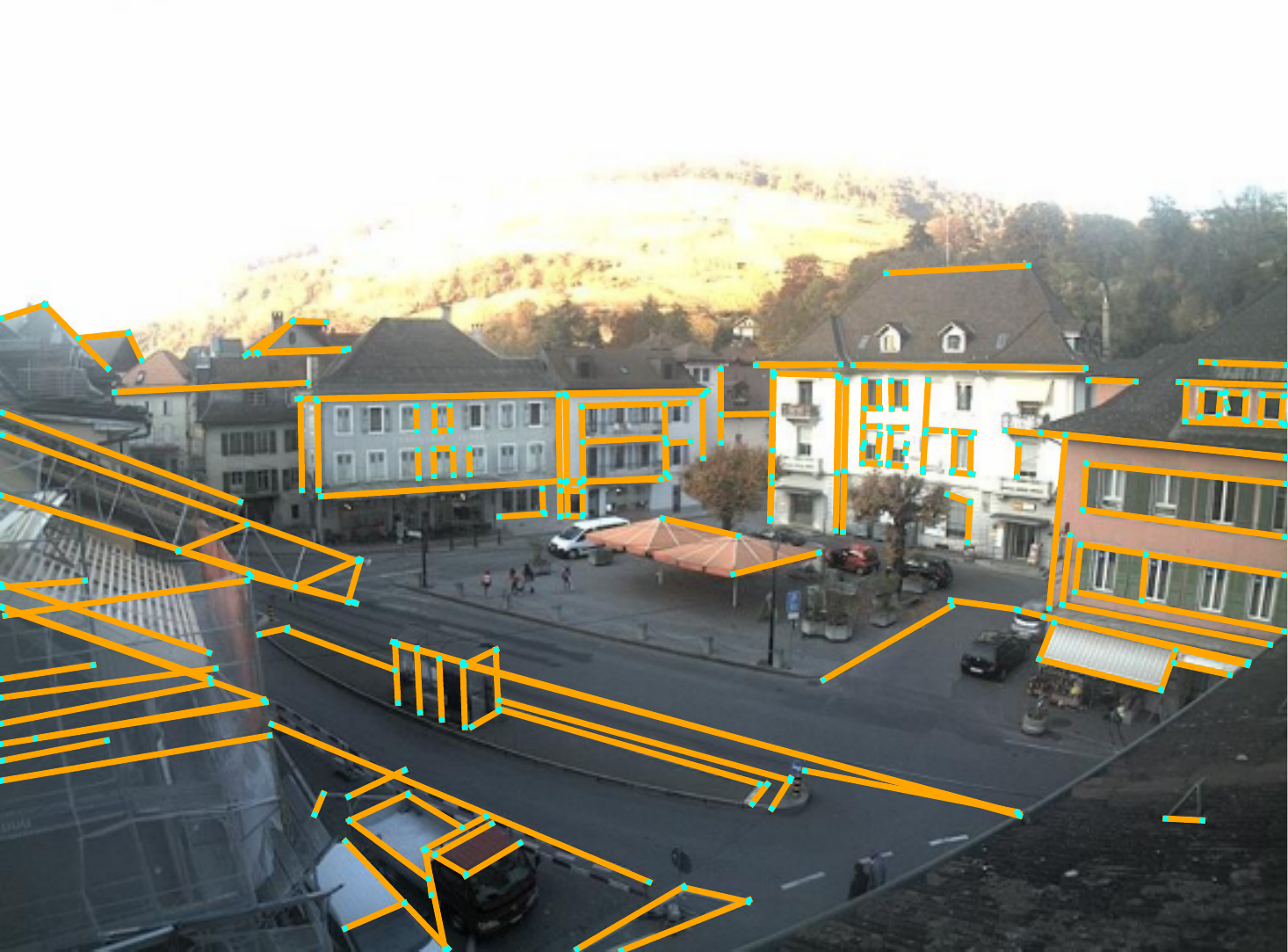}
        & \includegraphics[width=0.19\textwidth]{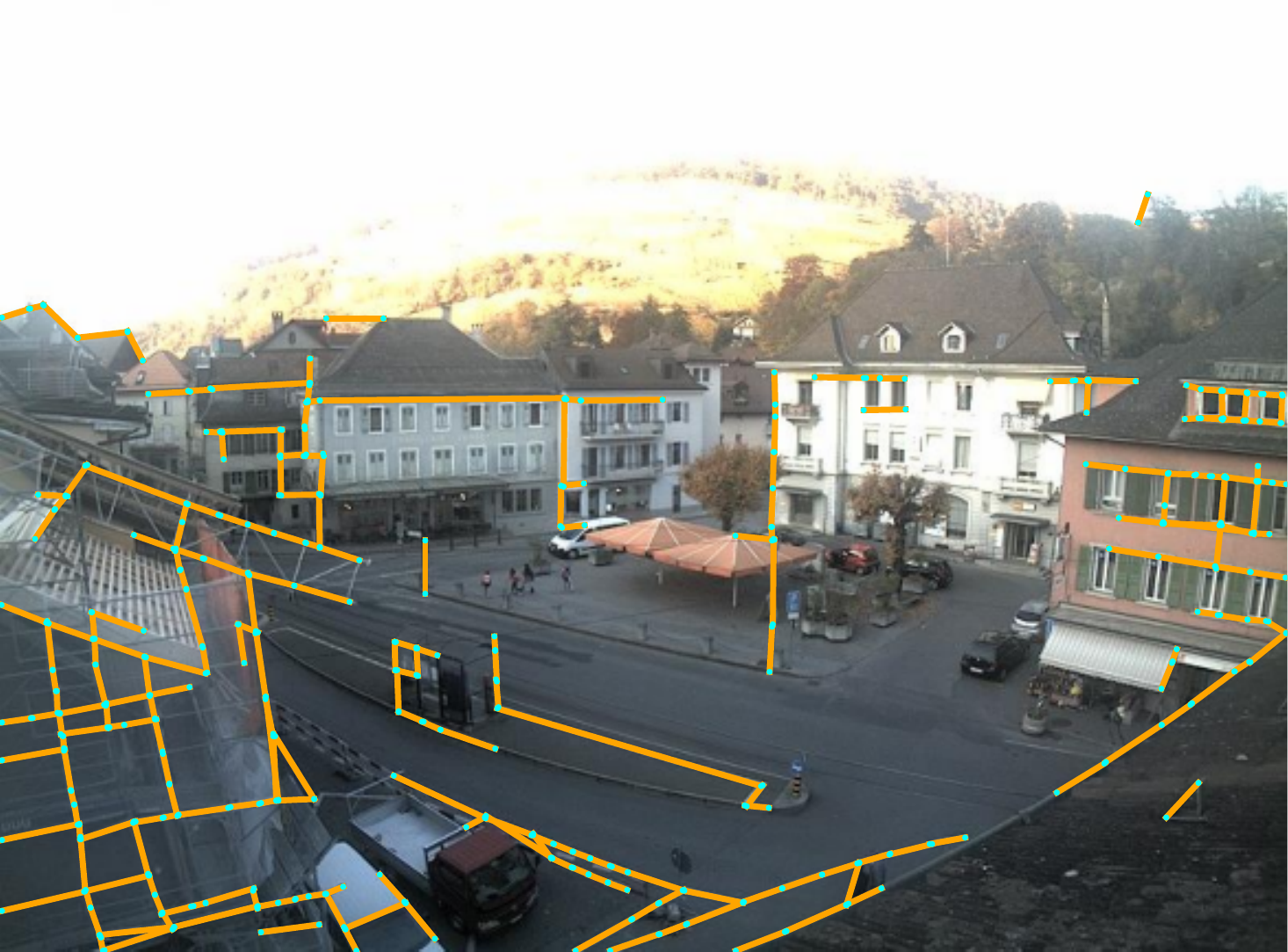}
        & \includegraphics[width=0.19\textwidth]{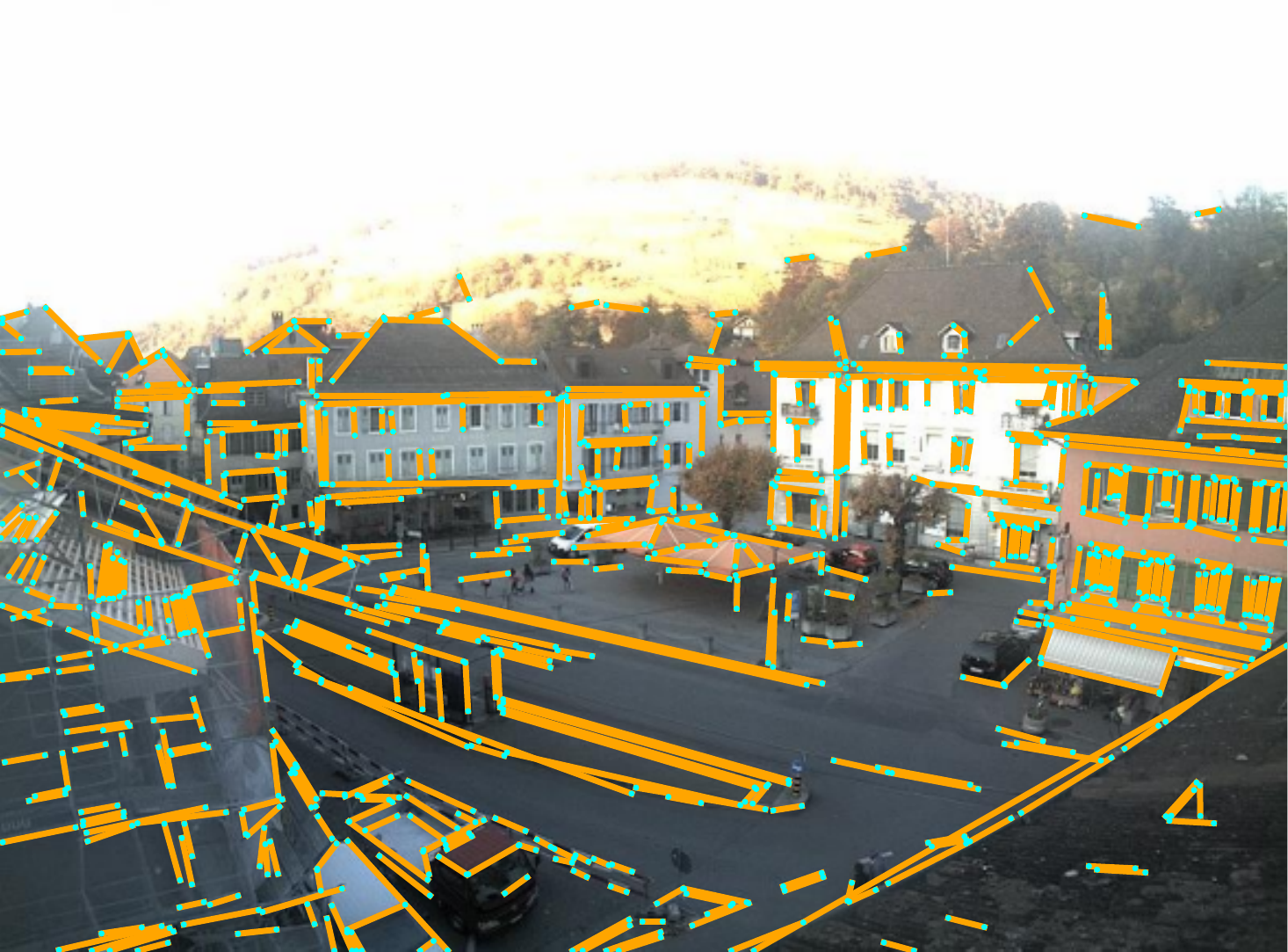}
        & \includegraphics[width=0.19\textwidth]{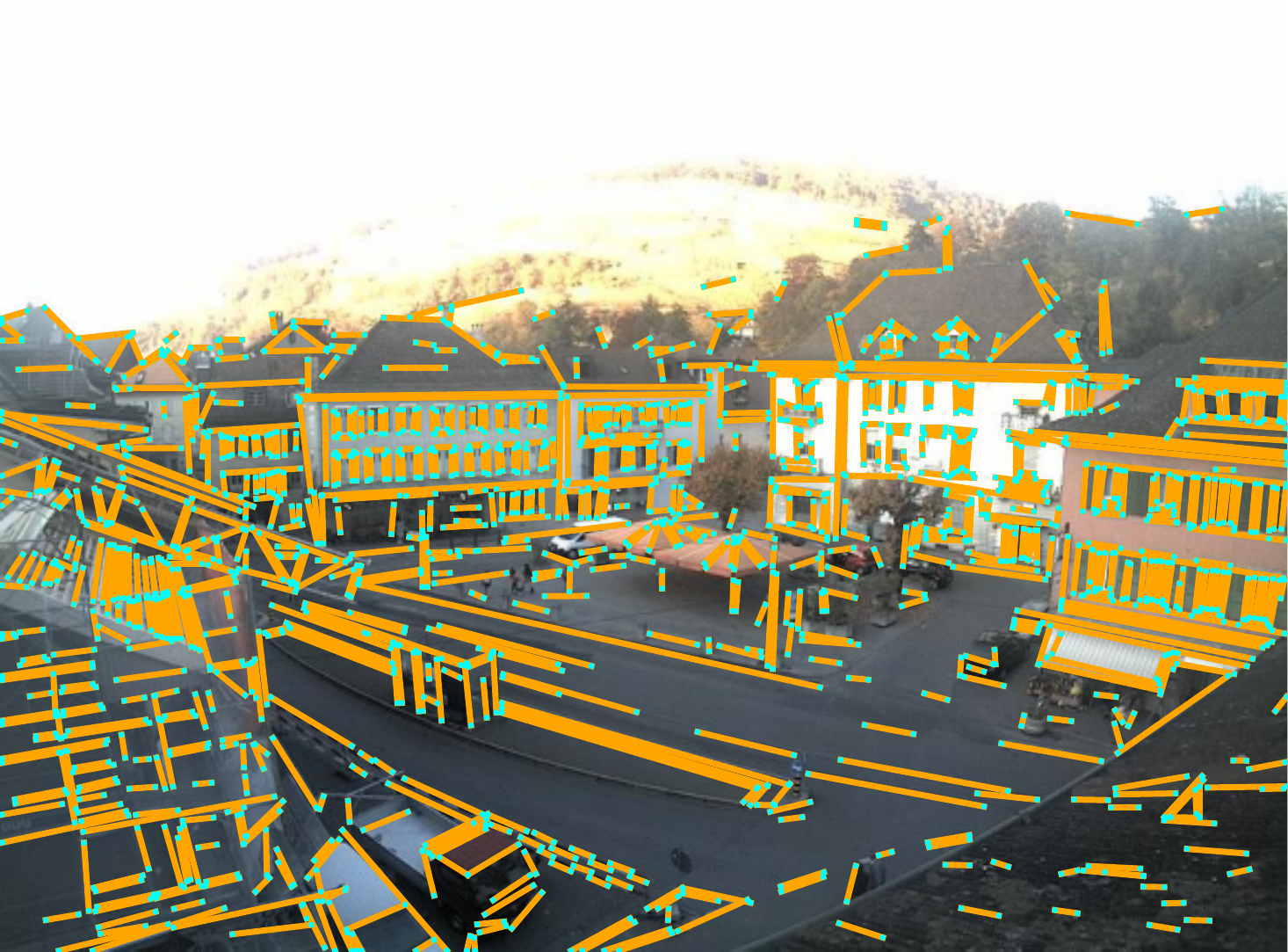} \\
    \end{tabular}
    \caption{\textbf{Visual comparison of line detectors.} \textbf{First five rows:} the lines of DeepLSD (here, without line refinement) are more complete and accurate in urban scenarios (images from the YorkUrbanDB dataset~\cite{yorkurban}). \textbf{Last three rows:} when employed in challenging scenarios such as by night, over-exposition and low image quality, DeepLSD can detect more relevant lines than the other baselines (images from the Day-Night Image Matching (DNIM) dataset~\cite{Zhou2016W}).}
    \label{fig:supp_visualizations}
\end{figure*}

%
We provide a visual comparison of our method and the other baselines for line detection in Figure~\ref{fig:supp_visualizations}. 
We first show line detection examples from the YorkUrbanDB dataset~\cite{yorkurban}, picturing indoor and outdoor urban scenes. DeepLSD offers more complete and accurate lines than its competitors. We also compare our method to the other line detectors on some images of the Day-Night Image Matching dataset~\cite{Zhou2016W}, where DeepLSD provides more lines than the other baselines in challenging scenarios such as night time, over-exposition and low image quality.

{\small
\bibliographystyle{ieee_fullname}
\bibliography{arxiv}

\begin{thebibliography}{10}\itemsep=-1pt

\bibitem{abdellali_2021}
Hichem Abdellali, Robert Frohlich, Viktor Vilagos, and Zoltan Kato.
\newblock {L2D2}: Learnable line detector and descriptor.
\newblock In {\em International Conference on 3D Vision (3DV)}, 2021.

\bibitem{agarap2018deep}
Abien~Fred Agarap.
\newblock Deep learning using rectified linear units ({ReLU}).
\newblock In {\em arXiv}, 2018.

\bibitem{ceres}
Sameer Agarwal and Keir Mierle.
\newblock Ceres solver.
\newblock \url{http://ceres-solver.org}.

\bibitem{akinlar2011edlines}
C. {Akinlar} and C. {Topal}.
\newblock {EDLines}: Real-time line segment detection by edge drawing.
\newblock In {\em International Conference on Image Processing (ICIP)}, 2011.

\bibitem{baker_2007}
Simon Baker, Daniel Scharstein, J.~P. Lewis, Stefan Roth, Michael~J. Black, and
  Richard Szeliski.
\newblock A database and evaluation methodology for optical flow.
\newblock In {\em International Conference on Computer Vision (ICCV)}, 2007.

\bibitem{hpatches_2017_cvpr}
Vassileios Balntas, Karel Lenc, Andrea Vedaldi, and Krystian Mikolajczyk.
\newblock Hpatches: A benchmark and evaluation of handcrafted and learned local
  descriptors.
\newblock In {\em Computer Vision and Pattern Recognition (CVPR)}, 2017.

\bibitem{Barath_2019_ICCV}
Daniel Barath and Jiri Matas.
\newblock Progressive-{X}: Efficient, anytime, multi-model fitting algorithm.
\newblock In {\em International Conference on Computer Vision (ICCV)}, 2019.

\bibitem{dsacstar}
Eric Brachmann and Carsten Rother.
\newblock Visual camera re-localization from rgb and rgb-d images using dsac.
\newblock {\em IEEE Transactions on Pattern Analysis and Machine Intelligence
  (PAMI)}, 44, 2022.

\bibitem{Camposeco2018CPVR}
Federico Camposeco, Andrea Cohen, Marc Pollefeys, and Torsten Sattler.
\newblock {Hybrid Camera Pose Estimation}.
\newblock In {\em Computer Vision and Pattern Recognition (CVPR)}, 2018.

\bibitem{dai2021fully}
Xili Dai, Xiaojun Yuan, Haigang Gong, and Yi Ma.
\newblock Fully convolutional line parsing.
\newblock In {\em arXiv}, 2021.

\bibitem{yorkurban}
Patrick Denis, James~H Elder, and Francisco~J Estrada.
\newblock Efficient edge-based methods for estimating manhattan frames in urban
  imagery.
\newblock In {\em European Conference on Computer Vision (ECCV)}, 2008.

\bibitem{superpoint}
Daniel DeTone, Tomasz Malisiewicz, and Andrew Rabinovich.
\newblock {SuperPoint}: Self-supervised interest point detection and
  description.
\newblock In {\em Computer Vision and Pattern Recognition Workshops (CVPRW)},
  2018.

\bibitem{Elder2020MCMLSDAP}
James~H. Elder, Emilio~J. Almaz{\'a}n, Yiming Qian, and Ron Tal.
\newblock {MCMLSD}: A probabilistic algorithm and evaluation framework for line
  segment detection.
\newblock In {\em arXiv}, 2020.

\bibitem{fu2020plvins}
Qiang Fu, Jialong Wang, Hongshan Yu, Islam Ali, Feng Guo, Yijia He, and Hong
  Zhang.
\newblock {PL-VINS}: Real-time monocular visual-inertial {SLAM} with point and
  line features.
\newblock In {\em arXiv}, 2020.

\bibitem{Gao2021PoseRW}
Shuang Gao, Jixiang Wan, Yishan Ping, Xudong Zhang, Shuzhou Dong, Jijunnan Li,
  and Yandong Guo.
\newblock Pose refinement with joint optimization of visual points and lines.
\newblock In {\em arXiv}, 2021.

\bibitem{gomez_2019}
Ruben Gomez-Ojeda, Francisco-Angel Moreno, David Zuñiga-Noël, Davide
  Scaramuzza, and Javier Gonzalez-Jimenez.
\newblock {PL-SLAM}: A stereo {SLAM} system through the combination of points
  and line segments.
\newblock {\em IEEE Transactions on Robotics}, 35, 2019.

\bibitem{gu2021realtime}
Geonmo Gu, Byungsoo Ko, SeoungHyun Go, Sung-Hyun Lee, Jingeun Lee, and Minchul
  Shin.
\newblock Towards real-time and light-weight line segment detection.
\newblock In {\em Conference on Artificial Intelligence (AAAI)}, 2022.

\bibitem{hofer_2017}
Manuel Hofer, Michael Maurer, and Horst Bischof.
\newblock Efficient 3d scene abstraction using line segments.
\newblock {\em Computer Vision and Image Understanding (CVIU)}, 157, 2017.

\bibitem{hough1962}
Paul~VC Hough.
\newblock Method and means for recognizing complex patterns, 1962.
\newblock US Patent 3,069,654.

\bibitem{wireframe}
Kun Huang, Yifan Wang, Zihan Zhou, Tianjiao Ding, Shenghua Gao, and Yi Ma.
\newblock Learning to parse wireframes in images of man-made environments.
\newblock In {\em Computer Vision and Pattern Recognition (CVPR)}, 2018.

\bibitem{huang2020tp}
Siyu Huang, Fangbo Qin, Pengfei Xiong, Ning Ding, Yijia He, and Xiao Liu.
\newblock {TP-LSD}: Tri-points based line segment detector.
\newblock In {\em European Conference on Computer Vision (ECCV)}, 2020.

\bibitem{huang2021vs}
Zhaoyang Huang, Han Zhou, Yijin Li, Bangbang Yang, Yan Xu, Xiaowei Zhou, Hujun
  Bao, Guofeng Zhang, and Hongsheng Li.
\newblock {VS-Net}: Voting with segmentation for visual localization.
\newblock In {\em Computer Vision and Pattern Recognition (CVPR)}, 2021.

\bibitem{ioffe_2015}
Sergey Ioffe and Christian Szegedy.
\newblock Batch normalization: Accelerating deep network training by reducing
  internal covariate shift.
\newblock In {\em International Conference on Machine Learning (ICML)}, 2015.

\bibitem{kingma2014}
Diederik Kingma and Jimmy Ba.
\newblock Adam: A method for stochastic optimization.
\newblock {\em International Conference on Learning Representations (ICLR)},
  2014.

\bibitem{kluger2020consac}
Florian Kluger, Eric Brachmann, Hanno Ackermann, Carsten Rother, Michael~Ying
  Yang, and Bodo Rosenhahn.
\newblock {CONSAC}: Robust multi-model fitting by conditional sample consensus.
\newblock In {\em Computer Vision and Pattern Recognition (CVPR)}, 2020.

\bibitem{kukelova2016efficient}
Zuzana Kukelova, Jan Heller, and Andrew Fitzgibbon.
\newblock Efficient intersection of three quadrics and applications in computer
  vision.
\newblock In {\em Computer Vision and Pattern Recognition (CVPR)}, 2016.

\bibitem{Lange_2019}
Manuel Lange, Claudio Raisch, and Andreas Schilling.
\newblock {LVO}: Line only stereo visual odometry.
\newblock In {\em 2019 International Conference on Indoor Positioning and
  Indoor Navigation (IPIN)}, 2019.

\bibitem{PoseLib}
Viktor Larsson.
\newblock {PoseLib - Minimal Solvers for Camera Pose Estimation}, 2020.

\bibitem{Lebeda2012loransac}
Karel Lebeda, Jiri Matas, and Ondrej Chum.
\newblock {Fixing the Locally Optimized RANSAC}.
\newblock In {\em British Machine Vision Conference (BMVC)}, 2012.

\bibitem{hao_2021}
Hao Li, Huai Yu, Jinwang Wang, Wen Yang, Lei Yu, and Sebastian Scherer.
\newblock {ULSD}: Unified line segment detection across pinhole, fisheye, and
  spherical cameras.
\newblock {\em Journal of Photogrammetry and Remote Sensing (ISPRS)}, 178,
  2021.

\bibitem{MegaDepthLi18}
Zhengqi Li and Noah Snavely.
\newblock {MegaDepth}: Learning single-view depth prediction from internet
  photos.
\newblock In {\em Computer Vision and Pattern Recognition (CVPR)}, 2018.

\bibitem{deephough}
Yancong Lin, Silvia~L Pintea, and Jan~C van Gemert.
\newblock Deep hough-transform line priors.
\newblock In {\em European Conference on Computer Vision (ECCV)}, 2020.

\bibitem{lindenberger2021pixsfm}
Philipp Lindenberger, Paul-Edouard Sarlin, Viktor Larsson, and Marc Pollefeys.
\newblock {Pixel-Perfect Structure-from-Motion with Featuremetric Refinement}.
\newblock In {\em International Conference on Computer Vision (ICCV)}, 2021.

\bibitem{mateus_2022}
André Mateus, Omar Tahri, A.~Pedro Aguiar, Pedro~U. Lima, and Pedro Miraldo.
\newblock On incremental structure from motion using lines.
\newblock {\em IEEE Transactions on Robotics}, 38, 2022.

\bibitem{meng_2020}
Quan Meng, Jiakai Zhang, Qiang Hu, Xuming He, and Jingyi Yu.
\newblock {LGNN}: A context-aware line segment detector.
\newblock In {\em ACM International Conference on Multimedia}, 2020.

\bibitem{micusik_2017}
Branislav Micusik and Horst Wildenauer.
\newblock Structure from motion with line segments under relaxed endpoint
  constraints.
\newblock {\em International Journal of Computer Vision (IJCV)}, 124, 2017.

\bibitem{silberman_2012}
Pushmeet~Kohli Nathan~Silberman, Derek~Hoiem and Rob Fergus.
\newblock Indoor segmentation and support inference from rgbd images.
\newblock In {\em European Conference on Computer Vision (ECCV)}, 2012.

\bibitem{Pautrat_2020_ECCV}
Rémi Pautrat, Viktor Larsson, Martin~R. Oswald, and Marc Pollefeys.
\newblock Online invariance selection for local feature descriptors.
\newblock In {\em European Conference on Computer Vision (ECCV)}, 2020.

\bibitem{Pautrat_Lin_2021_CVPR}
Rémi Pautrat, Juan-Ting Lin, Viktor Larsson, Martin~R. Oswald, and Marc
  Pollefeys.
\newblock {SOLD2}: Self-supervised occlusion-aware line description and
  detection.
\newblock In {\em Computer Vision and Pattern Recognition (CVPR)}, 2021.

\bibitem{pumarola_2017}
Albert Pumarola, Alexander Vakhitov, Antonio Agudo, Alberto Sanfeliu, and
  Francese Moreno-Noguer.
\newblock {PL-SLAM}: Real-time monocular visual {SLAM} with points and lines.
\newblock In {\em International Conference on Robotics and Automation (ICRA)},
  2017.

\bibitem{quan_2021}
Meixiang Quan, Zheng Chai, and Xiao Liu.
\newblock {LOF}: Structure-aware line tracking based on optical flow.
\newblock In {\em arXiv}, 2021.

\bibitem{roberts_2021}
Mike Roberts, Jason Ramapuram, Anurag Ranjan, Atulit Kumar, Miguel~Angel
  Bautista, Nathan Paczan, Russ Webb, and Joshua~M. Susskind.
\newblock {Hypersim}: {A} photorealistic synthetic dataset for holistic indoor
  scene understanding.
\newblock In {\em International Conference on Computer Vision (ICCV)}, 2021.

\bibitem{ronneberger_2015}
Olaf Ronneberger, Philipp Fischer, and Thomas Brox.
\newblock {U-Net}: Convolutional networks for biomedical image segmentation.
\newblock In {\em Medical Image Computing and Computer-Assisted Intervention
  (MICCAI)}, 2015.

\bibitem{salaun_2016}
Yohann Salaün, Renaud Marlet, and Pascal Monasse.
\newblock Multiscale line segment detector for robust and accurate {SfM}.
\newblock In {\em International Conference on Pattern Recognition (ICPR)},
  2016.

\bibitem{hloc}
Paul-Edouard Sarlin.
\newblock Visual localization made easy with hloc.
\newblock \url{https://github.com/cvg/Hierarchical-Localization/}.

\bibitem{sarlin2019coarse}
Paul-Edouard Sarlin, Cesar Cadena, Roland Siegwart, and Marcin Dymczyk.
\newblock From coarse to fine: Robust hierarchical localization at large scale.
\newblock In {\em Computer Vision and Pattern Recognition (CVPR)}, 2019.

\bibitem{sarlin_2020_superglue}
Paul-Edouard Sarlin, Daniel DeTone, Tomasz Malisiewicz, and Andrew Rabinovich.
\newblock {SuperGlue}: Learning feature matching with graph neural networks.
\newblock In {\em Computer Vision and Pattern Recognition (CVPR)}, June 2020.

\bibitem{sarlin21pixloc}
Paul-Edouard Sarlin, Ajaykumar Unagar, Måns Larsson, Hugo Germain, Carl Toft,
  Victor Larsson, Marc Pollefeys, Vincent Lepetit, Lars Hammarstrand, Fredrik
  Kahl, and Torsten Sattler.
\newblock {Back to the Feature: Learning Robust Camera Localization from Pixels
  to Pose}.
\newblock In {\em Computer Vision and Pattern Recognition (CVPR)}, 2021.

\bibitem{Sattler2019Github}
Torsten Sattler et~al.
\newblock {RansacLib - A Template-based *SAC Implementation}, 2019.

\bibitem{7scenes}
Jamie Shotton, Ben Glocker, Christopher Zach, Shahram Izadi, Antonio Criminisi,
  and Andrew Fitzgibbon.
\newblock Scene coordinate regression forests for camera relocalization in
  rgb-d images.
\newblock In {\em Computer Vision and Pattern Recognition (CVPR)}, 2013.

\bibitem{suarez2021elsed}
Iago Suárez, José~M. Buenaposada, and Luis Baumela.
\newblock {ELSED}: Enhanced line segment drawing.
\newblock {\em Pattern Recognition}, 2022.

\bibitem{Tardif_2009}
Jean-Philippe Tardif.
\newblock Non-iterative approach for fast and accurate vanishing point
  detection.
\newblock In {\em International Conference on Computer Vision (ICCV)}, 2009.

\bibitem{teplyakov2022}
Lev Teplyakov, Leonid Erlygin, and Evgeny Shvets.
\newblock Lsdnet: Trainable modification of lsd algorithm for real-time line
  segment detection.
\newblock {\em IEEE Access}, 10, 2022.

\bibitem{von2008lsd}
Rafael~Grompone Von~Gioi, Jeremie Jakubowicz, Jean-Michel Morel, and Gregory
  Randall.
\newblock {LSD}: A fast line segment detector with a false detection control.
\newblock {\em IEEE Transactions on Pattern Analysis and Machine Intelligence
  (PAMI)}, 32(4):722--732, 2008.

\bibitem{Xu_2021_CVPR}
Yifan Xu, Weijian Xu, David Cheung, and Zhuowen Tu.
\newblock Line segment detection using transformers without edges.
\newblock In {\em Computer Vision and Pattern Recognition (CVPR)}, 2021.

\bibitem{afm}
Nan Xue, Song Bai, Fudong Wang, Gui-Song Xia, Tianfu Wu, and Liangpei Zhang.
\newblock Learning attraction field representation for robust line segment
  detection.
\newblock In {\em Computer Vision and Pattern Recognition (CVPR)}, 2019.

\bibitem{hawp}
Nan Xue, Tianfu Wu, Song Bai, Fudong Wang, Gui-Song Xia, Liangpei Zhang, and
  Philip~HS Torr.
\newblock Holistically-attracted wireframe parsing.
\newblock In {\em Computer Vision and Pattern Recognition (CVPR)}, 2020.

\bibitem{HAWP-journal}
Nan Xue, Tianfu Wu, Song Bai, Fu-Dong Wang, Gui-Song Xia, Liangpei Zhang, and
  Philip~H.S. Torr.
\newblock Holistically-attracted wireframe parsing: From supervised to
  self-supervised learning.
\newblock {\em arXiv}, 2022.

\bibitem{Zhang_2021_ICCV}
Haotian Zhang, Yicheng Luo, Fangbo Qin, Yijia He, and Xiao Liu.
\newblock Elsd: Efficient line segment detector and descriptor.
\newblock In {\em International Conference on Computer Vision (ICCV)}, 2021.

\bibitem{zhang2013lbd}
Lilian Zhang and Reinhard Koch.
\newblock An efficient and robust line segment matching approach based on lbd
  descriptor and pairwise geometric consistency.
\newblock {\em Journal of Visual Communication and Image Representation}, 24,
  2013.

\bibitem{zhang2021}
Yongjun Zhang, Dong Wei, and Yansheng Li.
\newblock {AG3line}: Active grouping and geometry-gradient combined validation
  for fast line segment extraction.
\newblock {\em Pattern Recognition}, 113, 2021.

\bibitem{zhang2019ppgnet}
Ziheng Zhang, Zhengxin Li, Ning Bi, Jia Zheng, Jinlei Wang, Kun Huang, Weixin
  Luo, Yanyu Xu, and Shenghua Gao.
\newblock Ppgnet: Learning point-pair graph for line segment detection.
\newblock In {\em Computer Vision and Pattern Recognition (CVPR)}, 2019.

\bibitem{Zhou2016W}
Hao Zhou, Torsten Sattler, and David~W. Jacobs.
\newblock Evaluating local features for day-night matching.
\newblock In {\em European Conference on Computer Vision Workshops (ECCVW)},
  2016.

\bibitem{Zhou2018ASA}
Lipu Zhou, Jiamin Ye, and Michael Kaess.
\newblock A stable algebraic camera pose estimation for minimal configurations
  of 2d/3d point and line correspondences.
\newblock In {\em Asian Conference on Computer Vision (ACCV)}, 2018.

\bibitem{lcnn}
Yichao Zhou, Haozhi Qi, and Yi Ma.
\newblock End-to-end wireframe parsing.
\newblock In {\em International Conference on Computer Vision (ICCV)}, 2019.

\bibitem{zuo_2017}
Xingxing Zuo, Xiaojia Xie, Yong Liu, and Guoquan Huang.
\newblock Robust visual {SLAM} with point and line features.
\newblock In {\em International Conference on Intelligent Robots and Systems
  (IROS)}, 2017.

\end{thebibliography}
}

\end{document}